\documentclass[10pt, floatfix,amsmath,amssymb,superscriptaddress,longbibliography,twocolumn,showpacs,pre, aps]{revtex4-1}
\usepackage{amssymb}
\usepackage{amsmath}
\usepackage{bbold}
\usepackage{bm}
\usepackage{mathtools}
\usepackage{bbm}
\usepackage{braket}
\usepackage[title]{appendix}
\usepackage[svgnames]{xcolor}
\usepackage{soul}
\usepackage{graphicx}
\usepackage{textcomp}
\usepackage{xcolor}
\usepackage{enumerate}
\usepackage{multirow}
\usepackage[colorlinks=true, urlcolor=blue]{hyperref}
\usepackage{lipsum, babel}
\graphicspath{{../figura1/}{../figura2/}}

\usepackage{float}
\usepackage{algorithm}
\usepackage{algpseudocode}

\newcommand*\Tr{\mathop{}\!\mathrm{Tr}}
\newcommand{\tr}{\intercal}

\newcommand{\Id}{\mathbbmss{1}}

\usepackage[final]{changes}
\newcommand{\newtext}[1]{{#1}} 

\definechangesauthor[color=teal]{CK}
\definechangesauthor[color=blue]{PB}
\definechangesauthor[color=red]{rev1}  

\sethlcolor{LightGoldenrodYellow} 
\soulregister\cite7 
\soulregister\ref7
\soulregister\eqref7

\usepackage{cancel}

\begin{document}

\title{Kernel Renormalization in Bayesian Deep Neural Networks: the Equivalent Wishart Ansatz in the Proportional Regime}

\author{P. Baglioni}
\thanks{These authors contributed equally to this work.}
\affiliation{INFN, Sezione di Milano Bicocca, Piazza della Scienza 3, 20126, Milano, Italy}
\affiliation{INFN, Gruppo Collegato di Parma, Parco Area delle Scienze 7/A, 43124 Parma, Italy}

\author{C. Keup}
\thanks{These authors contributed equally to this work.}
\affiliation{INFN, Sezione di Milano Bicocca, Piazza della Scienza 3, 20126, Milano, Italy}
\affiliation{INFN, Gruppo Collegato di Parma, Parco Area delle Scienze 7/A, 43124 Parma, Italy}

\author{V. Zimbardo}
\thanks{These authors contributed equally to this work.}
\affiliation{Dipartimento di Scienze Matematiche, Fisiche e Informatiche,
Universit\`a degli Studi di Parma, Parco Area delle Scienze, 7/A 43124 Parma, Italy}
\affiliation{INFN, Sezione di Milano Bicocca, Piazza della Scienza 3, 20126, Milano, Italy}
\affiliation{INFN, Gruppo Collegato di Parma, Parco Area delle Scienze 7/A, 43124 Parma, Italy}

\author{R. Pacelli}
\thanks{These authors contributed equally to this work.}
\affiliation{INFN, sezione di Padova, Via Marzolo 8, 35131 Padova, Italy}

\author{A. Vezzani}
\affiliation{Istituto dei Materiali per l'Elettronica ed il Magnetismo (IMEM-CNR), Parco Area delle Scienze, 37/A-43124 Parma, Italy}
\affiliation{Dipartimento di Scienze Matematiche, Fisiche e Informatiche,
Universit\`a degli Studi di Parma, Parco Area delle Scienze, 7/A 43124 Parma, Italy}
\affiliation{INFN, Gruppo Collegato di Parma, Parco Area delle Scienze 7/A, 43124 Parma, Italy}

\author{R. Burioni}
\affiliation{Dipartimento di Scienze Matematiche, Fisiche e Informatiche,
Universit\`a degli Studi di Parma, Parco Area delle Scienze, 7/A 43124 Parma, Italy}
\affiliation{INFN, Sezione di Milano Bicocca, Piazza della Scienza 3, 20126, Milano, Italy}
\affiliation{INFN, Gruppo Collegato di Parma, Parco Area delle Scienze 7/A, 43124 Parma, Italy}

\author{P. Rotondo}
\affiliation{Dipartimento di Scienze Matematiche, Fisiche e Informatiche,
Universit\`a degli Studi di Parma, Parco Area delle Scienze, 7/A 43124 Parma, Italy}
\affiliation{INFN, Sezione di Milano Bicocca, Piazza della Scienza 3, 20126, Milano, Italy}
\affiliation{INFN, Gruppo Collegato di Parma, Parco Area delle Scienze 7/A, 43124 Parma, Italy}

\begin{abstract}
The scaling limit where both the size of the training set $P$ and the width $N$ of a deep neural network grow at the same rate, the so-called proportional-width regime, has been intensely studied for shallow, single-hidden-layer networks. However, extending these non-perturbative results from shallow architectures to deep non-linear networks has proven very challenging. Here we present an effective \emph{approximate} approach to predict the generalization performance of Bayesian multi-layer perceptrons (MLPs) of fixed depth $L$ on \replaced[id=rev1]{high-dimensional real}{arbitrary high-dimensional} data. We propose an \emph{equivalent Wishart Ansatz} to capture the dominant stochastic fluctuations of the hierarchical empirical kernels of MLPs. This allows us to perform a large deviation analysis for the \added[id=rev1]{approximated} partition function of MLPs in the proportional limit, expressed in terms of a renormalized NNGP kernel. In this description, even strong representation learning in the proportional limit is encoded in at most $L$ scalar order parameters, determined self-consistently. Extending the approach to  convolutional architectures (CNNs), we identify a hierarchical local kernel renormalization mechanism, which allows to quantify more complex data-dependent transformations of the large-width kernel in CNNs due to finite-width effects. We test our effective theory against sampling experiments from the Bayesian posterior of finite deep neural networks with depths $L \sim O(10)$ and $P\sim O(10^3)$ on classic benchmark datasets, finding overall very good agreement together with two distinct types of systematic deviations.
\end{abstract}

\maketitle

\section{Introduction}

Deep neural networks (DNNs) have emerged as a central technology to extract statistical regularities from data at unprecedented scale.
The core promise of this machine learning technique is that, aside from a computational speedup compared to classical kernel methods, DNNs can adapt their internal feature representation during training, and thereby approximate a kernel method where the kernel has been automatically adapted for each task (for good or worse).
However, the physical laws 
governing this kernel adaptation, and therefore the precise capabilities and limitations of this method, have turned out to be highly resistant to theoretical inquiry for networks that are both multi-layered and nonlinear.

DNNs are objects well suited for study via statistical physics techniques, because the network size, the dimensionality of input data, and the number of data examples are all simultaneously large. Knowing the physical laws of the system then amounts to finding an effective description in terms of low-dimensional order parameters which capture the system behavior.  What is less obvious, is the best large system limit in which to study deep networks, that is the relative scaling of layer sizes, initialization parameters, training procedures, and data set sizes to infinity.

The large variety of choices, affecting both conclusions and tractability, can be significantly reduced by adopting the Bayesian viewpoint of asking for the posterior given a prior over parameters closely related to the initialization scheme. In this way, the need to choose a training algorithm and its hyperparameter scalings is removed. 
\deleted[id=rev1]{Furthermore, while a}\added[id=rev1]{A}lgorithmic biases constitute an important field of research \cite{keskar2017,gunasekar18a,Jiang2020Fantastic,Lyu2020homogeneous,Agarwala2024}\replaced[id=rev1]{. However, the Bayesian network posterior at low temperature is a useful proxy of the generalization capabilities of a network architecture, typically offering comparable generalization to trained networks \cite{Avidan2025,wenzel2020how, bayesSGD}.}{, it has been shown that such biases may quantitatively but not qualitatively alter the performance compared to realizations drawn from the Bayesian network posterior \cite{Avidan2025}.}

We can then sketch the landscape of limits for Bayesian networks as follows: 
In the Neural Network Gaussian Process (NNGP) limit, also termed "lazy" infinite-width limit, where all layer widths are sent to infinity while the dataset is fixed, the network becomes equivalent to Gaussian process regression with a fixed NNGP kernel determined by the architecture \cite{Neal1995BayesianLF, LeeGaussian, matthews2018, novak2019bayesian, Hron2020}.
The lazy regime was contrasted by another, "rich" infinite width limit, where the network output in the prior or at initialization is scaled down to vanish with width, forcing a strong adaptation of the network features to fit the $O(1)$ target labels \cite{MeiMontanari18,Rotskoff18,ChizatBach18,tensoriv}.
However, while realistic deep networks are large, their width is typically not large compared to the number of data samples as assumed by these limits. Indeed, it was proposed that finite-width effects are responsible for the advantages of DNNs over kernel methods \cite{neurips_empirical_study}.
Another way to capture these effects is the proportional limit, where layer widths, input dimension, and sample size scale proportionally to infinity. This limit is classical in the statistical physics of shallow (at most one hidden layer) networks \cite{Gardner1988,Hopfield1982,Schwarze1993,Monasson1995}. 
For DNNs, which have a number of parameters quadratic in the hidden-layer width, also supraproportional limits with faster sample-size scaling could in some cases be relevant \cite{cui2023optimal,camilli2025information}. These are unexplored and expected to be extremely difficult to study for non-random data.

In this work, we focus on finite-width, multi-layer and nonlinear DNNs, adopting the proportional limit, and propose an effective low-dimensional, albeit approximate, theory. Considerable theoretical understanding exists when any one of these three ingredients is removed: nonlinearity, as in deep linear networks \cite{SompolinskyLinear,Bassetti:JMLR:2024, Saxelinear,HaninZlokapa23,bordelon25a,rubin25a}; depth, as in single-index models \cite{Barbier2019a,Mondelli2019,bietti22,troiani2024} and shallow networks \cite{MeiMontanari18,Rotskoff18,MeiMontanari21,cui2024asymptotics,Defilippis2025}; or finite-width effects, as in the infinite-width NNGP limit \cite{Neal1995BayesianLF,LeeGaussian,matthews2018,canatar2021,Misiakiewicz2024,Karkada2025}. 
The simultaneous presence of all three, arguably central for deep learning, so far evaded low-dimensional description however.
By choosing an approximate, Ansatz based theory, we trade off asymptotic exactness for practical insight into nonlinear DNN generalization when trained on real datasets.

\subsection{The proportional regime of Bayesian DNNs: previous results and open problems} \label{sec:proportional-limit-related-work}

The proportional limit in Bayesian deep linear neural networks was first considered in Ref.~\cite{SompolinskyLinear}. Here, the authors introduced a method, the \emph{Backpropagating kernel renormalization}, which allows for the incremental integration of the network weights layer by layer starting from the network output layer and progressing backward until the first layer’s weights are integrated out. Using this approach, properties of the network at fixed $\alpha = P/N$ are evaluated via a self-consistent equation for a one-dimensional order parameter.

Afterwards, it was shown that the validity of this approach in the proportional regime is a consequence of an exact representation for the partition function of deep linear networks at finite $N$ and $P$. Such a representation was first given in terms of Meijer-G functions for the case of fully-connected architectures with a single readout neuron \cite{zavatone-veth2021exact,HaninZlokapa23}, and later generalized to the case of networks with multiple outputs and convolutional layers \cite{Bassetti:JMLR:2024}. Here the authors noted that the joint prior distribution of the outputs for a finite deep linear architecture with $L$ hidden layers can be expressed as a mixture of Gaussians over $L$ low-dimensional positive-definite matrix ensembles. This exact integral representation for the prior directly transfers to the partition function in the case of Gaussian likelihood (square-loss), and it provides a mathematically rigorous framework for the Backpropagating kernel renormalization method introduced in \cite{SompolinskyLinear}.
\begin{figure*}
    \includegraphics[width=\textwidth]{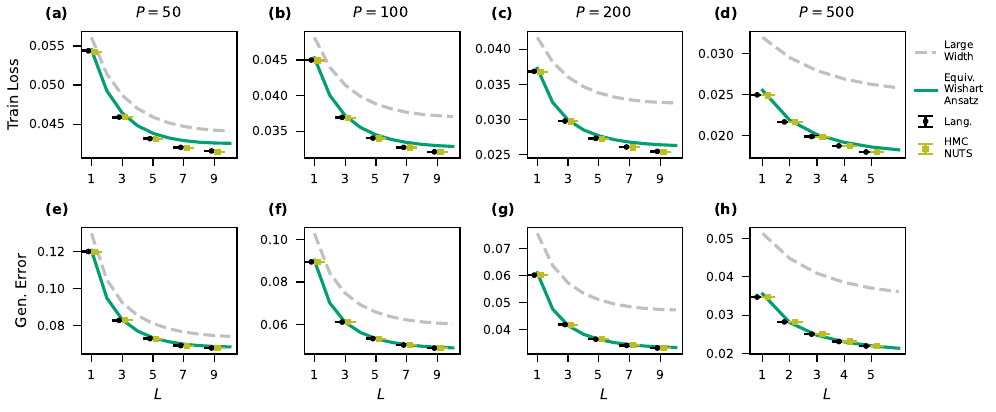} 
    \caption{\newtext{\textbf{Comparison between the learning curves obtained via the Equivalent Wishart Ansatz and numerical experiments for zero-mean activation functions on MNIST.} Numerical samples from the Bayesian posterior (black and yellow markers -- shifted slightly for better visibility) are compared against the large-width limit predictions (gray dashed lines) and the results of the EWA theory (green solid lines). Both the training loss (first row) and the generalization error (second row) are displayed as a function of the number of hidden layers $L$. We keep the number of neurons and test examples fixed at $N_\ell=200\, \forall \ell$ and $P_t=1000$, while varying the number of patterns $P$ across different columns ($P$ is constant within each column). These simulations refer to the Erf activation function, with Gaussian priors $\lambda = 1$ and temperature $T=0.1$. For all panels, we sample from the posterior using Langevin Monte Carlo with a learning rate $\eta=0.001$ (black dots) and HMC NUTS (yellow squares). The aggregated goodness-of-fit index value $R^2$ is 0.993 (train) and 0.999 (test).}}
    \label{fig:main-erf-mnist}
\end{figure*}

In Ref. \cite{SompolinskyLinear}, the authors also proposed a heuristic extension of the theory to fully-connected DNNs with ReLU activation, based on the observation that $\mathrm{ReLU}(\gamma x) = \gamma \mathrm{ReLU}(x)$, $\forall \, \gamma > 0$. Surprisingly, the resulting theory, which is obtained replacing the trivial NNGP linear kernel with the ReLU one, predicts the generalization performance of finite DNNs with ReLU activation for modest number of training patterns $P \sim 10^2$ and depth $L < 5$.

The work in Ref. \cite{pacelli2023statistical} showed that a variational low-dimensional free-energy (or effective action) emerges in non-linear one-hidden layer networks ($L = 1$) with generic activation function, thanks to a Gaussian equivalence informally justified using Breuer-Major theorems \cite{BM,bardet2013}. This framework was successfully generalized later to investigate 1HL convolutional architectures \cite{aiudi2023}, FC networks with multiple outputs \cite{baglioni2025kernel, baglioni2024predictive}, transfer and continual learning \cite{PhysRevLett.134.177301, doi:10.1073/pnas.2501899123}. In all these cases, new interesting forms of kernel shape renormalization arise, at variance with the 1HL fully-connected (and single output) case, where the renormalized kernel is found as a global, data-dependent rescaling of the NNGP kernel.

Crucially, neither Ref. \cite{SompolinskyLinear} nor Ref. \cite{pacelli2023statistical} elaborate convincing and principled arguments to establish whether an equivalent dimensional reduction of the partition function to the one found for $L = 1$ holds for deeper ($L > 1$) architectures with generic activation function. 

Another line of work seeking to explain feature learning in DNNs focuses not solely on going beyond the infinite width limit in the standard parametrization, but also on using a different scaling of the network output with $N$, termed the mean-field \cite{MeiMontanari18,Rotskoff18,ChizatBach18} or $\mu$P \cite{tensoriv} parametrization, which has become popular in the theoretical literature because even in the infinite-width limit the activations in all layers change by $O(1)$ during training. This has been termed the rich learning regime as the empirical kernels strongly differ between initialization and trained network, and between prior and posterior, indicating adaptation to the data. 
The empirical kernels here are defined as the Gram matrices of activations in a layer, and therefore encode the physically learned representation \cite{Ciceri2024, corti2025microscopic}. They are conceptually distinct from the effective kernels arising in the kernel shape renormalization approach, coinciding only in the NNGP limit, but both can be used to predict network outputs, as seen in linear networks where exact descriptions of both objects are available \cite{SompolinskyLinear,Bassetti:JMLR:2024,rubin25a}.
A series of ground-breaking papers has been seeking to predict the adaptation of the empirical kernels in nonlinear networks, yet by keeping the full $P\times P$ kernel matrices at each layer as high-dimensional observables \cite{seroussi2023natcomm,fischer24critical,lauditi25a,coding-schemes-2025}. For shallow $L=1$ networks, $P$-dimensional order parameters have been considered \cite{rubin25a,coding-schemes-2025}. In the mean-field setting, the empirical kernels concentrate in the infinite-width limit to nontrivial posterior configurations \cite{seroussi2023natcomm,lauditi25a,Andreis2026LDP,zavatoneveth2022contrasting}. In the standard parametrization, the more realistic proportional regime yields adaptation of the empirical kernels through finite-width fluctuations \cite{fischer24critical}. Overall, this approach is successful but recasts training and inference on a task in a coupled set of $P \times P$ dimensional self-consistency equations, which does not correspond to a dimensionality reduction with respect to the $N \times N$ parameter space in the proportional regime, leads to significant numerical difficulty in solving the self-consistency equations, and demonstrates the need for an effective low-dimensional theory for nonlinear DNNs.

\subsection{Summary of the major results}

In this work, we study the statistical mechanics of learning in overparameterized Bayesian DNNs with general nonlinearities in the hidden units, providing a low-dimensional approximate description of such architectures at finite width, thus going beyond the NNGP large-width limit. In particular, we make the following contributions: \vspace{.15cm} \\
(\textit{i}) We introduce the Equivalent Wishart Ansatz (EWA) in the proportional regime (Sec.~\ref{sec:EWA_for_MLP}), which provides an approximate description of the fluctuations of the hierarchical empirical kernels in nonlinear DNNs. Leveraging this novel framework, we obtain a low-dimensional integral representation of the characteristic function of the prior for generic $L$-layer NNs with fully connected (both scalar and multiple outputs) and convolutional layers (Sec.~\ref{sec:EWA_for_MLP}, \ref{subsec:multiple-outputs} and \ref{subsec:the-stacked-equivalent-wishart-ansatz-for-dnns-with-convolutional-layers}). We derive the corresponding Large Deviation Principle for the emerging $L$ order parameters, and test the validity of the EWA by comparing the theoretical rate functions with empirical simulations sampling the ground-truth MLP prior, showing convergence close to the asymptotic predicted behavior (Sec.~\ref{subsec:numerical-validation-ewa}). \vspace{.15cm} \\
(\textit{ii}) We identify different deep kernel renormalization schemes at the level of the prior, that encode distinct representation learning capabilities in deep architectures. In addition, we provide a low-dimensional expression for the posterior partition function in terms of an effective action (Sec.~\ref{subsec:effective-action-at-depth-l}), which explicitly depends only on the order parameters and the dataset, and non-perturbatively encodes the effect of depth and width. We conducted an extensive simulation campaign to assess the predictive power of the aforementioned theory, comparing the analytical learning curves obtained with the Equivalent Wishart Ansatz for the predictor statistics against Bayesian learning experiments employing finite networks\added[id=rev1]{, covering MLPs trained with square loss } (Sec.~\ref{subsec:learning-curves-for-fc-dnn})\added[id=rev1]{, multiple outputs and cross-entropy loss (Sec.~\ref{subsec:learning-curves-multiple-outputs-classification}) and CNNs (Sec.~\ref{subsec:learning-curves-for-cnns})}. Overall, we find that the theory captures the empirical behavior at finite width up to $L\sim O(10)$ and for different datasets, activation functions, and numbers of training examples. We also discuss the emergence of discrepancies when both the depth and the dataset size are simultaneously large (Sec.~\ref{subsec:emergent-performance-transitions}), in particular describing a so far unreported metastability phenomenon occurring in the posterior at moderate depths $L>5$. \vspace{.15cm} \\
(\textit{iii}) Using both the standard (SP) and mean-field ($\mu$P) parametrizations, we show that in the $P \sim N$ regime covered, a collective macroscopic description of feature learning is possible, without introducing $P \times P$ dimensional self-consistency equations (Sec.~\ref{subsec:mean-field-param}). In the EWA, improvements in generalization in the mean-field regime are shown to be driven by a suppression of the predictor variance as a function of the width, rather than by strong adaptation of the predictor bias. Overall, these Ansatz-based and empirically successful results pave the way for a low-dimensional and asymptotically exact theory of kernel adaptation in DNNs in the proportional regime.

\section{Setting of the problem and notations}

We consider deep neural networks with $L$ fully-connected hidden layers, where the pre-activations of each layer $h_{i_{\ell}}^{(\ell)}$ ($i_{\ell} = 1,\dots, N_{\ell}$; $\ell = 1, \dots, L$) are given recursively as a non-linear function of the pre-activations at the previous layer $h_{i_{\ell-1}}^{(\ell-1)}$ ($i_{\ell-1}= 1, \dots, N_{\ell-1}$):
\begin{align}
    \begin{aligned}
    h_{i_\ell}^{(\ell)}(x) &= \frac{1}{\sqrt {N_{\ell-1}}} \sum_{i_{\ell-1}=1}^{N_{\ell-1}} W^{(\ell)}_{i_{\ell}i_{\ell-1}} \sigma\!\left(h_{i_{\ell-1}}^{(\ell-1)}(x)\right) + b_{i_\ell}^{(\ell)}\,, \label{eq:preactivations-definition}\\
    h_{i_1}^{(1)}(x) &= \frac{1}{\sqrt {N_{0}}} \sum_{i_0=1}^{N_{0}} W^{(1)}_{i_1 i_0} x_{i_0} + b_{i_1}^{(1)}\ ,
    \end{aligned}
\end{align}
where $W^{(\ell)}$ and $b^{(\ell)}$ are respectively the weights and the biases of the $\ell$-th layer, whereas the input layer has dimension $N_0$ (the input data dimension). $\sigma$ is a non-linear activation function and it is common to each layer. For brevity, in the calculations we will restrict to the case without biases, $b^{(\ell)}=0$. We add one last readout layer and define the function implemented by the deep neural network as:
\begin{equation}
\label{eq:f_DNN}
f_{\theta} ( x) = \frac{1}{\sqrt{N_L}} \sum_{i_L=1}^{N_L} W^{(L+1)}_{i_L} \sigma \!\left[ h_{i_L}^{(L)} (x)\right]\,,
\end{equation}
where $W^{(L+1)}$ is the vector of weights of the last layer and $\theta$ indicates the collection of all the weights of the network, $\theta = \{ W^{(\ell)}\}_\ell$. 
We consider a supervised learning problem with fixed training set $\mathcal D_P = \{X,Y\}=\{x^\mu, y^\mu\}_{\mu=1}^P$, where each $x^\mu \in \mathbb R^{N_0}$ and the corresponding labels $y^\mu \in \mathbb R$. We analyse regression problems with 
squared-error loss function:
\begin{equation}
\mathcal L (\theta, \mathcal D_P) = \frac{1}{2} \sum_{\mu=1}^P \left[y^\mu - f_{\theta}(x^\mu)\right]^2 \,.
\label{eq:loss}
\end{equation}
As a standard practice in statistical mechanics of deep learning, we define the partition function of the problem as:
\begin{equation}
    Z_{\mathcal D_P} = \int  d p(\theta) \,e^{-\beta \mathcal L (\theta,  \mathcal D_P)}\ ,
\label{defpartition}
\end{equation}
where the symbol $d p(\theta)$ indicates the collective integration of the weights of the network over a prior measure and it will be used interchangeably with $d\theta\rho(\theta)$, being $\rho(\theta)$ the prior probability distribution. 
This has a natural Bayesian learning interpretation: the Gibbs probability $P_{\beta} (\theta  \mid \mathcal D_P )=Z^{-1} e^{-\beta \mathcal L (\theta, \mathcal D_P)} \rho(\theta)$ associated with the partition function in Eq.~\eqref{defpartition} is the posterior distribution of the weights. To keep the notation concise, we will often omit the explicit dependence on the dataset in the rest of the manuscript. Modeling the standard initialization scheme, the prior distribution over weights is Gaussian:
\begin{equation}
    W^{(\ell)}_{i_{\ell} i_{\ell-1}} \overset{\mathrm{i.i.d.}}{\sim} \mathcal N (0, 1/\lambda)\,, \label{eq:weight_prior}
\end{equation}
where the precision $\lambda$, the inverse of the variance, is the same at each layer for simplicity of the notation and as usually in practice. In this framework, the average test error over a new (unseen) example $(x^0, y^0)$ is given by:
\begin{equation}
\mathbb E_{\theta\mid\mathcal{D}_P}[\epsilon_{\textrm g} (x^0, y^0)] = \int d p(\theta) \, [y^0- f_{\theta}(x^0)]^2 \frac{e^{-\beta \mathcal L(\theta)}}{Z}\,.
\label{eq:err_g_def}
\end{equation}
In numerical experiments, we also consider the empirical generalization error (or simply the generalization error), which is defined as the average test error over $P_t$ different test examples, where $P_t$ denotes the number of patterns in the test set. The average training error at a given inverse temperature $\beta$ is given by:
\begin{equation}
\mathbb E_{\theta\mid\mathcal{D}_P}[\epsilon_{\textrm t}]
= \frac{1}{P} \sum_{\mu=1}^{P}\int d p(\theta) \, \left[y^\mu - f_{\theta}(x^\mu)\right]^2 \frac{e^{-\beta \mathcal L(\theta)} }{Z}\,.
\label{eq:err_t_def}
\end{equation}
Training and test errors represent two relevant observables that can be computed in the usual way by taking derivatives of the partition function with respect to appropriate source fields.

\section{Preamble: the Wishart distribution and its properties}

In statistics, the Wishart distribution arises as a matrix ensemble generalizing the Gamma distribution, and it is defined over symmetric, positive definite random matrices. Let us suppose that $G$ is a $P \times N$ matrix, and each column $G_i$ ($i = 1, \dots,\, N$) of $G$ is independently drawn from a $P$-dimensional normal distribution with zero mean and covariance matrix $V$:
\begin{align}    
    G&\coloneq(G_1, \dots, G_N), 
     & G_i\overset{\mathrm{i.i.d.}}{\sim} \mathcal{N}_P(0, V).
\end{align}
The Wishart distribution $\mathcal W_P(V,N)$ is the probability distribution of the $P \times P$ random matrix
\begin{equation}
    R \coloneq G G^{\top} = \sum_{i=1}^N  G_iG_i^\top \sim \mathcal W_P(V,N)\,.
    \label{eq:defWishartEnsemble}
\end{equation}
Using standard terminology in statistics, we will refer to $N$ as the number of degrees of freedom, and to $V$ as the scale matrix of the corresponding Wishart distribution.\replaced[id=rev1]{For $N\ge P$, t}{T}he probability density function of the Wishart distribution is:
\begin{equation}\label{eq:wishart1}
    \rho_{\mathcal W_P} (R;\, V,N) 
    = 
    \frac{\det(R)^{\frac{(N-P-1)}{2}} \,e^{-\frac{1}{2}\mathrm{Tr}\left(V^{-1} R\right)}}
    {2^{\frac{NP}{2}} \det(V)^{\frac{N}{2}}\Gamma_P(N/2)}\, ,
\end{equation}
where $\Gamma_P$ is the multivariate Gamma function. \added[id=rev1]{In the singular $N<P$ case, the Wishart law is still defined through Eq.~\eqref{eq:defWishartEnsemble}.}
In the special case $P = V = 1$, note that the Wishart distribution reduces to a chi-squared distribution $\chi_N^2$ with probability density:
\begin{equation}
    \rho_{\chi^2}(R;\,N) 
    = \rho_{\mathcal W_1}(R;\, 1,N) 
    = \frac{R^{\frac{N}{2}-1} \,e^{-\frac{R}{2}}}
           {2^{\frac{N}{2}}\, \Gamma (N/2)}\,.
\end{equation}
The mean and variance of a Wishart random matrix $R \sim \mathcal W_P(V/N,\,N)$ are given by
\begin{equation}
\begin{split}
    \label{eq:prop_of_wishart}
  \mathbb E (R_{\mu \nu}) &= V_{\mu\nu}\,,\\
  \mathrm {Var}(R_{\mu \nu}) &= \frac{V^2_{\mu\nu} \,+\, V_{\mu\mu}V_{\nu\nu} }{N}\,.
\end{split}
\end{equation} 
We now highlight an important property of Wishart matrices, which will be fundamental throughout this manuscript: If $R \sim \mathcal W_P(V,\,N)$ and $C$ is any fixed $S \times P$ rectangular matrix with rank $S$, then the $S\times S$ random matrix $C R C^\top$ is also Wishart distributed:
\begin{equation}
    \label{eq:wishart-law-under-congruences}
    C R C^\top \sim \mathcal W_S (C VC^\top,\, N)\,.
\end{equation}
\added[id=rev1]{This also holds in the singular $N<P$ case.} As a corollary, for $S = 1$ we get that, given a constant $P$-dimensional vector $s$, the scalar random variable $s^\top R s / s^\top V s $ is chi-squared distributed:
\begin{equation}
    \frac{s^\top R s}{s^\top V s} \sim \chi_N^2\,.
    \label{eq:propertychi}
\end{equation}

For properties of the non-central Wishart distribution, see Appendix~\ref{app:noncentral_contractions_and_tildeQ}.

\section{The Equivalent Wishart Ansatz for fully-connected DNNs in the proportional limit} \label{sec:EWA_for_MLP}

\subsubsection{Prior distribution of network outputs}
As the loss depends on the parameters $\theta$ only through the network outputs $f^\mu ,$ and thus the joint output prior $\rho(f|X)$, the partition function defined in Eq.~\eqref{defpartition} can be conveniently written in terms of the characteristic function $\varphi(\bar f|X)= \mathbb E_{f}[\exp(-if^\top \bar f)]$ of the output prior:
\begin{align}
    Z &= \int \prod_{\mu = 1}^P \frac{df^\mu d\bar f^\mu}{2\pi} e^{-\frac{\beta}{2} \sum_{\mu}\left(y^\mu -f^\mu\right)^2 + i \sum_\mu f^\mu \bar f^\mu} \varphi (\bar f| X)\nonumber \\
    &= \langle \varphi(\bar{f}|X) e^{i\sum_\mu y^\mu \bar{f}^\mu}   \rangle_{\bar f \sim \mathcal{N}_P(0,\beta \Id)}, \label{eq:Z_from_priorcharfunc}
\end{align}
where substituting $f^\mu\to f^\mu + y^\mu$ we performed the Gaussian $df$ integral, and use $\langle\cdot \rangle_{\bar f}$ only for brevity of notation of the dual variable measure. We thus need to compute the characteristic function of the joint output prior $\varphi (\bar f| X)$ by performing the integral over the parameter space:
\begin{equation}
    \varphi (\bar f| X) = \int d p(\theta) \,e^{-i \sum_{\mu} \bar f^\mu f_\theta(x^\mu)}\,.
\label{characteristicoutput}
\end{equation}
Note that the whole complexity of the learning problem lies in computing the prior or its characteristic function - specifically if we can find an integral representation 
$\varphi(\bar{f}|X) = \int dQ e^{-S_\varphi(Q,\bar{f})}$ where $S_\varphi(Q,\bar{f})$ is at most a quadratic function in $\bar{f}$, and $Q$ are a set of low-dimensional order parameters, the $d\bar{f}$ integration to obtain the posterior partition function in Eq.~\eqref{eq:Z_from_priorcharfunc} reduces to Gaussian. The EWA leads to just such an integral representation of $\varphi(\bar{f}|X)$\deleted[id=rev1]{, as shown next}.

\added[id=rev1]{For cross-entropy loss in classification tasks, the integral is no longer explicit, but an analytic expression for the prior output distribution still yields a strong simplification: In $D$-output classification, $f^\mu\in\mathbb R^D$ are the logits and we denote by $\pi_a(f^\mu)=e^{f_a^\mu}/\sum_b e^{f_b^\mu}$ the softmax probability of class $a$, giving the cross-entropy loss $\mathcal L_{\rm CE}(f,Y)=-\sum_\mu\log\pi_{y^\mu}(f^\mu)$. Correspondingly,}
\begin{equation}
    \added[id=rev1]{Z_{\rm CE}
      = \int\!df\, e^{-\beta\mathcal L_{\rm CE}(f,Y)}
        \int\!\frac{d\bar f}{(2\pi)^{PD}}\,
        e^{i\bar f^\top f}\varphi(\bar f | X)\,.}
    \label{eq:ce-partition-characteristic}
\end{equation}
\added[id=rev1]{A continuous Gaussian mixture representation of $\varphi(\bar f\mid X)$ as above directly translates to a corresponding representation of the prior $\rho(f|X)$. The remaining $f$ integral can then be evaluated by a Laplace approximation, conditional on the order parameters.}

\subsubsection{Output prior as an ensemble of random kernel matrices}

Due to the i.i.d. normal prior of the weights Eq.~\eqref{eq:weight_prior}, the activations Eq.~\eqref{eq:preactivations-definition} at layer $\ell$ in the Bayesian network prior are always a sum of Gaussian variables when conditioning on the pre-activations in the layer below. This is to make the standard observation that by introducing the empirical kernels at each layer $\ell = 1,\,\dots,\, L$:
\begin{equation}
    K^{(\ell)}_{\mathrm{E}} =   \frac{\sigma \left(H^{(\ell)}\right) \sigma \left(H^{(\ell)}\right)^\top} {\lambda N_\ell} \,,
    \label{empiricalkernel}
\end{equation}
where the activation function $\sigma$ is applied element-wise to the $P \times N_\ell$ matrix of pre-activations $H^{(\ell)},$ 
and the kernel- or Gram-matrix of the data
\begin{equation}
     K_{\mathrm{E}}^{(0)}=C=\frac{XX^\top}{\lambda N_0},
     \label{eq:def_data_Gram_matrix_C}
\end{equation}
the columns $H^{(\ell)}_{i_\ell} \coloneq (h^{(\ell)}_{1i_{\ell}}, \dots, h^{(\ell)}_{Pi_{\ell}})^\top\in \mathbb R^P$
are conditionally i.i.d. normal distributed:
\begin{equation}
\label{eq:preactivations-distribution}
    H^{(\ell)}_{1}, \dots, H^{(\ell)}_{N_\ell} \mid H^{(\ell-1)} \overset{\mathrm{i.i.d.}}{\sim} \mathcal{N}_P(0,K_{\mathrm{E}}^{(\ell-1)}).
\end{equation}
Correspondingly, also the outputs $f$ are Gaussian when conditioned on $K^{(L)}_E$. Since the distributions of $H^{(\ell)}$ each induce a distribution of $K^{(\ell)}$ in Eq.~\eqref{empiricalkernel} that again only depends on the realization of the kernel $K^{(\ell-1)}$ below, this leads to an alternative description of the prior, and its characteristic function, in terms of $L$ random matrix ensembles \cite{aitchison2020bigger}:
\begin{equation}
    \varphi (\bar f| X) 
    = \prod_{\ell = 1}^L\int_{\mathcal S^+_P} dK_{\mathrm{E}}^{(\ell)} \rho (K_{\mathrm{E}}^{(\ell)}|K_{\mathrm{E}}^{(\ell-1)}) e^{-\frac{1}{2}\bar f^\top K^{(L)}_E\bar f}\,,
    \label{eqprioralternative}
\end{equation}
where the integrations are performed over the cone $\mathcal S_P^+$ of semi-positive definite symmetric $P\times P$ matrices\added[id=rev1]{ including the boundary}, and the conditional probabilities $\rho (K_{\mathrm{E}}^{(\ell)}|K_{\mathrm{E}}^{(\ell-1)})$ are defined as: 
\begin{equation}
   \rho (K_{\mathrm{E}}^{(\ell)}|K_{\mathrm{E}}^{(\ell-1)}) 
   = \mathbb{E}\left[ \delta \bigg( 
       K_{\mathrm{E}}^{(\ell)}
       - \frac{\sigma (H^{(\ell)}) \sigma (H^{(\ell)})^\top} {\lambda N_\ell} \bigg)
       \right],
  \label{conditionaldistrib} 
\end{equation}
where the expectation is taken with respect to $H_i^{(\ell)} \mid H^{(\ell-1)} \overset{\mathrm{i.i.d.}}{\sim} \mathcal N (0, K_E^{(\ell -1)})$.
\added[id=rev1]{For the $N<P$ case, $dK_{\mathrm{E}}^{(\ell)} \rho (K_{\mathrm{E}}^{(\ell)}|K_{\mathrm{E}}^{(\ell-1)})$ is understood as integration with respect to the corresponding singular probability measure.}
Coming from a slightly different perspective, in Ref.~\cite{fischer24critical} a joint (matrix-)Gaussian approximation of these layer-wise distributions is done, leading to a high-dimensional adaptive kernel theory with $L$ coupled $P\times P$ dimensional equations of state for the layer-wise empirical kernels in the posterior. Here, we seek an approach to obtain an intensive, low-dimensional equation of state. 
The fundamental observation that will lead us to formulate the Equivalent Wishart Ansatz arises from the analysis of deep linear networks.
In this special case, $\sigma \left(H\right) \sigma \left(H\right)^\top = H H^\top $ and by Eq.~\eqref{eq:defWishartEnsemble}  the conditional probability in Eq.~\eqref{conditionaldistrib}, is non-asymptotically characterized as a Wishart distribution:
\begin{equation}
    \added[id=rev1]{
   K_{\mathrm{E,lin}}^{(\ell)}\big|K_{\mathrm{E,lin}}^{(\ell-1)} 
   \sim \mathcal W_P \left(K_{\mathrm{E}}^{(\ell-1)}/ N_\ell,\,N_\ell\right)\,. }
\end{equation}
This result is implicitly at the core of the drastic dimensional reduction that occurs in finite Bayesian deep linear networks, related to the property Eq.~\eqref{eq:propertychi} of Wishart distributions.

\subsubsection{The Equivalent Wishart Ansatz}
In the non-linear case, we cannot characterize non-asymptotically the conditional probability distributions $\rho (K_{\mathrm{E}}^{(\ell)}|K_{\mathrm{E}}^{(\ell-1)})$ layer by layer. Still we can hope that simplifications arise in the proportional regime, where $P,\, N_\ell \to \infty$ and the ratios $\alpha_\ell = P/N_\ell$ are finite. The fixed reference point is that, starting from the last layer, the mean of the distribution must be the Neural Network Gaussian Process (NNGP) kernel
\begin{equation}
    \mathbb E \left[K_E^{(L)} | K_E^{(L-1)}\right] = \Theta\big(K_E^{(L-1)}\big) \, ,\label{eq:empiricalkernel_mean_conditioned_L-1}
\end{equation}
describing the large-width limit of Bayesian DNNs. The matrix elements of the NNGP kernel are given as
\begin{equation}
    \left[\Theta(K)\right]_{\mu\nu} 
    = \frac{1}{\lambda}
      \mathbb E_h
      [
      \sigma(h^\mu)\sigma(h^\nu)
      ]
      \label{eq:def_NNGP_kernel_one-layer}
\end{equation}
with $h \sim \mathcal{N}\left(0, \Sigma_{2} \right),$ and
$
\Sigma_2 = 
      \left(\begin{smallmatrix}
                        K_{\mu\mu} & K_{\mu\nu}\\
                        K_{\nu\mu} & K_{\nu\nu}
                \end{smallmatrix} \right)
        \in \mathcal{S}_2^+.
$
The weight precision $\lambda$ appears since $\Theta(K^{(\ell)})_{\mu\nu}=\mathrm{Cov}[h_\mu^{\ell+1}h_\nu^{\ell+1}]$.

The mean in Eq.~\eqref{eq:empiricalkernel_mean_conditioned_L-1} is again a random variable when moving the conditioning down to the next layer, with mean
\added[id=rev1]{
\begin{equation}
     \mathbb E \left[\Theta\big(K_E^{(L-1)}\big)| K_E^{(L-2)}\right] =
 \Theta \circ \Theta\big(K_E^{(L-2)}\big) + O(1/N),
\end{equation}
Neglecting errors of order $O(1/N)$, we can iterate this process and get: 
}
\begin{equation}
    \mathbb E \left[ \Theta^{L-\ell}\big( K_E^{(\ell)} \big) | K_E^{(\ell-1)} \right] = \Theta^{L-(\ell-1)}\left(K_E^{(\ell-1)}\right). \label{eq:empiricalkernel_mean_conditioned_l}
\end{equation}
Here we introduced the $l$-layer NNGP kernel function 
\begin{equation}
    \Theta^\ell(K) = \Theta \circ \dots \circ \Theta (K). \label{eq:def_NNGP_kernel_lth-layer}
\end{equation}

The \emph{Equivalent Wishart Ansatz (EWA)} 
amounts to state that in the proportional regime we can perform analytical calculations \emph{as if} the fluctuations around each of these means are still Wishart.
The Ansatz is thus 
\begin{equation*}
    \Theta^{L-\ell}\big( K_E^{(\ell)} \big) \,|\, K_E^{(\ell-1)} \overset{\mathrm{EWA}}{\sim} \mathcal W_P \left(V^{(\ell)}, N_\ell\right) \, , \label{eq:EWA-statement}
\end{equation*}
with scale matrix (see Eq.~\eqref{eq:prop_of_wishart}):
\begin{equation}
    V^{(\ell)} = \frac{1}{N_\ell} \Theta^{L-(\ell-1)}\big( K_E^{(\ell-1)}\big).
    \label{eq:EWA-scalematix}
\end{equation}
For example $K_E^{(L)}|K_E^{(L-1)}\sim\mathcal W \big( \Theta\big(K_E^{(L-1)}\big)/N_L, N_L\big)$ and $\Theta^{L-1}\big(K_E^{(1)}\big) \sim \mathcal W \big( \Theta^L\big(C\big)/N_1, N_1\big)$. Note that we thereby obtained a new closed sequence of random matrix ensembles
\begin{equation}
    \rho\big(K_E^{(L)} \big) = \prod_{\ell=L}^1 \rho\left(\Theta^{L-\ell}\big( K_E^{(\ell)} \big) \,|\, \Theta^{L-(\ell-1)}\big( K_E^{(\ell-1)} \big) \right).
\end{equation}
The EWA is rooted in the idea that in first approximation the fluctuations in the kernels of a nonlinear network will be similar to those arising in a linear network. In the proportional limit in particular, for the large majority of vectors $\bar f$ the contraction $\bar f^\top K \bar f$ of these random matrices behaves as if $K$ was Wishart. This is demonstrated numerically in Section \ref{subsec:numerical-validation-ewa}.
Note that the EWA is not equivalent to a Gaussian approximation of the network’s pre-activations $h$, which would strictly yield the lazy NNGP result, nor to a Gaussian approximation of the post-activations at each layer or to a Wishart approximation of each empirical kernel. Rather, propagating fluctuations from each of the lower layer weight priors forward through the nonlinearities, induces prior fluctuations of the \textit{last layer} empirical kernel, which approximately acts as if Wishart distributed in the proportional regime.

\subsubsection{Computation of $\varphi(\bar f |X)$ given the EWA}
A direct consequence of the EWA is that in Eq.~\eqref{eqprioralternative} we can exploit the contraction property Eq.~\eqref{eq:propertychi} and form of the scale matrix Eq.~\eqref{eq:EWA-scalematix} to iteratively substitute 
\begin{align}
    e^{-\frac{1}{2} \bar{f}^\top K_E^{(L)} \bar{f}} 
    &=  \exp\bigg[  -\frac{1}{2} 
                    \underbrace{
                    \frac{\bar{f}^\top K_E^{(L)} \bar{f}}
                         {\bar{f}^\top V^{(L)} \bar{f}}\;
                         }_{\coloneq Q\sim \chi_{N_L}^2}
                    \bar f^\top V^{(L)} \bar f
            \bigg] \nonumber \\
    &= \exp\left[  -\frac{1}{2} 
                    \frac{Q_L}{N_L}
                    \bar{f}^\top \Theta(K_E^{(L-1)}) \bar{f}
            \right] \nonumber \\
    &= \exp\left[  -\frac{1}{2} 
                    \bigg(\prod_{\ell=1}^L \frac{Q_\ell}{N_\ell} \bigg)
                    \bar{f}^\top \Theta^L(C) \bar{f}
            \right] \, ,  \label{eq:iterated_chi2_intoduction}
\end{align}
where all $Q_\ell \sim \chi_{N_\ell}^2$ are independent of each other, of $\bar{f}$, and of the data Gram matrix $C$. This makes the EWA extremely powerful in reducing the complexity of the intractable MLP prior while keeping the, we argue, 
main source of fluctuations in the proportional regime.
The characteristic function is thus
\begin{equation}
    \varphi(\bar{f}|X)
    \overset{\mathrm{EWA}}{=} \int \big(\prod_\ell dQ_\ell\big)
       e^{- S_\varphi(Q, \bar{f}) + \sum_\ell \log\big(\rho_{\chi^2}(Q_\ell;\,N_\ell)\big)}  \label{eq:charfunc_prior_EWA_singleoutput}
\end{equation}
where 
$S_\varphi(Q, \bar{f}) 
 =  \frac{1}{2} 
    \big(\prod_{\ell=1}^L \frac{Q_\ell}{N_\ell} \big)
    \bar{f}^\top \Theta^L(C) \bar{f}$
is only quadratic in $\bar{f}$, as desired to compute the partition function via Eq.~\eqref{eq:Z_from_priorcharfunc}. 

In the following, we provide extensive direct and indirect evidence of the validity of the EWA in the proportional limit:
\begin{itemize}
    \item First, in Section~\ref{sec:noncentral_EWA} we show why the EWA also extends to activation functions with non-zero mean $\mathbb E_h [ \sigma(h^\mu)] > 0$, such as the ReLU activation function. While in principle requiring a generalization to non-central Wishart distributions and leading to significant complications, we show that asymptotically in the proportional limit central and non-central EWA nonetheless yield equivalent predictions, also for non-zero mean activation functions.
    \item In Section~\ref{subsec:numerical-validation-ewa} we provide a numerical verification of the validity of the EWA; we introduce a Large Deviation Principle and compare the rate function for the variables $q_\ell \coloneq Q_\ell/N_\ell$ to samples from the true MLP prior. We leverage the simplified expression for the characteristic function, Eq.~\eqref{eq:charfunc_prior_EWA_singleoutput}, to compute the posterior asymptotic free energy density in Section~\ref{subsec:effective-action-at-depth-l} and provide an interpretation in terms of data-dependent Gaussian processes. 
    \item In Section~\ref{sec:multiple_outputs_and_CNNs}, we extend this setting to DNNs with multiple outputs and introduce the Stacked Equivalent Wishart Ansatz to describe deep networks with convolutional layers.
    \item Finally, in Section~\ref{sec:Posterior_MCMC_results}, we present a systematic numerical study comparing the posterior predictor of the EWA on real data against numerical experiments with DNNs.
\end{itemize}

\subsection{EWA in the case of general activation functions.} \label{sec:noncentral_EWA}

In principle, the fact that the mean activation $m(K) = \mathbb{E}_{h\sim \mathcal{N}(0, K)}[\sigma(h)] \neq 0$ for activation functions such as ReLU suggests that $\sigma(h)$ should not be replaced by an effective zero-mean variable as supposed by the EWA above. Rather, the natural extension would be to approximate all of the $\Theta^{L-\ell}(K_E^{(\ell)}) | K_E^{(\ell-1)}$ as \emph{noncentral} Wishart random matrices. While this ensemble has a more complicated density function, one can iteratively introduce the $Q_\ell$ variables analogously to Eq.~\eqref{eq:iterated_chi2_intoduction}, only that now the $Q_\ell$ are no longer $\chi^2$-distributed. Instead, now the quantity $\tilde{Q}_{\ell}\coloneq(1 + \lambda_{\mathrm{nc}})Q_{\ell}$ follows a noncentral $\chi^2$-distribution $\tilde{Q}_{\ell} \sim \chi^2_{\mathrm{nc}}(N_\ell, N_\ell \lambda_\mathrm{nc})$ with so-called noncentrality parameter $N_\ell \lambda_\mathrm{nc}$, where 
\begin{equation}
\lambda_\mathrm{nc}(\bar{f}, K^{({\ell-1})}_E) 
= \frac{(\bar{f}^\top m)^2}
           {\bar{f}^\top \Sigma \bar{f}}. 
\end{equation}
See Appendix~\ref{app:noncentral_contractions_and_tildeQ} for the derivation.
Here $\Sigma(K) = \mathrm{Cov}[\sigma(h)]_{h\sim \mathcal N(0,K)}$ is the covariance kernel instead of the second moment, and for brevity we drop the $\ell$-dependent arguments, which play no important role in the following, writing $\Sigma,m$ and $\Theta$ instead of $\Sigma\big(\Theta^{L-\ell-1}(K^{\ell-1}_E)\big)$, $m\big(\Theta^{L-\ell-1}(K^{\ell-1}_E)\big)$ and $\Theta^{L-\ell}\big(K^{(\ell-1)}_E\big)$.

Alas, unlike for the central EWA, this distribution still depends on $\bar{f}$ and $K^{({\ell-1})}_E$, requiring to introduce an additional order parameter and a complex dependence between the $Q_{\ell}$ across layers. Indeed the distributions of $Q_\ell$ and $Q_{\ell,\mathrm{nc}}$ differ significantly, for example in their variances by a factor 
$1 - \frac{\mathrm{Var}[Q_{\ell,\mathrm{nc}}]}{\mathrm{Var}[Q_\ell]} = \frac{(\bar{f}^\top m)^4}{(\bar{f}^\top \Theta\bar{f})^2} $
which is $O(1)$ even for random overlaps $\bar{f}^\top m$. 

In the lazy infinite-width limit $N_\ell \to \infty, P=\mathrm{const.}$, these differences vanish since both distributions have the same mean and concentrate in prior and posterior to $\lim_{N_\ell \to \infty}(Q^\ast_{\ell,nc}/N_\ell) = \lim_{N_\ell \to \infty}(Q^\ast_\ell/N_\ell) = 1$, giving the usual NNGP result. The nonzero mean enters trivially in the form of the NNGP kernel, defined as the covariance of the pre-activations $h$ and therefore given through the second moment of the activations, $\Theta = \lambda^{-1}\mathbb{E}[\sigma \sigma^\top] = \lambda^{-1}(\Sigma + m m^\top)$ with $\Sigma$ the activation covariance, and has no further effects.

In the proportional limit, the fluctuations of $Q$ become important in the posterior. However, another mechanism suppresses the difference due to the central and noncentral distributions of $Q$ on the level of both prior and posterior outputs. Intuitively, the reason is that the kernel $\Theta$ has an $O(P)$ outlier eigenvalue close to the mean direction caused by the rank-1 spike $m m^\top$ which suppresses any contributions to the prior from $(\bar{f}^\top m)^2 > O(1)$ under the $d\bar f$ integral. The denominator in contrast concentrates to $\bar{f}^\top \Sigma \bar{f} \sim P$, such that only noncentrality parameters $\lambda_\mathrm{nc}(\bar{f}, K^{({\ell-1})}_E)= O(1/P)$ contribute to both prior and posterior; any contributions where $\lambda_\mathrm{nc} \neq 0$ and $\rho(Q_{\mathrm{nc}})$ would asymptotically differ from $\chi^2(N_\ell)$ are removed.

In the following we detail an argument showing this asymptotic equivalence holds at any layer, both in the prior- and the posterior output distribution.
Consider the contribution to the partition function from any single layer $l$\begin{align}
    Z(K^{(\ell-1)}_E) 
    &= \big\langle e^{i y^\top \bar{f}} \varphi(\bar{f}|K^{(\ell-1)}_E) \big\rangle_{\bar f \sim \mathcal{N}(0 , \beta \Id_P)} \label{eq:noncentra-central_Z}\\
    &= \int  d\bar{f}\, d\tilde{Q}_{\ell}\; \rho_{\chi^2_{\mathrm{nc}}}\big( \tilde{Q}_{\ell}; N_\ell, N_\ell \lambda_\mathrm{nc}(\bar{f}, K^{({\ell-1})}_E) \big) \nonumber \\
    & \qquad \quad   \times e^{- \frac{1}{2} \bar{f}^\top \big(\beta^{-1} \Id + Q_{\ell}\Theta \big) \bar{f} + i y^\top \bar{f}} \, . \nonumber
\end{align}
Note that Eq.~\eqref{eq:noncentra-central_Z} is of the form 
\begin{equation}
    \int d\tilde Q
    \left\langle F_{\tilde Q}\left( \lambda_\mathrm{nc}(\bar f) \right)
    \right\rangle_{\bar f \sim \mathcal{N}(\mu, K_\beta)} \label{eq:noncentral_function_under_fbar_measure}
\end{equation}
with $\lambda_\mathrm{nc}(\bar f) =\frac{(\bar{f}^\top m)^2}{\bar{f}^\top \Sigma \bar{f}}$, covariance $K_\beta =\big[ \beta^{-1} \Id + Q(\Sigma + m m^\top) \big]^{-1}$ and $\mu = i K_\beta\, y$. Here $F$ depends on $\bar f$ only through $\lambda_\mathrm{nc}(\bar f)$ and is a positive continuous function. 
Assuming that $\Sigma$ has an extensive number of eigenvalues above the noise floor $\beta^{-1}$, we \replaced[id=rev1]{show in App.~\ref{app:selfaveraging-noncentrality}}{now show} that $\lambda_{nc}$ is self-averaging and vanishes as $O(1/P)$ under the $\bar f$ measure, such that
\begin{equation}
    \left\langle F_{\tilde Q}\left( \lambda_\mathrm{nc}(\bar f) \right)
    \right\rangle_{\bar{f}} 
    = 
    F_{\tilde Q}\left( \langle\lambda_\mathrm{nc}(\bar f) \rangle_{\bar f} \right)
    =
    F_{\tilde Q}\left( O(1/P) \right),\label{eq:small_noncentrality}
\end{equation}
and we can replace $d\tilde Q_{\ell} \rho_{\chi^2_{\mathrm{nc}}}\big( \tilde{Q}_{\ell}\big) \to dQ_\ell\, \rho_{\chi^2}(Q_\ell)$ in Eq.~\eqref{eq:noncentra-central_Z}, recovering Eq.~\eqref{eq:charfunc_prior_EWA_singleoutput}.

These arguments can be applied layer-by-layer, first replacing $\rho_{\chi^2_{\mathrm{nc}}}\big( \tilde{Q}_{L}\big) \to \rho_{\chi^2}(Q_L)$ for $\ell = L$, then introducing $\tilde{Q}_{L-1}$ as in Eq.~\eqref{eq:iterated_chi2_intoduction} and repeating the replacement, down to $\ell=1$. This asymptotic equivalence of noncentral and central EWA not only holds at the level of the \emph{posterior} partition function Eq.~\eqref{eq:Z_from_priorcharfunc}, but also for the \emph{prior} output distribution $\rho(f|X)$, as can be seen by setting $\beta\to\infty$ in Eq.~\eqref{eq:noncentra-central_Z} such that the computation corresponds to the Fourier back-transform of the prior characteristic function.

\added[id=rev1]{For the posterior under cross-entropy loss, we restrict ourselves to noting that in many cases a similar argument still seems possible. Assume that the Laplace approximation at fixed order parameters is accurate. This corresponds to a Gaussian approximation of the logits around the mode, giving again a Gaussian dual $\bar f$ with covariance $(K+W^{-1})^{-1}$, where $K$ is the prior logit covariance and $W$ the Hessian of $\beta\mathcal L_{\rm CE}$ at the posterior mode (App.~\ref{appsec:numerical-computation-of-the-saddle-point}). The spike in $K$ again suppresses fluctuations along $m$ and the noncentrality.}

\replaced[id=rev1]{Overall, t}{T}he more complicated noncentral EWA therefore reduces to the zero-mean Ansatz, and is not considered further. Only at $P \approx 50$ we could find small differences between noncentral and zero-mean theories as a finite-size effect (not shown). For additional numerical evidence and discussion of the mean contribution, see Appendix~\ref{app:noncentral_EWA}.

\subsection{Numerical validation of the EWA}
\label{subsec:numerical-validation-ewa}

\begin{figure}
    \centering
    \includegraphics[width=.45\textwidth]{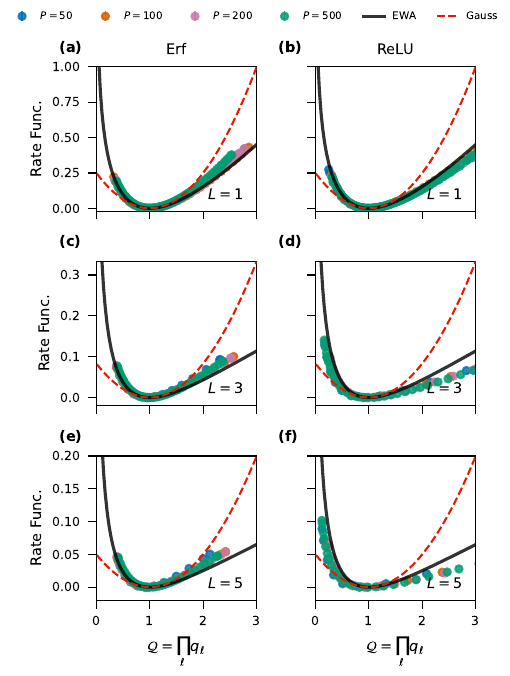}
    \caption{\textbf{Rate function for the product random variable $\mathcal{Q}$.} Numerical estimates of the network rate function for the CIFAR-10 dataset (coloured dots) are compared to the expected theoretical rate function under the EWA \added[id=rev1]{(black solid lines) and the Gaussian approximation (red dashed lines)}, both for Erf (first column) and ReLU activation function (second column). \added[id=rev1]{The empirical rate function is obtained through thermodynamic integration along a ladder of tilted distributions. The procedure is repeated} at a fixed value of $\alpha=1.0$ but for increasing value of $P\in\{50,100,200,500\}$, necessary condition to see the asymptotic convergence and therefore to assess the validity of the LDP.}
    \label{fig:product-rate-function}
\end{figure}

The aim of this section is to provide numerical evidence for the EWA using the framework of Large Deviation Theory.
As discussed in the previous section, possible contributions due to a non-zero mean of the activation function are negligible in the asymptotic regime. Therefore, it is sufficient to investigate whether a Large Deviation Principle (LDP) holds for the central EWA, which states that:
\begin{equation}
    \Theta^{L-\ell}(K_\mathrm{E}^{(\ell)}) \mid K_\mathrm{E}^{(\ell-1)} \sim \mathcal{W}_P
    \left(
        \frac{\Theta^{L-\ell+1}(K_\mathrm{E}^{(\ell-1)})}{N_\ell}, N_\ell
    \right),
\end{equation}
where $\Theta$ is the usual NNGP kernel function.
Assuming the validity of the EWA implies that property in Eq.~\eqref{eq:propertychi} needs to be verified, \textit{i.e.} that, at each layer, the variables $Q_\ell$ defined as
\begin{equation}
    Q_\ell = N_\ell \frac{\bar f^\top \Theta^{L-\ell}(K_\mathrm{E}^{(\ell)}) \bar f}{\bar f^\top \Theta^{L-\ell+1}(K_\mathrm{E}^{(\ell-1)}) \bar f}
\end{equation}
are chi-squared distributed for every fixed realization of the $\bar f \in \mathbb{R}^P$ (conditioned on the empirical kernel at the previous layer):
\begin{equation}
    Q_\ell \mid K_\mathrm{E}^{(\ell-1)} \sim \chi^2_{N_\ell} = \Gamma(N_\ell/2, 2).
\end{equation}
Notice that the distribution of $Q_\ell \mid K_\mathrm{E}^{(\ell-1)}$ actually does not depend on $K_\mathrm{E}^{(\ell-1)}$, implying that the $Q_\ell$ variables are decoupled across layers, and it is also independent on $\bar{f}$. Since $\mathbb{E}[Q_\ell] = N_\ell$, let us define the corresponding intensive variables $q_\ell = Q_\ell / N_\ell \sim \Gamma(N_\ell/2, 2/N_\ell)$, which concentrates around $q_\ell=1$ in the infinite-width limit.
The intensive variables $q_\ell$, each viewed as a sequence over the corresponding $N_\ell$, satisfy a LDP of the form:
\begin{align}
    \rho_{q_\ell}(x) &\sim e^{-a_{N_\ell} \mathcal{I}_\ell(x)} & &\text{as } N_\ell \to \infty.
\end{align}
$\mathcal{I}_\ell(x)$ is called rate function, and it is calculated via the Gärtner-Ellis theorem as the Legendre transform of the scaled cumulant generating function $\Lambda_{q_\ell}$:
\begin{align}
    \mathcal{I}_{q_\ell}(x) &= \sup_t \{ tx - \Lambda_{q_\ell}(t) \}, \\
    \Lambda_{q_\ell}(t) &\coloneq \lim_{N_\ell \to \infty} \frac{1}{a_{N_\ell}} \ln M_{q_\ell}(a_{N_\ell} t).
\end{align}
Using for $M_{q_\ell}$ the moment generating function of a $\Gamma(N_\ell/2, 2/N_\ell)$ distribution, it follows that the scales for these LDPs are $a_{N_\ell}=N_\ell$ and that the rate function is the same for all layers:
\begin{equation}
\label{eq:single-q-rate-func}
    \mathcal{I}_{q_\ell}(x) = \mathcal{I}_q(x) = \frac{1}{2} (x-1-\ln x).
\end{equation}
If we take $N_\ell=N \, \forall l$, the random variables $q_\ell$ are all independent and identically distributed, and we can derive a multivariate LDP with a single scale $a_N=N$ (otherwise a multi-scale LDP should be considered):
\begin{align}
\label{eq:LDP-multivariate-q}
    \rho_{(q_1,\dots,q_L)}(x_1, \dots, x_L) &\sim e^{-N \sum_{\ell=1}^L \mathcal{I}_q(x_\ell)} & &\text{as }N\to\infty.
\end{align}
\newtext{We note that, although the EWA directly implies the contraction property, Eq.~\eqref{eq:propertychi}, the approximation may be more or less accurate depending on the vector $\bar f$ considered. In some cases, this may result in leading-order corrections to the rate function of $q_\ell$ (an explicit example is discussed in App.~\ref{app:equicorrelated})}.

As shown in the previous section, see in particular Eq.~\eqref{eq:charfunc_prior_EWA_singleoutput}, the characteristic function of the prior over the network output actually depends only on the product $\mathcal{Q}\coloneq\prod_{\ell=1}^L q_\ell$. We can derive the rate function for the random variable $\mathcal{Q}$ using the contraction principle:
\begin{equation}
\begin{aligned}
\label{eq:product-rate-function}
    \mathcal{I}_{\mathcal{Q}}(y) &= \inf \left\{ \sum_{\ell=1}^L \mathcal{I}_q(x_\ell) : \prod_{\ell=1}^L x_\ell = y \right\} \\
    &= L \, \mathcal{I}_q(y^{1/L}).
\end{aligned}
\end{equation}

\begin{figure}
    \centering
    \includegraphics[width=.45\textwidth]{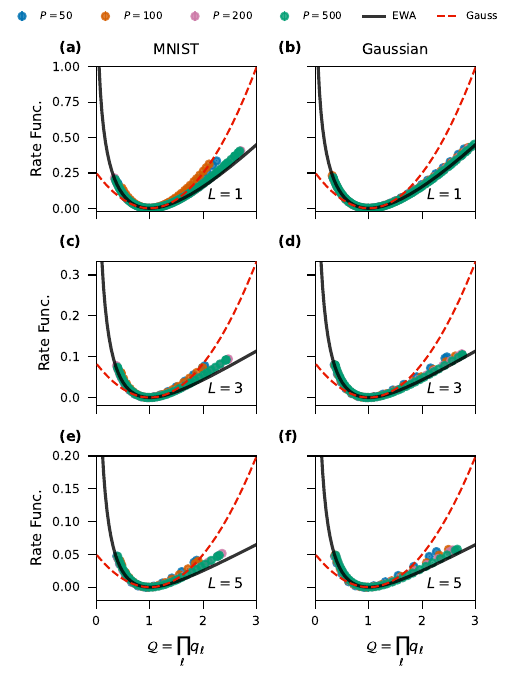}
    \caption{\textbf{Large deviation principle \added[id=rev1]{for different datasets and Erf activations}.} Numerical samples of the empirical rate function (coloured dots) for the MNIST dataset (first column) and on Random Gaussian data (second column) are compared to the expected theoretical rate function for the $\mathcal Q$ variable under the EWA (black lines) and its Gaussian approximation (red dashed lines). \added[id=rev1]{The empirical rate function is obtained through thermodynamic integration along a ladder of tilted distributions. The entire procedure is repeated} for increasing value of $P\in\{50,100,200,500\}$ in order to assess the asymptotic convergence of the empirical rate function to the theoretical one.}
    \label{fig:rate_function_TI_dataset}
\end{figure}

\added[id=rev1]{Fig.~\ref{fig:product-rate-function} shows the rate function for the $\mathcal{Q}$ variable for both Erf and ReLU activation functions and the CIFAR-10 dataset, for several networks of different depth. Black solid lines denote the theoretical rate function obtained under the EWA, Eq.~\eqref{eq:product-rate-function}, while the red dashed lines represent its Gaussian approximation, related to the CLT associated with the EWA variance. The colored points correspond to statistical estimates of the actual network rate function, obtained by performing thermodynamic integration along a ladder of polynomially tilted distributions. The tilting method was employed in order to sample more efficiently further in the tails of the prior distribution, where the number of prior samples needed to estimate the rate function becomes exponentially prohibitive. For an overview of the method, including details on the numerical implementation, we refer the reader to App.~\ref{app:tilting_and_thermointegration}. Statistical error bars calculated from the integrated autocorrelation time of each chain and then propagated through the thermodynamic integration, are smaller than the symbols on the scale of the figure. In Fig.~\ref{fig:rate_function_TI_dataset} we show the analog results for different datasets, and similar conclusions can be drawn as for Fig.~\ref{fig:product-rate-function} both for the MNIST dataset (first column) and for Gaussian data (second columns). In all panels the measured rate function is manifestly asymmetric, as predicted by the EWA: it rises steeply for $\mathcal Q<1$ and grows sub-quadratically for $\mathcal Q>1$, and the data lie far from the Gaussian parabola on both branches. A central-limit description of the kernel fluctuations is therefore inadequate for values of $\mathcal Q$ further in the tails of the distribution, potentially in the range relevant for the posterior, while the EWA captures the leading non-Gaussian behavior. 
The agreement is nonetheless not exact, since the EWA is exact only for deep linear networks. The residual discrepancies grow with depth and depend on the activation function. For Erf networks they are small: at $L=3$ and $L=5$ on MNIST and CIFAR-10 the left branch is predicted correctly while the right branch of the data lies slightly above the EWA curve, meaning that large values of $\mathcal Q$ are less likely than predicted, and on Gaussian input data the agreement is nearly quantitative. For ReLU networks the deviations are larger and of opposite sign: at $L=3$ and $L=5$ the measured rate function lies below the EWA prediction on both branches.}

\deleted[id=rev1]{It is important to underline that an independent sampling of the $q_\ell$ variables is required in order to build the proper $\mathcal{Q}$ variable satisfying the LDP with rate function in Eq.~\eqref{eq:product-rate-function}, in addition to keeping the $\bar f$ vector fixed across samples. About the sampling procedure, the independent samples of the vectors $(q_1,\dots,q_L)$ are obtained by sampling Gaussian preactivations, according to Eq.~\eqref{eq:preactivations-distribution}, and building the corresponding empirical kernels at each layer. The NNGP kernel function $\Theta$ is then applied recursively to build the $q_\ell$ variables at each layer, and this entire procedure is repeated for each sample keeping the same $\bar f$. At the end, the $\mathcal{Q}=\prod_\ell q_\ell$ variables (one for each network) are calculated from the independent samples.}

For more details about the large deviation analysis, we refer the reader to App.~\ref{appsec:rate-function-prior}.

\subsection{Effective action for MLPs of depth $L$}
\label{subsec:effective-action-at-depth-l}

Using the EWA expression for the prior characteristic function, Eq.~\eqref{eq:charfunc_prior_EWA_singleoutput}, a Gaussian integration gives the partition function Eq.~\eqref{eq:Z_from_priorcharfunc} as the following expectation value over the $\{q_\ell\}_\ell$ variables:
\begin{equation}
\label{eq:Z-as-expectation-for-Varadhan}
    Z = \mathbb{E}_{q_1,\dots,q_L} \left[e^{-\frac{N}{2} U(q_1,\dots,q_L)}\right],
\end{equation}
where $U(q) \coloneq U(q_1, \dots, q_L)$ is given by:
\begin{equation}
\begin{aligned}
    U(q) &= \frac{\alpha}{P} \log \det \left[\Id + \beta K^{(\text{R})}_{\mathcal{Q}}\right] + \\
    & \quad \quad \quad + \frac{\alpha}{P} y^\top \left[\beta^{-1}\Id + K^{(\text{R})}_{\mathcal{Q}}\right]^{-1}y,
\end{aligned}
\end{equation}
and the renormalized kernel is:
\begin{align}
\label{eq:renormalized_kernel}
    K^{(\text{R})}_{\mathcal{Q}} &= \mathcal{Q} \Theta^L(C), & \mathcal{Q}&=\prod_{\ell=1}^L q_\ell. 
\end{align}

Where the $\{q_\ell\}_\ell$ satisfy the joint LDP given by Eq.~\eqref{eq:LDP-multivariate-q}. \added[id=rev1]{Because $U(q)$ is always non-negative, the moment condition for Varadhan's Lemma (see, e.g., Ref. \cite{dembo2010large}) is automatically satisfied. This allows us to apply the lemma directly to calculate the asymptotic free energy density of the system \added[id=rev1]{under the EWA}:}
\begin{equation}
\label{eq:min_of_effective_action}
    - \frac{1}{N} \ln Z \to \inf_{q_1, \dots, q_L} \underbrace{\left[\sum_{\ell=1}^L \mathcal{I}_q(q_\ell) + \frac{1}{2} U(q_1, \dots, q_\ell)\right]}_{S(q)/2}.
\end{equation}
\deleted[id=rev1]{We notice that, in order to make this step we need to make the technical assumption that the scaling of the two terms in $U$ are well behaved in the proportional limit, when $P\to\infty$, and in particular that the limit exists and it is finite.}
\added[id=rev1]{
We notice that, in order for the large deviation analysis to be valid, we need to assume a property of the sequence of subsets of the datasets: $U$ depends on $P$ through $\Theta^L(C)$ and $y$, and we assume that, at fixed $\alpha$, the two terms in $U(q)$ converge for $P\to\infty$ to continuous functions of $\mathcal Q$. The same assumption on the data is needed in the deep linear case, where the EWA is exact, so this is a condition on the dataset sequence rather than on the ansatz. Eq. \eqref{eq:min_of_effective_action} is thus exact for the EWA surrogate; whether it describes the original network rests on the EWA itself, which we have tested in Sec. \ref{subsec:numerical-validation-ewa}.
}
We also point out that the application of Varadhan's Lemma and of LDT in general is a more powerful method than the calculation of $Z$ through saddle point/Laplace approximation, which would formally yield the same result, because it does not make assumptions of the convergence properties of the partition function itself, but directly provide\added[id=rev1]{s} the asymptotic free energy density, which is, after all, the only quantity we are actually interested in. Indeed, there exist sequences of random variables for which no density--or no useful asymptotic density--is available, yet they satisfy a large deviation principle at the level of measures, and this is sufficient to Varadhan’s lemma to be applicable. In this context, it is indeed very hard to make mathematically rigorous statements about the partition function itself in the proportional limit.
Note that in Eq.~\eqref{eq:min_of_effective_action} we introduce a new quantity $S(q)$, called effective action in the physics language, whose minimum is \added[id=rev1]{(twice)} the free energy density of the system and which is given by (up to irrelevant additive constants):
\begin{equation}
\begin{aligned}
\label{eq:action_singleoutput}
    S(q) = \sum_{\ell=1}^L [q_\ell - \log q_\ell] & + \frac{\alpha}{P} \log \det \left[\Id + \beta K^{(\text{R})}_{\mathcal{Q}}\right] + \\
    & + \frac{\alpha}{P} y^\top \left[\beta^{-1}\Id + K^{(\text{R})}_{\mathcal{Q}}\right]^{-1}y.
\end{aligned}
\end{equation}
We notice that the obtained free energy, corresponding to the marginal log-likelihood in Bayesian language, is that of a hierarchical Gaussian process regression model, where the kernel is being renormalized by a function of the order parameters $\{q_\ell\}_\ell$, which are in principle distributed according to their own hyper-prior but that concentrates around some specific data-dependent values in the proportional limit (for a detailed discussion in the 1HL case, we refer the reader to Ref.~\cite{pacelli2024GP}). Indeed, in the effective action, the first term corresponds to the order-parameter hyper-prior (and physically to the entropic contribution in the free energy), while the function $U(q_1, \dots, q_\ell)$ is exactly the marginal log-likelihood of a Gaussian process regression model, conditioned on the value of the hyperparameters $\{q_\ell\}_\ell$. This implies the possibility to do exact inference with this theoretical model, and in particular to find a closed form expression for the posterior predictive distribution and therefore for relevant observables like the generalization error, see for example Ref. \cite{williams2006gaussian}. \replaced[id=rev1]{Notice that for ReLU activations the effective action Eq.~\eqref{eq:action_singleoutput} leads to the same heuristic theory proposed in \cite{SompolinskyLinear}, which was obtained by replacing the linear NNGP kernel with the ReLU one.}{, based on the homogeneity property common to linear and ReLU activations}

\added[id=rev1]{For cross-entropy loss, Eq.~\eqref{eq:Z-as-expectation-for-Varadhan} has the same form. The likelihood contribution becomes}
\begin{equation}
    \added[id=rev1]{U_{\rm CE}(q) = -\frac{2}{N}\log
    \mathbb{E}_{f\sim\mathcal N(0,K^{(\mathrm R)}_{\mathcal Q})}
    \!\left[\prod_{\mu=1}^P \pi_{y^\mu}(f^\mu)^\beta\right]\,,}
    \label{eq:ce-conditional-evidence}
\end{equation}
\added[id=rev1]{In contrast to the network weight posterior, the logit posterior is typically unimodal. We can therefore in practice evaluate this expectation by a Laplace approximation about the most probable logits at fixed $\mathcal Q$ (see App.~\ref{appsec:numerical-computation-of-the-saddle-point}).}

As discussed in Sec.~\ref{subsec:numerical-validation-ewa}, the $\{q_\ell\}_\ell$ variables have \added[id=rev1]{prior} expectation value $\mathbb{E}[q_\ell]=1$ and variance $\mathrm{Var}[q_\ell]=2N_\ell^{-1}$, such that they concentrate in the proportional limit. The action Eq.~\eqref{eq:action_singleoutput} governing the posterior however, is minimized in general by values of $q^\ast_\ell \ne 1$, thus far in the tail of the prior distribution. This arises from the fact that the likelihood (loss term) scales as $\sim P$, adding a constraint for each data sample.  The dominant contribution which a good approximation of the network prior must capture is therefore not the mode of the distribution, but its large deviations behavior.

\subsection{Theory for predictor statistics}

\replaced[id=rev1]{Given the saddle-point of the effective action, t}{T}he theoretical predictions\replaced[id=rev1]{ on train- and test-points are then}{, on the other hand, are}straight-forward to evaluate. Since the characteristic function of the prior in Eq.~\eqref{eq:charfunc_prior_EWA_singleoutput} is a Gaussian mixture in the variables $\bar{f}$ and the mixture weight concentrates on $q_1^*, \dots, q_L^*$ in the posterior, the predictor statistics given squared-error likelihood Eq.~\eqref{eq:loss} are those of a Gaussian process regressor. Under these circumstances, analytical expressions for the train and test losses can be obtained by means of the bias--variance decomposition:
\begin{equation}
    \epsilon_{\text{g/t}}(x,y) = \left(y-\Gamma(x)\right)^2 + \sigma^2(x)\, ,
\end{equation}
where the bias and variance of the network outputs are those of a Gaussian process regressor \cite{Neal1995BayesianLF} with the large-width NNGP kernel replaced by the renormalized kernel evaluated at the minimum of the effective action (see Eq.~\eqref{eq:min_of_effective_action}). Details regarding the numerical implementation employed to obtain the minimum $q_1^*, \dots, q_L^*$ are provided in App.~\ref{appsec:numerical-computation-of-the-saddle-point}.
Thus, for a generic input $x$ the bias and variance read:
\begin{align}
    \label{eq:bias-definition}
    \Gamma(x) = & \left[ K^{(\text{R})}_{\mathcal{Q}^*}(Xx) \right]^T \left[ \frac{\Id}{\beta} + K^{(\text{R})}_{\mathcal{Q}^*}(C)\right]^{-1} Y \, , \\
    \label{eq:sigma2-definition}
    \sigma^2(x) = & \, K^{(\text{R})}_{\mathcal{Q}^*}(x^Tx)\ - \left[ K^{(\text{R})}_{\mathcal{Q}^*}(Xx) \right]^T \nonumber \\
    & \times \left[ \frac{\Id}{\beta} + K^{(\text{R})}_{\mathcal{Q}^*}(C) \right]^{-1} \left[ K_{\mathcal{Q}^*}^{R}(Xx) \right]  \, ,
\end{align}
where $Xx \in \mathbb R^{P}$, $x^Tx \in \mathbb{R}$, $\mathcal{Q}^*=\prod_{\ell=1}^{L}q_\ell^*$ and 
\begin{equation*}
    K^{(\text{R})}_{\mathcal{Q}^*}(Xx) = \mathcal{Q}^* \Theta^L(Xx)\, , \quad
    K^{(\text{R})}_{\mathcal{Q}^*}(x^Tx) = \mathcal{Q}^* \Theta^L(x^Tx)\, .
\end{equation*}
In the equations above, $\Theta^L(Xx) \in \mathbb{R}^P$ and $\Theta^L(x^Tx) \in \mathbb{R}$ denote in analogy with Eq.~\eqref{eq:def_NNGP_kernel_lth-layer} the $L$-th composition of the NNGP kernel function, acting respectively on a vector and on a scalar entry; specifically, 
\begin{align}
    [ \Theta(Xx) ]_\mu & = \mathbb{E}_{h^\mu, h}\left[ \sigma(h^\mu)\sigma(h)\right] / \lambda  \nonumber \\
    \Theta(x^Tx) & = \mathbb{E}_{h}\left[ \sigma(h)\sigma(h)\right]/\lambda
\end{align}
where $(h^\mu, h)\sim\mathcal{N}(0,\Sigma_\mu)$ and $h\sim\mathcal{N}(0,\Sigma_s)$, being $
\Sigma_\mu = \frac{1}{N_0 \lambda}
      \left(\begin{smallmatrix}
                        (x^\mu)^T x^\mu & (x^\mu)^T x\\
                        x^T x^\mu & x^Tx
                \end{smallmatrix} \right)
$ and $\Sigma_s = \frac{x^Tx}{N_0 \lambda}$. It is worth noting that in the limit $\alpha \to 0$, the saddle-point solutions reduce to $q_1^* = \ldots = q_L^* = 1$, and Eqs.~\eqref{eq:bias-definition} and \eqref{eq:sigma2-definition} for the bias and variance reproduce the large-width asymptotics \cite{LeeGaussian}, as expected.

\section{Kernel Renormalization schemes for deep architectures with Multiple Outputs and Convolutional Layers} \label{sec:multiple_outputs_and_CNNs}

\subsection{Fully-connected DNNs with multiple outputs}
\label{subsec:multiple-outputs}

The multiple outputs architecture implements a function $f_\theta : \mathbb{R}^{N_0} \to \mathbb{R}^D$, with $D>1$. The last-layer weights thus form a $D \times N_L$ matrix, whose elements are labeled by an additional index, leading to the following definition of the network function:
\begin{equation}
    f_{a}(x) = \frac{1}{\sqrt{N_L}} \sum_{i_L=1}^{N_L}W^{(L+1)}_{a i_L} \sigma\Bigl(h^{(L)}_{i_L}\Bigl) \, ,
\end{equation}
where the pre-activations $h^{(L)}_{i_L}$ are defined in Eq.~\eqref{eq:preactivations-definition}. The square-loss function is the natural generalization of Eq.~\eqref{eq:loss}, replacing the square with the squared Frobenius norm. 
Let us start from the 1HL case. The EWA amounts to assume that the characteristic function of the prior over outputs is given by:
\begin{equation}
\begin{split}
    \varphi(\bar{f}|X) &= \int_{S_P^+} dp(K_\mathrm{E}) e^{-\frac{1}{2} \Tr (\bar{f} K_\mathrm{E} \bar{f}^\top)}, \\
    K_\mathrm{E} \mid C &\sim \mathcal{W}_P(\Theta(C)/N_1, N_1).
\end{split}
\end{equation}
In the last equation, $\bar{f} \in \mathbb{R}^{D \times P} $ denotes the matrix whose elements are the dual variables $\bar{f}_a^\mu $. To proceed with the calculation, we will need two mathematical identities:
\begin{itemize}
    \item Ingham-Siegel integral \cite{ingham_1933, siegel_1935}, or Laplace transform of a Wishart distribution: \added[id=rev1]{ For $\alpha>0,\;A\succ0$,} 
    \newline
    \begin{equation}
    \label{eq:Ingham-Siegel} 
         \added[id=rev1]{\mathbb E_{M\sim\mathcal{W}_P(V,N)}\big[}e^{-\frac{\alpha}{2} \Tr(AM)}\added[id=rev1]{\big]} =
        [\det(\Id_P + \alpha VA)]^{-N/2} \,.
    \end{equation}
    \item Weinstein–Aronszajn identity \cite{pozrikidis2014introduction}: given $A$ and $B$ matrices of size $m \times n$ and $n \times m$ respectively, the following identity holds
    \begin{equation}
    \label{eq:Weinstein-Aronszajn-identity}
        \det (\Id_m + A \cdot B) = \det (\Id_n + B \cdot A). 
    \end{equation}
\end{itemize}
After using the cyclic property of the trace operation, we can proceed with the calculation as
\begin{equation}
\begin{split}
    \varphi(\bar{f}|X) &= \left[
        \det \left(
            \Id_P + \frac{1}{N_1} \Theta(C) \bar{f}^\top \bar{f}
        \right)
    \right]^{-N_1/2} \\
    &= \left[
        \det \left(
            \Id_D + \frac{1}{N_1}\bar{f} \Theta(C) \bar{f}^\top
        \right)
    \right]^{-N_1/2} \\
    &= \int_{S_D^+} dp(q) e^{-\frac{1}{2} \Tr (q \bar{f} \Theta(C) \bar{f}^\top)} \\
    &= \int_{S_D^+} dp(q) e^{-\frac{1}{2} \bar{f} (q \otimes \Theta(C)) \bar{f}^\top},
\end{split}
\end{equation}
where $q$ is a Wishart distributed random matrix, $q\sim\mathcal{W}_D(\Id_D/N_1, N_1)$.
To understand how to generalize this result to the deep case, let us focus to a 2HL architecture, for which the EWA reads (recall that $K_\mathrm{E}^{(0)} = C$):

\begin{equation}
\label{eq:2HL-EWA-multiple-output}
\begin{split}
    K_E^{(2)} \mid K_E^{(1)} &\sim \mathcal{W}_P\left(\Theta(K_E^{(1)})/N_2, N_2\right), \\
    \Theta(K_E^{(1)}) \mid K_E^{(0)} &\sim \mathcal{W}_P\left(\Theta^2(C)/N_1, N_1\right) .
\end{split}    
\end{equation}
The characteristic function of the prior over outputs is in this case given by:
\begin{equation}
    \varphi(\bar{f}| X) = \int dp(H^{(1)}) \int_{S_P^+} dp(K_\mathrm{E}^{(2)} \mid H^{(1)}) \, e^{-\frac{1}{2} \Tr (\bar{f} K_\mathrm{E}^{(2)} \bar{f}^\top)},
\end{equation}
where in the distribution of $K_\mathrm{E}^{(2)} \mid H^{(1)}$ the dependence on the pre-activations $H^{(1)}$ is only through the empirical kernel $K_\mathrm{E}^{(1)}$, allowing for a simple change of the integration variable to $K_\mathrm{E}^{(1)}$. The inner integral has the same structure as in the 1HL case, leading to:
\begin{equation}
\begin{split}
    \varphi(\bar{f}|X) = \int_{S_P^+} d p(K_\mathrm{E}^{(1)}) \int_{S_D^+} dp(q_2) e^{
        -\frac{1}{2} \Tr (q_2 \bar{f} \Theta(K_E^{(1)}) \bar{f}^\top)
    } \, , 
\end{split}
\end{equation}
where $q_2 \sim \mathcal{W}_D\left(\Id_D / N_2, N_2 \right)$. Given that $q_2$ is a Wishart matrix, it is symmetric, positive-definite. Therefore we can decompose it (using for example the Cholesky decomposition) as $q_2=U_2U_2^\top$.
In this way, we can write the exponent as
\begin{equation}
    \Tr (q_2 \bar{f} \Theta(K_E^{(1)}) \bar{f}^\top) = \Tr ( \underbrace{U_2^\top \bar{f}}_{g} \Theta(K_E^{(1)}) \underbrace{\bar{f}^\top U_2}_{g^\top}) \, .
\end{equation}
At this point, we can apply directly the 2HL EWA, Eq.~\eqref{eq:2HL-EWA-multiple-output}, obtaining the same integral expression as in the last-layer (up to a replacement $\bar f \to g\coloneq U_2^\top \bar f$), leading to:
\begin{equation}
\begin{split}
    \varphi(\bar{f}|X) & = \int_{(S_D^+)^2} dp(q_1, q_2) e^{
        -\frac{1}{2} \bar{f} (\mathcal{Q} \otimes \Theta^2(C)) \bar{f}^T}, \\
    \mathcal{Q} &\coloneq U_2U_1U_1^TU_2^T, \\
    q_\ell & \sim \mathcal{W}_D\left(\Id_D / N_\ell, N_\ell \right), \quad U_\ell U_\ell^T = q_\ell.
\end{split}
\end{equation}
It is now clear that the same procedure can be carried on also for deeper networks, leading to the following kernel renormalization scheme:
\begin{align}
    K^{(\text{R})}_{\mathcal Q} &= \mathcal{Q} \otimes \Theta^L(C), \label{eq:multiple-output-kernelrenorm}\\
    \mathcal{Q} &= \left( \prod_{\ell=1}^{L}U_{\ell}^\top \right)^\top\left( \prod_{\ell=1}^{L}U_{\ell}^\top \right) \, .
\end{align}
\added[id=rev1]{The effective action and its optimization are described in App.~\ref{appsec:numerical-computation-of-the-saddle-point}, and we compare the predictions with posterior sampling in Sec.~\ref{subsec:learning-curves-multiple-outputs-classification}.}
Notice that, in the linear case, we recover the \deleted[id=rev1]{same} result of Refs. \cite{Bassetti:JMLR:2024,SompolinskyLinear}.

\subsection{The \deleted[id=rev1]{stacked} Equivalent Wishart Ansatz for DNNs with convolutional layers}
\label{subsec:the-stacked-equivalent-wishart-ansatz-for-dnns-with-convolutional-layers}
\newtext{
One additional advantage of the EWA formalism is that it can be used to obtain, for the first time, the kernel renormalization scheme for non-linear deep neural networks with convolutional layers (CNNs). Here we analyze the general case in which each convolutional layer has an arbitrary stride $S_\ell$ and filter masks $\mathcal M_\ell$ of integer size $M_\ell=|\mathcal M_\ell|$, the filter index can take $M_\ell$ corresponding integer values $m_\ell \in \mathcal M_\ell$. The input data carry the usual positional index $i_0 = 1, \ldots, p_0$ and also display different channels, i.e. each input has the form $x^\mu_{i_0 a_0}$, where $a_0 = 1, \ldots, A_0$ (with $A_0$ being the total number of input channels). At a generic layer $\ell$, the network has $p_\ell$ spatial positions (patches) and $A_\ell$ channels, labelled by the indices $i_\ell$ and $a_\ell$, respectively. To treat different convolutional geometries in a unified way, we denote by $i\oplus_\ell m$ the position in layer $\ell-1$ that is read by position $i$ of layer $\ell$ through the filter offset $m$ (for a one-dimensional convolution $i\oplus_\ell m=S_\ell\,i+m$). In the first layer, convolutional pre-activations are defined as
\begin{equation}
    h_{i_{1} a_{1}}^{(1)^\mu}=\frac{1}{\sqrt{M_1 A_{0}}}\sum_{a_{0}=1}^{A_{0}}\sum_{m\in\mathcal M_1}W_{a_{1}a_{0}m}^{(1)}x_{i_1\oplus_1 m, a_{0}}^{\mu}\,,
\end{equation}
while at generic depth $\ell$ pre-activations read:
\begin{equation}
    h_{i_{\ell} a_{\ell}}^{(\ell)^\mu}=\frac{\sum_{a_{\ell-1}=1}^{A_{\ell-1}}\sum_{m\in\mathcal M_\ell}W_{a_{\ell}a_{\ell-1}m}^{(\ell)}\sigma\Bigl(h^{(\ell-1)^\mu}_{i_\ell \oplus_\ell m, a_{\ell-1}}\Bigr)} {\sqrt{M_\ell A_{\ell-1}}}\,,
\end{equation}
In this context, the readout layer performs a flatten and fully-connected dense readout, so that a deep CNN implements the following function:
\begin{equation}
    f_{\theta}(x^{\mu})=\frac{{1}}{\sqrt{{A_{L}{p_L}}}}\sum_{a_{L}=1}^{A_{L}}\sum_{i_{L}=1}^{{p_L}}W_{a_{L}i_{L}}^{(L+1)}\sigma\left(h_{i_{L}a_{L}}^{(L)^{\mu}}\right)\, .
\end{equation}
Since the readout weights enter linearly, they can be integrated out exactly, giving
\begin{equation}
    \varphi(\bar f\mid X)=
    \mathbb E_{K_\text{E}^{(L)}} e^{-\frac{1}{2p_L}\Tr \bar{F}_L^\top K_\text{E}^{(L)} \bar{F}_L}, \quad \bar{F}_L = \Id_{p_L}\otimes\bar{f} \, ,
    \label{eq:characteristic-function-prior-cnn}
\end{equation}
where $K_\text{E}^{(\ell)}\in \mathbb{R}^{Pp_\ell\times Pp_\ell}$ is the empirical kernel of layer $\ell$, with entries:
\begin{equation}
    \left(K_\text{E}^{(\ell)}\right)_{i_\ell j_\ell}^{\mu\nu}=A_\ell^{-1}\sum_{a_\ell} \sigma\left( h_{i_\ell a_\ell}^{(\ell)^\mu}\right) \sigma \left(h_{ j_\ell a_\ell}^{(\ell)^\nu}\right)\, .
\end{equation}
Note that in Eq.~\eqref{eq:characteristic-function-prior-cnn} and in what follows we set $\lambda_\ell=1$ to lighten the notation. The dimensional reduction rests on a single backward integration identity, which we use repeatedly. Suppose that $K_E^{(L)}$ is a Wishart kernel of scale matrix $V_L/A_L$ and $A_L$ degrees of freedom, and let $S$ be a positive semidefinite matrix weighting its $p_L$ spatial coordinates. Then the transform of Eq.~\eqref{eq:characteristic-function-prior-cnn} reads
\begin{align}
    \label{eq:cnn-backward-transform}
    \mathbb E_{K_\text{E}^{(L)}} e^{-\frac{1}{2p_L}\Tr[S\,\bar{F}_L^\top K_\text{E}^{(L)} \bar{F}_L]}  = & \mathbb E_{q} e^{-\frac{1}{2p_L}\Tr[q \bar{F}_L^\top V_L \bar{F}_L]},\\
    q\sim&\mathcal W_{p_L}(S/A_L,A_L)\,.\nonumber
\end{align}
This follows by applying Eq.~\eqref{eq:Ingham-Siegel}, then Eq.~\eqref{eq:Weinstein-Aronszajn-identity}, and finally again Eq.~\eqref{eq:Ingham-Siegel} once more, as an inverse transform. For one hidden layer, the pre-activation covariance is the Gram matrix of input patches, normalized by the number of filter offsets and input channels:
\begin{equation}
    \omega_1(C)^{\mu\nu}_{i_1 i_1'}=\frac{1}{M_1}\sum_{m\in\mathcal M_1} C^{\mu\nu}_{i\oplus_1 m, i'\oplus_1 m},
\end{equation}
where 
\begin{equation}
    \quad C^{\mu\nu}_{i_0 i_0'}=\frac{1}{A_0}\sum_{a_0}x^\mu_{i_0 a_0} x^\nu_{i_0' a_0} \, .
\end{equation}
In this case the scale matrix of the empirical kernel $K_{\text{E}}^{(1)}$ is therefore $V_1 / A_1 = \Theta(\omega_1(C)) / A_1 \in \mathbb{R}^{Pp_1\times Pp_1}$. Applying Eq.~\eqref{eq:cnn-backward-transform} to $K_{\text{E}}^{(1)}$ in Eq.~\eqref{eq:characteristic-function-prior-cnn} in the special case where $L=1$ and $S=\Id_{p_1}$, we obtain that the resulting characteristic function of the output prior is a Gaussian mixture:
\begin{equation}
    \varphi(\bar f\mid X) = \mathbb E_{q_1} e^{-\frac{1}{2} \bar{f}^\top K^{(\mathrm R)}_{q_1} \bar{f} }\, ,
\end{equation}
with
\begin{equation}
\begin{aligned}
 [K^{(\mathrm R)}_{q_1}]^{\mu\nu}
 &=\frac1{p_1}\sum_{ij}(q_1)_{ij}\Theta(\omega_1(C))^{\mu\nu}_{ij},\\
 q_1&\sim\mathcal W_{p_1}(\Id_{p_1}/A_1,A_1)\,.
\end{aligned}
\label{eq:cnn-shallow-local}
\end{equation}
For a single input channel this is the local renormalization found in Ref.~\cite{aiudi2023}, while for $A_1\to\infty$, $P$ fixed, $q_1\to \Id_{p_1}$ giving the large-width kernel first introduced in Ref.~\cite{garriga-alonso2018deep}.

For two layers we again integrate the readout and apply the EWA to the last hidden layer, now conditionally on $K_{\text{E}}^{(1)}$:
\begin{align}
    \varphi(\bar f\mid X)
    &= \mathbb E_{K_E^{(2)}}\, e^{-\frac{1}{2p_2}\Tr \bar F_2^\top K_E^{(2)}\bar F_2} \\
    &= \mathbb E_{K_E^{(1)}}\,\mathbb E_{K_E^{(2)}\mid K_E^{(1)}}\, e^{-\frac{1}{2p_2}\Tr \bar F_2^\top K_E^{(2)}\bar F_2}
  \nonumber\\
    &\stackrel{\rm EWA}{=} \mathbb E_{K_E^{(1)}}\,\mathbb E_{q_2}\, e^{-\frac{1}{2p_2}\Tr[q_2\bar F_2^\top V_2\bar F_2]}\, .
\end{align}
In the previous expression $q_2\sim\mathcal W_{p_2}(\Id_{p_2}/A_2,A_2)$, and the remaining covariance is $V_2 = \Theta(\omega_2(K_{\text{E}}^{(1)})) \in \mathbb{R}^{Pp_2\times Pp_2}$, still fluctuating with the first-layer weights. Note that the convolution averages over matched offsets,
\begin{equation}
    [\omega_2(K_{\text{E}}^{(1)})]^{\mu\nu}_{ij}=M_2^{-1}\sum_{m\in\mathcal M_2} [K_{\text{E}}^{(1)}]^{\mu\nu}_{i\oplus_2 m,\,j\oplus_2 m}\, ,
\end{equation}
i.e. it sums spatial blocks of $K_E^{(1)}$, which in general changes the Wishart fluctuation law even for a linear activation. We therefore specify its fluctuations before summing over offsets. Our strategy amounts to writing the action of the NNGP map $\Theta$ as an entrywise multiplication by a gain $b$, $\Theta(Z)=b\odot Z$ with $b=\Theta(Z)/Z$, and freezing the gain at its large-width value.
Specifically, in the exponent we represent $\Theta(\omega_2(K_{\text{E}}^{(1)}))$ by $\Tr_{m_2}(\widehat H)$, where
\begin{align}
    [\Tr_{m_2}(\widehat H)]^{\mu\nu}_{ij}  & = \sum_{m_2\in\mathcal M_2} \widehat{H}^{\mu\nu}_{m_2i,m_2j} \\
    \widehat H^{\mu\nu}_{m_2i,m_2'j} & = \frac{(b_2)^{\mu\nu}_{ij}}{M_2}\left[ K_\text{E}^{(1)}\right]^{\mu\nu}_{i\oplus_2 m_2,\, j\oplus_2 m_2'}\, ,  \nonumber
\end{align}
and
\begin{equation}
    (b_2)^{\mu\nu}_{ij} = \frac{ [ (\Theta \circ \omega_2 \circ \Theta \circ \omega_1)(C)]^{\mu\nu}_{ij}\,}{[(\omega_2 \circ \Theta \circ \omega_1)(C)]^{\mu\nu}_{ij}}\, ,
\end{equation}
defined by continuity where the denominator vanishes. Note that with this choice we recover exactly the large-width limit. By direct inspection, introducing $\bar{F}_1 = \Id_{M_2 p_2}\otimes\bar{f}$, we can rewrite the exponent as
\begin{equation}
    \Tr[ q_2 \, \bar{F}_2^\top \Tr_{m_2}(\widehat H) \bar{F}_2] = \Tr\Big(\bar F_1^\top\widehat H\bar F_1\,(\Id_{M_2}\otimes q_2)\Big) \, .
\end{equation}
Assuming a Wishart ansatz for $\widehat H$, i.e.
\begin{align}
    \widehat H\ & \sim\  \mathcal W_{M_2 {p_2}P}\Big(\mathcal H/A_1,\ A_1\Big) \\
    \mathcal H^{\mu\nu}_{mi,\,m'j} & = \frac{(b_2)^{\mu\nu}_{ij}}{M_2}\,\big[(\Theta \circ \omega_1) (C)_{i\oplus_2 m,\;j\oplus_2 m'}\big]^{\mu\nu} \nonumber \, ,
\end{align}
we obtain
\begin{align}
    \varphi(\bar f\mid X) = & \mathbb E_{q_2}
    \mathbb E_{\widehat H}
       \exp{\Big[-\frac{1}{2p_2}\Tr\bar F_1^\top\widehat H\bar F_1\,(\Id_{M_2}\otimes q_2)\Big]} \\
       = & \mathbb E_{q_2}\mathbb E_{\mathcal{Q}|q_2} \exp{\Big[-\frac{1}{2} \bar{f}^\top K^{\text{(R)}}_{\mathcal{Q}} \bar{f}\Big]} \nonumber
\end{align}
where
\begin{align}
    q_2 & \sim\mathcal W_{{p_2}}\!\Big(\Id_{{p_2}}/A_2,\;A_2\Big) \label{eq:cnn_conditional_q1q2}\\
    \mathcal{Q}\,\big|\, q_2 & \sim\mathcal W_{M_2 {p_2}}\!\Big( (\Id_{M_2}\otimes q_2)/A_1,\;A_1\Big), \nonumber
\end{align}
and
\begin{equation}
    [K^{(\mathrm R)}_{\mathcal{Q}}]^{\mu\nu}=\frac{1}{p_2}\sum_{m_2,m_2'\in\mathcal M_2}\sum_{ij=1}^{{p_2}}(\mathcal{Q})_{m_2i,\,m_2'j}\;\mathcal H^{\mu\nu}_{m_2i,\,m_2'j} \, . \label{eq:cnn_L2_kernelrenorm}
\end{equation}
It is worth noting that for CNNs the order parameters of different layers are coupled: $q_2$ enters the scale matrix of $\mathcal{Q}$, whose conditional mean $\Id_{M_2}\otimes q_2$ is block-diagonal in the offset index. The two variables can be decoupled by exploiting the invariance of the Wishart law under congruences (see Eq.~\eqref{eq:wishart-law-under-congruences}). Explicitly, the conditioned change of variables
\begin{equation}
    \mathcal{Q}=\big(\Id_{M_2}\otimes q_2^{1/2}\big)\,q_1\,\big(\Id_{M_2}\otimes q_2^{1/2}\big)\, ,\label{eq:mathcalQ_L2_CNN}
\end{equation}
with $q_2^{1/2}$ the symmetric square root of $q_2$, maps the conditional law of $\mathcal{Q}$ into a Wishart law for $q_1$ with identity scale, so that the order parameters become independent,
\begin{equation}
    q_2\sim\mathcal W_{p_2}\big(\Id_{p_2}/A_2,\,A_2\big)\,,\quad
    q_1\sim\mathcal W_{M_2p_2}\big(\Id_{M_2p_2}/A_1,\,A_1\big)\,. 
\end{equation}
In these variables, the block-structured coupling is transferred from the prior of the order parameters to the kernel renormalization scheme, that is still defined through Eq.~\eqref{eq:cnn_L2_kernelrenorm} and the product in Eq.~\eqref{eq:mathcalQ_L2_CNN}, which here corresponds to the compound $\mathcal Q$ matrix of the CNN. 

The construction extends to $L$ hidden layers by iterating the same two steps, 
giving an $\mathcal H^{\mu\nu}$ and a cumulative product $\mathcal Q$ which are indexed by all distinct paths $\pi$ leading from the output to the spatial patch positions of the first hidden layer, with kernel renormalization scheme
\begin{equation}
\relax [K^{(\mathrm R)}_{\mathcal Q}]^{\mu\nu}
 =\frac{1}{p_L\lambda}\sum_{\pi,\pi'}\mathcal H^{\mu\nu}_{\pi\pi'}\mathcal Q_{\pi\pi'}\,.
\label{eq:cnn-path-kernel-renorm}
\end{equation}
Additional details and the full derivation for deep architectures with convolutional filters, including with non-zero mean activation functions, a final average pooling operation and multiple outputs are given in App~\eqref{app:cnn-spatial}.
} 

\section{Posterior sampling experiments at finite width compared to theory predictions}
\label{sec:Posterior_MCMC_results}

\begin{figure*}
    \includegraphics[width=\textwidth]{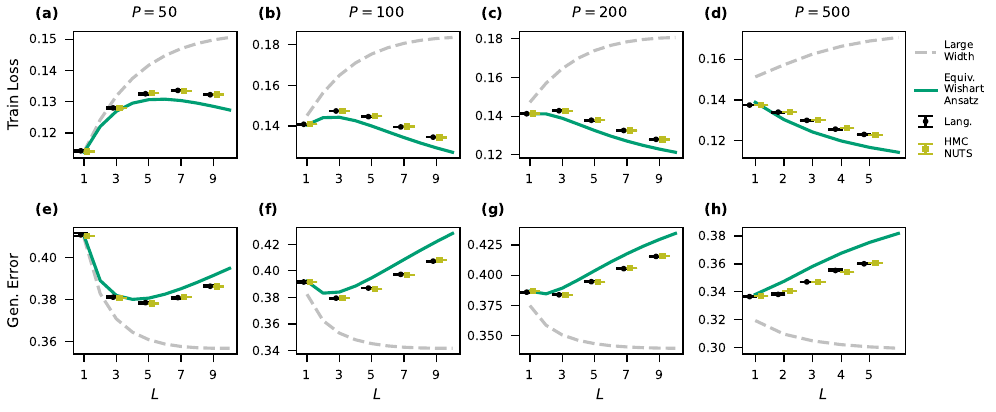} 
    \caption{\newtext{\textbf{Comparison between the learning curves obtained via the Equivalent Wishart Ansatz and numerical experiments for zero-mean activation functions on the CIFAR-10 dataset.} Numerical samples from the Bayesian posterior (black and yellow markers – shifted slightly for better visibility) are compared against the large-width limit predictions (gray dashed lines) and the results of the EWA theory (green solid lines). Both the training loss (first row) and the generalization error (second row) are displayed as a function of the number of hidden layers $L$. We keep the number of neurons and test examples fixed at $N_\ell=200\, \forall \ell$ and $P_t=1000$, while varying the number of patterns $P$ across different columns ($P$ is constant within each column). These simulations refer to the Erf activation function, with Gaussian priors $\lambda = 1$ and temperature $T=0.1$. For all panels, we sample from the posterior using Langevin Monte Carlo with a learning rate $\eta=0.001$ (black dots) and HMC NUTS (yellow squares). The aggregated goodness-of-fit index value $R^2$ is 0.982 (train) and 0.955 (test).}}
    \label{fig:main-erf-cifar}
\end{figure*}

In this section we present an extensive comparison between the learning curves resulting from the EWA analysis and the outcomes of numerical simulations. The main purpose is to assess and quantify the predictive power of the EWA formalism in describing the behavior of fully-connected NNs as a function of the depth $L$, the number of training patterns $P$, and the  number of neurons $N_1,\ldots,N_L$. In contrast to Section \ref{subsec:numerical-validation-ewa},  where the focus was on the rate function of $\mathcal Q$ within the context of the characteristic of the prior, this numerical analysis centers on the posterior predictor statistics. It is worth noting that an additional numerical analysis is required at the level of the posterior, due to the impossibility of checking the EWA for each configuration of the dual outputs $\bar{f}$ over which we integrate in Eq.~\eqref{eq:Z_from_priorcharfunc}. In view of this, numerical checks at the posterior level provide a consistent probe of the capabilities of the EWA, taking into account all the approximations involved.

Here we focus on the training loss and the generalization error, as defined in Eq.~\eqref{eq:err_t_def} and Eq.~\eqref{eq:err_g_def} respectively. These two observables are expected to capture the fundamental behavior of the networks, probing both interpolation and generalization capabilities.
Examples of posterior predictor distributions for individual test points can be found in supplementary Figs.~\ref{Sfig:history_analysis_L5_wellbehaved}-\ref{Sfig:history_analysis_muP_difficult_sampling}.
Since the numerical evaluation of Eqs.~\eqref{eq:err_t_def} and \eqref{eq:err_g_def} requires sampling from high-dimensional and complex distributions, we employed various Markov Chain Monte Carlo (MCMC) techniques to approximate 
these ensemble averages via time-averaging over appropriate stochastic evolutions that can be implemented numerically. We provide the main details regarding the MCMC implementations in Sec.~\ref{subsec:algorithms-used}, while extended discussions are presented in App.~\ref{appsec:ease-and-difficulty-of-sampling-the-overparametrized-MLP-posterior}.

\subsection{MCMC Sampling}
\label{subsec:algorithms-used}

In this work, we primarily employ the Langevin Monte Carlo (LMC) algorithm, a standard method for sampling from high-dimensional distributions that take the form of a Gibbs distribution. It is based on the continuous Langevin equation
\begin{equation}
    \label{eq:langevin-continuous}
    \dot{\theta}(t) = - \nabla_\theta \mathcal{L}_{\text{reg}}(\theta(t)) + \sqrt{2T} \epsilon(t) \, ,
\end{equation}
where $T=1/\beta$ and $\epsilon(t)$ is Gaussian white noise with moments
\begin{equation}
    \langle \epsilon(t)\rangle=0\, ,\quad \langle\epsilon(t)\epsilon(t')\rangle = \delta(t-t') \, . 
\end{equation}
Indeed, the stationary probability distribution $P_\beta(\theta) = \lim_{t \to \infty} P_{t}(\theta)$ defined by the dynamics in Eq.~\eqref{eq:langevin-continuous}, which results from averaging over all possible white noise realizations, is the posterior distribution $P_\beta(\theta) \propto \exp{(-\beta \mathcal{L}_{\text{reg}})}$. In this work, we use the discretized version of Eq.~\eqref{eq:langevin-continuous} for the numerical implementation, namely:
\begin{equation}
    \label{eq:langevin-discrete}
    \theta_{t+1} = \theta_t -\eta\nabla_\theta\mathcal{L}_{\text{reg}}(\theta_t)+\sqrt{2T\eta}\epsilon_t\, ,
\end{equation}
where $\langle\epsilon_t\rangle=0$, $\langle\epsilon_t\epsilon_{t'}\rangle=\delta_{t,t'}$ and $\eta$ is the step size for the numerical integration (or learning rate). 
Note that in Eqs.~\eqref{eq:langevin-continuous} and~\eqref{eq:langevin-discrete}, the gradients are computed using the regularized loss function: this function includes both the Squared Error likelihood contribution and the Gaussian term arising from the prior, i.e.
\begin{equation}
    \mathcal{L}_{\text{reg}}(\theta_t) = \mathcal{L}(\theta_t) - \frac{1}{\beta} \log{\rho(\theta_t)}\, .
\end{equation}
At different time scales, the stochastic evolution driven by Eq.~\eqref{eq:langevin-discrete} exhibits two distinct regimes. Close to the initialization, the dynamics is dominated by the gradient contribution over the white noise, as the system is initialized in a random configuration $\theta_0$ that is typically far from equilibrium. This stage, known as the thermalization phase, persists for a small number of updates and is discarded from the statistical analysis presented below. At larger time scales, once the system reaches the typical configurations of the target distribution---for instance, after $s$ updates---the equilibrium phase begins. This phase persists until the end of the simulation, and the configurations $\{\theta_s, \theta_{s+1}, \ldots, \theta_{s+T}\}$ generated during this stage are used to approximate the ensemble average as:
\begin{equation}
    \langle B\rangle \approx \frac{1}{T} \sum_{t=s}^{s+T} B(\theta_t) \, ,
\end{equation}
where $B(\theta_t)$ is a generic observable computed on the network configuration $\theta_t$. Under general hypotheses, the Monte Carlo average converges to the ensamble average as the inverse square root of the number of samples, i.e., $\langle B \rangle = \frac{1}{T} \sum_t B(\theta_t) + O(1/\sqrt{T})$. The rate of convergence depends on the characteristic time required by the algorithm to generate independent samples, known as the autocorrelation time \cite{Wolff2004}. To account for autocorrelation effects in the estimation of the statistical error, we computed the standard deviation of the mean using the blocking method (see App.~\ref{appsubsec:additional-details-on-monte-carlo-sampling} for further details). \newtext{Note that we use standard deviation to denote the statistical error associated with a sampling measurement, distinguishing it from variance, which we use to denote the fluctuations in predictors and random variables}. In this work, we generated chains of $O(10^7)$ network samples for each experimental point, depending on the computational load at hand, which proved to be sufficient for a reliable estimation of both the mean and the autocorrelation time, and thus of the statistical uncertainty.

In addition to statistical errors, the discretized LMC also introduces systematic errors. Indeed, the stationary distribution of Eq.~\eqref{eq:langevin-discrete} is only an approximation of the Gibbs distribution, with an additional systematic bias arising from the numerical integration that scales linearly with the step size $\eta$, i.e., $P_\beta(\theta) \propto e^{-\beta\mathcal{L}(\theta)}\rho(\theta) + O(\eta)$. To mitigate finite step size effects, we used a small learning rate of the order $\eta \sim 10^{-3}$ such that the systematic error remains smaller than the statistical uncertainty. We verified this condition \textit{a posteriori} by further reducing the learning rate and showing that the new estimation of the mean is compatible with the previous one within our statistical precision (we report a representative example in Fig.~\ref{Sfig:diff-lr}). \newtext{Throughout what follows, results of Langevin Monte Carlo runs are often presented alongside independent simulations performed using the No-U-Turn Sampler, a Hamiltonian Monte Carlo method (NUTS HMC) \cite{hoffman2014no}. The latter, in contrast, produces exact samples and is therefore free from systematic error (see Sec.~\ref{subsec:emergent-performance-transitions} and App.~\ref{appsec:additional-mcmc-samplers} for further details).}

It is worth noting that Eq.~\eqref{eq:langevin-discrete} corresponds to the standard Gradient Descent (GD) dynamics with the addition of a noise source proportional to the model temperature. \added[id=rev1]{The Bayesian NNs explored in this manuscript sample solutions from the canonical Gibbs ensemble at temperature $1/\beta$. They differ from (full-batch) GD-trained models which reach a single solution after training.} Following this analogy, discretized LMC is a form of training dynamics that in the long term limit samples from the Bayesian posterior. Although sampling with MCMC algorithms is not the conventional approach for training modern deep networks due to their high computational cost, it represents a well-established numerical framework for investigating their behavior and performance (see for instance Refs. \cite{Neal1995BayesianLF, izmailov2021bayesian}). \added[id=rev1]{Furthermore, outcomes from Bayesian inference at low temperatures $T$ have been shown to be close to those obtained with GD \cite{wenzel2020how, smith2018bayesian, li2022globally, baglioni2025kernel,Avidan2025}, which further motivates the interest in such training algorithms.}

\subsection{Learning curves for fully-connected DNNs}
\label{subsec:learning-curves-for-fc-dnn}

\begin{figure*}
    \includegraphics[width=\textwidth]{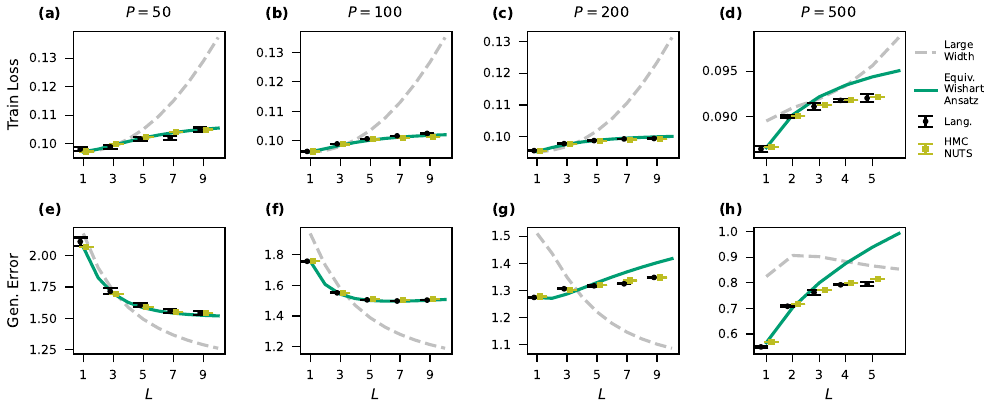} 
    \caption{\newtext{\textbf{Comparison between the learning curves obtained via the Equivalent Wishart Ansatz and numerical experiments for non-zero-mean activation functions on Random Gaussian data.} Numerical samples from the Bayesian posterior (black and yellow markers -- shifted slightly for better visibility) are compared against the large-width limit predictions (gray dashed lines) and the results of the EWA theory (green solid lines). Both the training loss (first row) and the generalization error (second row) are displayed as a function of the number of hidden layers $L$. We keep the number of neurons and input dimensionality fixed at $N_\ell=200\, \;\forall \ell\ge1$ and $N_0 = 300$, while varying the number of patterns $P$ across different columns ($P$ is constant within each column). These simulations refer to ReLU activation function, with critical Gaussian priors $\lambda = 1/2$, $P_t = 1000$ test samples and temperature $T=0.1$. For all panels, we sample from the posterior using Langevin Monte Carlo with a learning rate $\eta=0.001$ (black dots) and HMC NUTS (yellow squares). The aggregated goodness-of-fit index value $R^2$ is 0.992 (train) and 0.914 (test)}.}
    \label{fig:main-relu-gaussian}
\end{figure*}

We conducted extensive simulation campaigns to compare the predictions of the EWA with numerical outcomes obtained via Bayesian sampling. In particular, we measured both the training and generalization losses in DNNs using two stereotypic activation functions: (\textit{i}) Erf, as a non-linear zero-mean activation function, and (\textit{ii}) ReLU, as a non-linear activation function with mean. We focus on three different datasets: the MNIST dataset (classes ``$0$'' and ``$1$''), the CIFAR-10 dataset (classes ``cars'' and ``planes''), and a synthetic Gaussian dataset with linear teacher rule $y^\mu=w^\ast x^\mu$. For MNIST and CIFAR-10, no one-hot encoding was applied to the labels, resulting in single-output DNNs. Additional details regarding the datasets are extensively discussed in App.~\ref{appsec:details-on-datasets}. 

In Figs.~\ref{fig:main-erf-mnist} and ~\ref{fig:main-erf-cifar} we show the learning curves for DNNs with Erf activation functions, trained on the MNIST and CIFAR-10 dataset respectively. The first and the second rows report the training and test loss as function of the number of layers $L$. The number of neurons is kept fixed at $N_1 = \ldots = N_L = 200$, while the number of training patterns $P$ varies across columns, such that different columns correspond to different values of $\alpha_1 = \ldots = \alpha_L = \alpha$. In addition to the EWA predictions (green solid lines) and numerical simulations (black dots), each panel also reports the large-width NNGP prediction obtained at the same $P$ and $\alpha=0$ (gray dashed lines). 
In all panels, numerical simulations differ quantitatively from the large-width theory (and often also qualitatively, see especially Fig.~\ref{fig:main-erf-cifar}) as the network depth increases. In contrast, the EWA predictions are in good agreement, even for a significant number of hidden layers. As $\alpha$ increases, corresponding to the right columns in the figure, the large-width theory becomes increasingly inaccurate even for small depths, while the EWA continues to quantitatively track the experimental behavior, apart from small discrepancies due to the EWA approximation that nonetheless remain limited even for large depths. In the Appendix, Fig.~\ref{fig:main-erf-gaussian}, we show simulations for DNNs with Erf activations on the random Gaussian dataset. Again, the learning curves obtained with the EWA are predictive across the entire range of depths considered, especially for $\alpha < 2.5$. In panel (h) ($\alpha=2.5$) we observe discrepancies between numerical outcomes and EWA predictions for larger depths; nevertheless, even in the worst cases, the EWA approximations for the generalization error exhibit a mismatch no larger than $5$--$8\%$.

In Fig.~\ref{fig:main-relu-gaussian} we show learning curves for DNNs with ReLU hidden units for the random Gaussian dataset. Here the parameter of the simulations are the same as Figs.~\ref{fig:main-erf-mnist} and~\ref{fig:main-erf-cifar}, with the exception of the choice of the Gaussian prior precisions. While in the latter we used standard priors $\lambda=1$, in the ReLU case we tune the precision to critical priors $\lambda=1/2$ so that the relative NNGP kernels are well-behaved at large depth \cite{hanin2024random} \newtext{(in Fig.~\ref{fig:off-crit} we also report numerical simulations with non-critical priors)}. For ReLU the large-width theory again fails to qualitatively capture the behavior of the DNNs. As shown, in these regimes learning is completely dominated by finite-width effects which are neglected by the infinite-width theory. Instead, the EWA learning curves do successfully describe the behavior of ReLU DNNs at finite width. In addition, in Fig.~\ref{fig:main-relu-gaussian} (f, g), the large-width and EWA learning curves cross around $L=3$, with numerical simulations consistently matching the EWA. This indicates that our working regime is far from one where finite-width effects can be treated as mere first-order perturbative corrections. In such regimes, first-order corrections to an observable $A$ are expressed as $A = A^{(0)} + \alpha L A^{(1)} + \dots$, where $A^{(0)}$ is the Gaussian infinite-width prediction and $A^{(1)}$ is a constant first-order perturbative correction \cite{hanin2024random}. Since learning curves that take into account only first-order corrections to the Gaussian large-width theory can correspondingly never cross the zero-order prediction as a function of depth, our numerics clearly shows that the EWA accounts for strong, non-perturbative finite-width effects in a layer-dependent manner. \newtext{To quantify how much of the finite-width effect is captured by the EWA, we introduce, for each observable $\mathcal O$ (train loss or generalization error), the dimensionless goodness-of-fit index
\begin{equation}
\label{eq:skill-score}
    R^2 = 1-\frac{\sum_{i}\big[(\mathcal O_i^{\rm EWA}-\mathcal O_i^{\rm MC})/\mathcal O_i^{\rm MC}\big]^2}{\sum_{i}\big[(\mathcal O_i^{\infty}-\mathcal O_i^{\rm MC})/\mathcal O_i^{\rm MC}\big]^2}\, ,
\end{equation}
where $i$ labels the numerical points (dataset, $P$ and $L$), $\mathcal O^{\infty}$ is the infinite-width prediction, $\mathcal O^{\text{MC}}$ and $\mathcal O^{\text{EWA}}$ are the measured and predicted finite-width corrections, respectively. Eq.~\eqref{eq:skill-score} is a coefficient of determination $R^2$ in which the null model, usually the sample mean of the data, is replaced by the NNGP theory. The denominator is the total finite-width effect present in the data, and the numerator is the part of it that the EWA fails to reproduce, so that $R^2$ is the fraction of the finite-width effect, in the mean-square sense, explained by the EWA. $R^2=1$ corresponds to a perfect prediction and $R^2=0$ to a theory that does not improve on the infinite-width limit. The values obtained separately for the two observables and the two activation functions, averaged over the three datasets, are $R^2=0.990$ (train loss) and $0.953$ (generalization error) for Erf networks, and $0.978$ and $0.970$ for ReLU networks (dataset by dataset, $R^2$ ranges from $0.905$ to $0.999$), implying that the EWA explains between $95\%$ and $99\%$ of the finite-width effect.}

We conclude this section noting that in Fig.~\ref{fig:main-relu-gaussian}(h) in analogy to Fig.~\ref{fig:main-erf-gaussian}(h), the EWA starts to break down for ReLU at $\alpha=2.5$ and depths $L>3$, when the network is trained on random Gaussian inputs. These deviations are expected to gradually emerge as the network depth increases, reflecting the approximate nature of the EWA. 
Similar effects have already been discussed in recent literature, where in the Bayes-optimal setting of student MLPs trained to match random teacher networks, an upper-bound for the generalization can be rigorously proved, proceeding by a layer-wise reduction \cite{camilli2025information}. The upper-bound scales as $O(L/\sqrt{N})$, accumulating errors for each layer reduction. While in our setting with general data a different scaling with $N$ and $P$ is possible, we expect the same scaling of deviations with $L$ holds due to the similar layer-wise structure of the computation. 
Interestingly, for simulations with MNIST and CIFAR-10 the EWA remains predictive also at $\alpha=2.5$ and $L>3$ (see panels (d, h) of Appendix Figs.~\ref{fig:main-relu-mnist} and ~\ref{fig:main-relu-cifar}). The same scaling argument nevertheless suggests that consistent deviations could still appear at larger depths than those explored here.

\subsection{Emergent metastable regime at $L\alpha\gg 1$}
\label{subsec:emergent-performance-transitions}

\begin{figure}[h!]
    \centering
    \includegraphics[width=.5\textwidth]{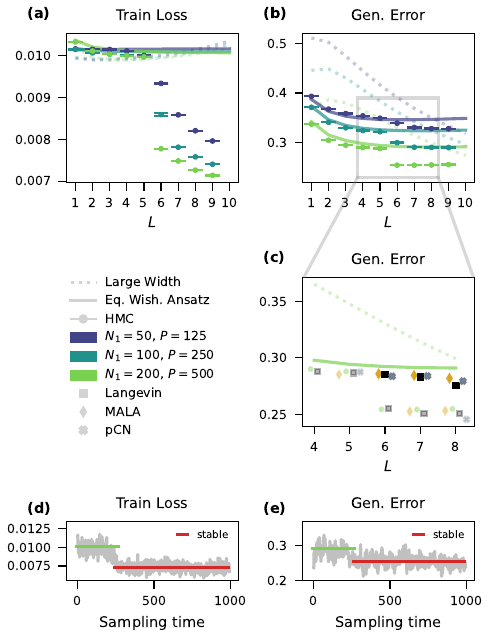}
    \caption{\textbf{Finite-size analysis of an emergent metastable regime at large depth and load $\alpha$.} In panels (a) and (b), the training loss and generalization error are displayed as function of the depth $L$ at fixed network load $\alpha = 2.5$ for the CIFAR-10 dataset. Dashed lines represent the large-width predictions, solid lines are the EWA learning curves, and circles denote numerical simulations performed  with the HMC NUTS algorithm. We employed the ReLU activation function with critical priors $\lambda = 1/2$ at each layer,  temperature $T = 0.01$, and $1000$ test examples. Different colors indicate  different values of $N_1$ and $P$ at the same network load, as shown in the legend (error bars and lines with the same color refer to the same $N_1$ and $P$).  The inset (c) shows numerical outcomes from different sampling algorithms: HMC NUTS (circles), Langevin  (squares), MALA (diamonds), and pCN (crosses). Faded points \replaced[id=rev1]{show the}{indicate that the}  Monte Carlo average \deleted[id=rev1]{is} taken over the \added[id=rev1]{long-term} stable equilibrium \added[id=rev1]{excluding the transient plateau}, while high-contrast  points represent the Monte Carlo average over the transient \added[id=rev1]{plateau} phase.  Panels (d) and (e) show both the transient and equilibrium phases along the HMC NUTS sampling time for both the training loss and generalization error, in the case  of $L=8$, $N_\ell=200$, and $P=500$. Green and red lines indicate the EWA prediction and the final Monte Carlo average over the stable minimum, respectively.}
    \label{fig:finite-size-relu-cifar}
\end{figure}

While the EWA learning curves provide an overall proper description of the behavior of NNs across a broad range of network architectures and learning tasks, in the regime where both $L$ and $\alpha$ are sufficiently and simultaneously large, numerical simulations reveal sudden improvements in performance. In particular, we find a sharp reduction in both the train loss and the generalization error in some cases as depth increases, unveiling an emergent and potentially new form of finite-width contributions. In this section, we provide a systematic investigation of this phenomenon, relating these empirical observations to a metastability in the Bayesian learning dynamics, and perform a finite-size study against the EWA predictions.

In Fig.~\ref{fig:finite-size-relu-cifar} we show MCMC experiments on MLPs with ReLU nonlinear hidden units trained on the same CIFAR-10 task as in Figs.~\ref{fig:main-erf-cifar} and~\ref{fig:main-relu-cifar}. All the simulations in this case refer to the temperature $T=0.01$, $\alpha = 2.5$, and critical priors. Panels (a) and (b) report the main message: at $L=6$, both the train and the test loss exhibit sudden drops (see colored circles). It is worth mentioning that these numerical outcomes are obtained using NUTS HMC, which is a state-of-the-art MCMC sampling algorithm that ensures improved performance compared to LMC. While being built on top of the vanilla HMC, NUTS HMC integrates features such as automatic fine-tuning of hyperparameters, including the integration step and trajectory length. In our case, the more powerful but less interpretable NUTS HMC is implemented to \replaced[id=rev1]{guard against possible}{avoid} finite learning-rate systematics and \added[id=rev1]{to} counteract slowdowns in the Bayesian sampling at large $L\alpha$. Posterior sampling experiments clearly show that: (\textit{i}) the transition is not a finite-size effect that vanishes in the proportional regime $N_\ell, P \to \infty$, since the drop becomes more and more pronounced as we simulate networks with a larger number of neurons and training patterns (see the different colored circles in panels (a, b), which show increasing $N_\ell$ and $P$ at fixed $\alpha$); (\textit{ii}) both the large-width theory and the EWA continue to predict smooth behavior in these regimes. Since consistent Bayesian sampling is known to be challenging under such conditions, i.e. where one has simultaneously large depth, a large number of training examples, and low temperature, we implemented several MCMC samplers to ensure the robustness of our results. In panel (c) of Fig.~\ref{fig:finite-size-relu-cifar} (shaded points) we also display results obtained with the Metropolis-Adjusted Langevin algorithm (MALA) \cite{besag1994comments, roberts1996exponential} and the preconditioned Crank--Nicolson MCMC algorithm (pCN) \cite{cotter2013mcmc}, along with LMC and NUTS HMC results. We highlight that while LMC and MALA are gradient-based samplers, pCN is an energy-based sampler, whereas NUTS HMC is a momentum-driven sampler, so that biases arising from common strategies in exploring the configuration space are avoided. The results of the numerical experiments are fully in agreement with each other, providing strong numerical evidence that additional finite-width effects in DNNs are at play. 

Quite interestingly, we found that the Bayesian sampling dynamics exhibit for those points at $L>5$ a new intermediate phase between the thermalization and the equilibrium phases, which we refer to as a transient phase. Configuration updates in the transient phase are characterized by a balance between gradient forces and random fluctuations, so that the system seems to be at thermal equilibrium for long simulation times, until the thermalization dynamics restart and drive the system toward the true long-term equilibrium distribution. The number of epochs needed to escape the transient phase depends on the algorithm and can last up to millions of updates, and therefore may persist longer than the sampling experiment even for simulations with a large number of epochs. An example comparing two independently initialized LMC chains is shown in Fig.~\ref{Sfig:history_analysis_L7_bistability}. It is worth mentioning that in our experience we observed drifts from the \newtext{transient plateau to the equilibrium phase, and never up-transitions: the empirical absence of an up-transition in these histories could be due to a larger energetic- or entropic barrier in this direction.} Taking advantage of these equilibrium-like properties of the transient phase, we computed the ensemble averages over those configurations and found that they not only smoothly continue the empirical learning trends observed at smaller depths $L\le5$, but are also in agreement with the smooth behavior predicted by the EWA theory. In Fig.~\ref{fig:finite-size-relu-cifar}, panels (d, e), we display the onset of the transient phase in NUTS HMC for the train and test losses. Even if for NUTS HMC the transient phase lasts a few hundred epochs, due to the optimal preconditioning of the simulation parameters and much larger step distance per sample, it is evident that the EWA predictions (green lines) are in agreement with the \replaced[id=rev1]{MCMC sampling during the transient plateau (green line)}{Bayesian sampling}, while after the drop the \replaced[id=rev1]{putative long-term equilibrium}{system} fluctuates around different loss minima \added[id=rev1]{no longer captured by the EWA} (the red line indicates the \added[id=rev1]{empirical} average over the equilibrium phase). In Fig.~\ref{fig:finite-size-relu-cifar}, panel (c), we show all the different averages together: shaded points correspond to equilibrium averages, while unshaded points represent transient \added[id=rev1]{plateau} averages. Up to $L=5$ no drops are detected, while for $L \ge 6$ we report both equilibrium and transient measurements, with the latter being in good agreement with the smooth learning curves predicted by the EWA.

\subsection{\added[id=rev1]{Multiple outputs and classification}}
\label{subsec:learning-curves-multiple-outputs-classification}

\begin{figure}
    \centering
    \includegraphics[width=\columnwidth]{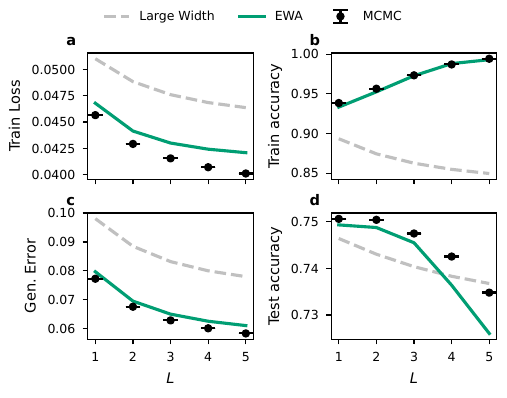}
    \caption[Multiple outputs and classification on FashionMNIST.]{\textbf{\added[id=rev1]{Multiple outputs and classification on FashionMNIST.}} \added[id=rev1]{Training (top) and test (bottom) observables as a function of the number of hidden layers $L$ for MLPs with $D=10$ outputs, corresponding to all 10 classes of FashionMNIST. (a,c) The posterior-averaged squared error for regression on one-hot targets. (b,d) Posterior-averaged accuracy for the same networks trained with cross-entropy loss.
    Network parameters are $N_\ell=250$, $P=500$, Erf activation, $\lambda=1$ and at $T=0.1$. Numerical samples from the posterior and their uncertainty estimates (black) are obtained with HMC, the size of the test set is $P_t =2000$.}
    }
    \label{fig:multioutput-mlp-depth}
\end{figure}

\added[id=rev1]{We test the multiple-output theory of Sec.~\ref{subsec:multiple-outputs} on all ten classes of FashionMNIST, also comparing regression models trained with square loss and classification models trained with cross-entropy loss for the same network architectures and dataset. The MLPs had $D=10$ outputs and depth from $L=1$ to $5$ at fixed load $\alpha=2$ with $N=250$, $P=500$. For square loss, Fig.~\ref{fig:multioutput-mlp-depth}(a,c) shows both the training loss and the generalization error decreasing with depth. The EWA prediction stays within $5\%$ of the sampled test error, and the scale of the training loss makes a small but systematic difference for larger depths visible.
For this multi-output experiment, the $\mathcal Q$ is a $D\times D$ symmetric matrix of $55$ parameters, therefore the saddle point approximation in $\mathcal Q$ for $P=500$ may carry a more significant finite-size error than in the scalar output examples. }

\added[id=rev1]{For the cross-entropy case with the same architecture and 10-class data, shown in Fig.~\ref{fig:multioutput-mlp-depth}(b,d), the training accuracy rises instead from $93\%$ at $L=1$ to over $99\%$ at $L=5$, while the test accuracy falls from $75$ to $73.5\%$. The EWA prediction captures both of these behaviors. Due to the use of the Laplace method in evaluating the non-Gaussian logit posterior integral, the theory here contains an additional approximation both in the EWA and the Large Width (NNGP) lines. However, the relative Laplace error appears to be small in this setting, as we checked by fixing the EWA saddle-point value $\mathcal Q_\ast$ and sampling the integral numerically at the $L=4$ point, yielding a difference to the Laplace value of $<1\%$ (not shown).}

\begin{figure*}[t]
    \centering
    \includegraphics[width=\textwidth]{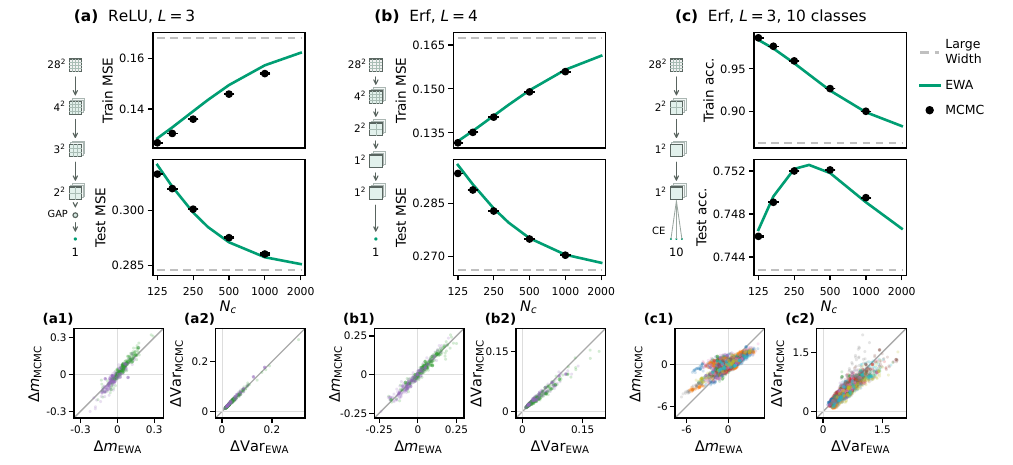}
    \caption[Spatial kernel renormalization in deep CNNs.]{\textbf{\added[id=rev1]{Spatial kernel renormalization in deep CNNs on FashionMNIST.}} \added[id=rev1]{The upper panels show posterior-averaged training and test errors or accuracies versus the number $N_c$ of channels per layer, for $P=500$, $P_t=2000$ and $T=0.1$. (a) A three-layer ReLU CNN with global average pooling and (b) a four-layer Erf CNN, both trained with square loss on Pullover vs. Coat regression (classes 2 and 4). (c) A three-layer Erf CNN classifier using all ten classes with cross-entropy loss. The prior precision is $0.5$ for ReLU and $1$ for Erf. The green curves give the  EWA saddle, the gray dashed curves the large-width posterior, and the black points finite-network HMC estimates. In the lower panels, for the respective $N_c=125$ ($P/N_c=\alpha=4$) points, each axis gives the change relative to the large-width predictor. We compare changes predicted by EWA (horizontal) and observed in HMC (vertical) for the test predictive mean (a1--c1) and marginal variance (a2--c2) on 400 individual test points. Colors denote the true class. For (c1,c2) each classification image contributes ten centered logits. }
    }
    \label{fig:cnn}
\end{figure*}

\subsection{Learning curves for \added[id=rev1]{convolutional networks}}
\label{subsec:learning-curves-for-cnns}

\added[id=rev1]{Fig.~\ref{fig:cnn} tests the deep spatial renormalization scheme of Sec.~\ref{subsec:the-stacked-equivalent-wishart-ansatz-for-dnns-with-convolutional-layers} as the number of channels $N_c = A_\ell \; \forall \ell$ varies at fixed training set size $P=500$. Smaller $N_c$ therefore corresponds to large network load $\alpha =P/N$, while large $N_c$ eventually recovers the large-width limit. The three tested CNN architectures are sketched next to the panels (a,b,c). 
We evaluate Eq.~\eqref{eq:cnn-path-kernel-renorm} by summing over pairs of convolutional paths, specified by their filter offsets at successive layers (see App.~\ref{app:cnn-spatial}). The four-layer Erf network in panel (b), for example, reduces the $28\times28$ images to a $4\times4$ grid in the first layer, then to $2\times2$ and to a single position in layer 3, after which furthermore a fully-connected layer gives the last-layer representation. The final readout has no choice of position in the last layer, yet the predictive kernel can still reweight contributions that passed through different positions in the earlier layers. In the three-layer ReLU network of panel (a), averaging over the last $2\times2$ grid by global average pooling (GAP) also retains contributions from pairs of earlier positions. Compared with the one-hidden-layer result of Ref.~\cite{aiudi2023}, the spatial structure available for renormalization now depends on the entire hierarchy of convolutions.}

\added[id=rev1]{For the two single-output regression models in panels (a,b), reducing the channel count lowers the training loss while increasing the generalization error in this setting. In contrast to scalar-output MLPs where a scalar $\mathcal Q$ multiplies the whole kernel and cancels from the mean predictor at zero temperature, in the CNNs, finite width produces appreciable shifts in the posterior mean on individual images. The spatial renormalization captured by the EWA for CNNs reproduces much of this image dependence in both architectures, see Fig.~\ref{fig:cnn}(a1,b1), as well as the increase in predictive variance (a2, b2). In these scatter plots the difference of mean and variance of individial sample predictors is shown relative to the large-width NNGP predictors; for $N_c \to\infty$ all points would therefore become concentrated at zero in these plots.  
In Fig.~\ref{fig:cnn}(c), a three-layer CNN classifier that has multiple outputs and is trained with cross-entropy loss on all 10 classes of FashionMNIST is shown. As in the previous section, the EWA and large-width theory lines here also use the Laplace approximation to compute observables averaged over the posterior predictor.
In this ten-class experiment the training accuracy grows as the channels become fewer and network load increases, while the test accuracy has a shallow maximum at intermediate width. The EWA captures both trends, although it overestimates the magnitude of many shifts of the mean class logits, Fig.~\ref{fig:cnn}(c1). A class-resolved comparison across true class- and  logit-labels (Fig.~\ref{Sfig:cnn-class-shifts}) and an additional experiment with three different activation functions at fixed CNN architecture (Fig.~\ref{Sfig:cnn-gap-activations}) are described in App.~\ref{app:cnn-spatial}.}

\subsection{Mean field versus Standard parametrization of the weights}
\label{subsec:mean-field-param}


\begin{figure}
    \centering
    \includegraphics[]{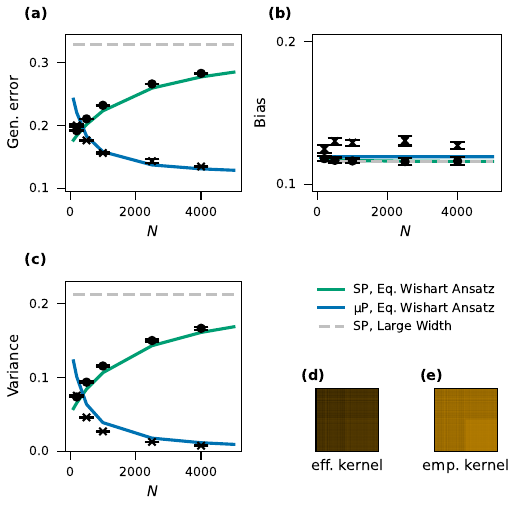}
    \caption{
    \textbf{Comparison of deep networks in maximal-update ($\mu P$) and standard parametrization (SP) to EWA predictions.}
    (a,b,c) Generalization error and its bias - variance decomposition Eqs.~\eqref{eq:bias-definition},\eqref{eq:sigma2-definition} for a $L=4$ hidden layer MLP with $P=1000$ and ReLU activation on the CIFAR10 "cars" vs. "planes" task.
    Mean-field / $\mu$P scaling results in theory (blue) and LMC experiments (black crosses), SP results in theory (green) and LMC experiments (black dots). The sizable performance difference between SP and $\mu$P is mostly driven by predictor variance. Both proportional regimes differ from the lazy large-width limit (gray dashed).
    (d) The effective rescaled kernel $K^{(\text{R})}_{\mathcal Q^\ast}(C)$ underlying the theory prediction for the $N=4000$ point\replaced[id=rev1]{ in $\mu$P}{s}. Here $\mathcal Q^\ast = 238.5$ and the rescaling factor relative to the prior kernel is $\frac{\mathcal Q^\ast}{\gamma^2 \lambda} = 0.119$.
    (e) The empirical Gram matrix of last-layer activations in the posterior for the $N=4000$ point\replaced[id=rev1]{ in $\mu$P}{s}, showing strong learned block structure. Both panels share the same color scale, and indices are sorted by the $\{0,1\}$ class labels.
    Temperatures for all points are $T_{\mathrm{SP}}=0.1$ and $T_{\mu P}= T_{\mathrm{SP}}/\gamma^2 =0.1/N$, with feature learning scale $\gamma_0=1$.
    }
    \label{fig:muPvsSP}
\end{figure}

Here, we ask in how far the low-dimensional effective action obtained through the EWA can also be valid in the rich learning regime.
To do so, we change from standard parametrization to the $\mu$P by defining $f_{\mu\mathrm{P}} = \frac{1}{\gamma} f_{\mathrm{SP}}$, where $\gamma = \gamma_0 \sqrt{N}$. 
From the Bayesian perspective, this corresponds to reducing the scale of the prior by factor $\gamma$ and is equivalent to reducing the variance of the weights in the readout layer as $\lambda_L^{-1} \to \gamma^{-2} \lambda_L^{-1}$. Simply changing $\lambda_L$ is \added[id=rev1]{thus} convenient in the theoretical expressions. \deleted[id=rev1]{However, i}\added[id=rev1]{I}n the gradient-based LMC simulations\added[id=rev1]{ however,} we use instead the explicit factor in the network output, $\frac{1}{\gamma} f_{\mathrm{SP}}$, which \replaced[id=rev1]{gives}{corresponds to} the same output posterior $P_\beta(f)$ but is a choice of weight parameters in which gradients in all layers are reduced equally, allowing to use the standard increase of learning rate in $\mu$P of $\eta \to \gamma^2 \eta$.

As a final consequence of the change from SP to $\mu$P, also the temperature has to be scaled down \added[id=rev1]{with network size} as $\frac{1}{\gamma^2} T$ to avoid dominance of the sharper prior over the likelihood term \cite{lauditi25a,coding-schemes-2025}. This means that SP and $\mu$P \added[id=rev1]{are two different Bayesian models that }can not be compared consistently at finite temperatures\replaced[id=rev1]{.}{, since $\mu$P requires $T\to 0$ to perform better than samples from the prior in the high-dimensional limit.} 
\deleted[id=rev1]{This susceptibility to noise can be seen as a limitation of the $\mu$P parametrization. However, the zero-temperature limit is not unrealistic since in practical training, output noise due to floating point error is small and noise arising from finite batch sizes and learning rates is progressively reduced over training time. }
\added[id=rev1]{As the ideal $T=0$ cannot be sampled by MCMC, comparing the two models at $T_{\mu P}= T_{\mathrm{SP}}/\gamma^2$ is a balanced choice in the sense that the relative scale of prior and likelihood terms is kept constant.}
At the finite sizes of $N\sim 10^3$ that we probe in our experiment, we found that using a temperature $T_{\mu P}=10^{-1}\gamma^2 \sim 10^{-4}$ was both feasible to sample and sufficient to obtain behavior qualitatively similar to that in the zero-temperature limit. 

Overall, we find that the EWA successfully predicted generalization performance even in the rich learning regime.
Fig.~\ref{fig:muPvsSP} shows a comparison of SP (green) and $\mu$P (blue) for ReLU MLPs of depth $L=4$ trained on the CIFAR-10 “cars” vs. “planes” task with $P=1000$. The behavior of the generalization error is well captured in both parametrizations, and again clearly differs from the large-width limit (dashed gray). In panels (b,c), the error is decomposed into the contributions arising from the bias and variance of the posterior predictor distribution. The substantial improvement of the performance in $\mu$P seen in (a) is almost entirely explained by a reduction of variance in $\mu$P (c), while the bias remains similar to SP (b).

It appears counter-intuitive that the EWA can capture the behavior of $\mu$P, because the renormalized kernel Eq.~\eqref{eq:renormalized_kernel} does not adapt in the same way as the empirical hidden-layer kernels do in the rich regime. It is a characteristic of $\mu$P that on a binary classification task as shown in Fig.~\ref{fig:muPvsSP}, the posterior empirical kernels show a strong block-structure when the sample indices are sorted by class label; while the renormalized kernel $K_{\mathcal Q}^{(R)}$ retains the prior structure and only receives a global rescaling factor.
However, this rescaling can \textit{effectively} take into account the adaptation of the output posterior to the data. 
\added[id=rev1]{The EWA with effective Gaussian process covariance kernel $K_{\mathcal Q}^{(R)}$ is a theory of the posterior network outputs, not of the posterior Gram matrices encoding the representation in each layer.}
Indeed, Fig.~\ref{fig:muPvsSP}(e) shows that the empirical kernel of the last-layer representation has acquired a block structure consistent with strong feature learning of the $\{0,1\}$ class structure, while the generalization error remains well predicted by the unstructured effective kernel (d).
This at first surprising behavior is \replaced[id=rev1]{known in}{consistent with} deep linear networks however. Here it can be shown that the \replaced[id=rev1]{rescaling of the posterior output}{renormalized effective} kernel is equivalent to \deleted[id=rev1]{adding} a learned low-rank contribution $\propto yy^\top$ \replaced[id=rev1]{in}{to} the empirical \replaced[id=rev1]{representation kernels of the training set}{kernels} at each layer \cite{Bassetti:JMLR:2024, SompolinskyLinear,PhysRevE.105.064118}, which corresponds to a block structure like that observed in the empirical kernel \added[id=rev1]{of panel (e)} undergoing feature learning.

For the single output task as in Fig.~\ref{fig:muPvsSP}, in the zero temperature limit the renormalization factor of the kernel cancels out in the mean predictor Eq.~\eqref{eq:bias-definition} while entering linearly into the predictor variance Eq.~\eqref{eq:sigma2-definition}, which is strongly reduced in $\mu$P. The effect is that at low temperature, 
the improvement of MLPs in the rich learning regime over SP is mostly attributable to a difference in variance, while the biases are similar when the EWA is a valid approximation. This can explain the behavior observed in the experiments on real data we performed in the proportional sample-width regime. 
However, this does not preclude the existence of tasks where the EWA breaks down more strongly and a qualitative difference in the biases appears.
A detailed empirical and theoretical exploration of output posteriors $P_\beta(f)$ in the rich learning regime will therefore be the subject of a follow-up work.

\section{Discussion and Conclusions}\label{sec:discussion}

We have presented an approximate theory describing the generalization error in Bayesian multi-layer, nonlinear networks at finite width. In the proportional scaling regime where the number of samples $P$ and the width $N$ are comparably large, this theory successfully captures the empirical performance of Monte-Carlo sampling experiments across multiple architectures and datasets, particularly for moderate network load $\alpha = \frac{P}{N} \lesssim 1$.

Building upon work on deep linear networks, we show how an Equivalent Wishart Ansatz reduces the intractable hierarchy of empirical kernel fluctuations to a low-dimensional effective theory in terms of a renormalized kernel and a small number of self-consistent order parameters. Within this framework, both fully-connected and convolutional layers can be treated, and we expect that generalizations to other architectures admitting a well defined large-width limit are possible.

A few additional aspects are worth emphasizing: First, the effect of non-zero activation means is asymptotically suppressed, so that the zero-mean EWA remains the relevant effective description also for activation functions such as ReLU. Second, the theoretical predictions are supported by an extensive set of Bayesian sampling experiments across depths, datasets, and activation functions, which to our knowledge constitutes the broadest such test so far performed on deep MLPs. Finally, the results suggest that learning regimes in large but finite-width networks can be  organized by the network load $\alpha=P/N$, and therefore by the relative scaling of data and width, rather than only by the usual lazy-versus-rich parametrization distinction.

\subsection{Focusing on the layer-wise kernels as random matrix ensembles}

In this paper, we focus on the kernel matrices as the fundamental random variables of the learning system. A theory describing deep network learning outcomes in a Bayesian setting is then always a more or less faithful approximation of the prior ensemble of kernel matrices. In the EWA, we aggregate all randomness in the lower-layer kernels into approximately Wishart fluctuations of the last-layer kernel matrix. This notably differs from a Gaussian approximation in the pre-activations in each layer, which would give the infinite-width NNGP result. 

Since a Wishart kernel by definition in Eq.~\eqref{eq:defWishartEnsemble} also follows from Gaussian fluctuations in the post-activations, a Gaussian Ansatz for the last-layer post-activation distributions arising from each of the lower layer weight priors would also imply the EWA. However, this transports no physical insight, since the individual nonlinear post-activations in reality are never Gaussian distributed, also not in the infinite-width limit. The kernel matrix entries (suppressing layer indices for brevity) $K^{\mu \nu}=\frac{1}{\lambda N} \sum_i \sigma(h^\mu_i)\sigma(h^\nu_i)$ instead, being always high-dimensional sums of products, behave close to Wishart even if the $\sigma(h_i)$ are not Gaussian, and also the continuation to the infinite-width limit holds. The perspective of approximating the kernel distribution is therefore physically more natural. It is nonetheless important to note that it is still an \textit{equivalent} Wishart Ansatz: For the EWA to hold, the kernel needs not be exactly Wishart but we only need the implication $Q=\frac{\bar f^\top K \bar f}{\bar f^\top \mathbb E[K] \bar f} \sim \chi^2_N$ to hold for those $\bar f$ which dominate the partition function Eq.~\eqref{eq:Z_from_priorcharfunc}.

Overall, \deleted[id=rev1]{ we argue that} the ensemble of kernel matrices may be the more convenient object to study learning in these systems\added[id=rev1]{, as is also supported by previous literature \cite{SompolinskyLinear,zavatone2021depth,aitchison21a, milsom2024convolutional} }. In deep linear networks, a full understanding of the posterior appeared as a highly complex problem \cite{zavatone-veth2021exact,HaninZlokapa23} until by considering the ensemble of kernel matrices the posterior distribution could be characterized exactly at arbitrary values of $N,P$ and $L$ \cite{Bassetti:JMLR:2024}. With this framing as a random matrix problem we hope to stimulate also mathematically rigorous work on the kernel ensembles in nonlinear multi-layer networks.

It may appear as a limitation that pairwise kernels are naturally sufficient statistics for square loss, while cross-entropy loss takes also higher-order information of the output prior into account. However, we note that the joint output prior Eq.~\eqref{eqprioralternative} can always be expressed as an integral representation in terms of the pairwise kernel. Using cross-entropy loss only means that the posterior is no longer an analytic Gaussian integral of the prior characteristic function. \added[id=rev1]{Posterior expectations can still be calculated in the Laplace approximation, by numerically finding the maximum a posteriori (MAP) value and Hessian of the output logits.}

\subsection{Finite- versus infinite-width generalization performance}

The proportional sample/width limit provides a natural framework to study large but finite-width overparametrized networks. Throughout, we have seen that the finite-width generalization performance can differ strongly from the infinite-width (lazy) NNGP prediction. 

At the same time, our experiments and approximate theory show that for MLPs the finite-width correction takes a surprisingly simple form: to leading order, the effect is captured by a scalar renormalization of the kernel Eq.~\eqref{eq:renormalized_kernel} by a factor $\mathcal Q$. This indicates that at network load $\alpha = O(1)$ the amount of data structure extracted by deep fully-connected networks is limited. The significant difference to the infinite-width limit $\alpha \to 0$ is here driven mostly by a difference in the predictor variance. This is in contrast to deep convolutional architectures, where the \replaced[id=rev1]{spatial}{local} kernel renormalization Eq.~\eqref{eq:cnn-path-kernel-renorm} is naturally more structured\added[id=rev1]{: Here contributions of individual paths through the spatial patch hierarchy and their correlations can be reweighted by a matrix $\mathcal Q$}. Because of this, at $\alpha = O(1)$ also the bias of finite-width CNNs can differ substantially from the infinite-width limit. From a broader perspective, while much theoretical work has focused on the rich learning regime to go beyond the lazy infinite width limit, we argue that the relative scaling of the number of samples and the width, encoded by $\alpha$, may be a more fundamental organizing principle for finite-width learning than the lazy-versus-rich dichotomy formulated only in terms of output scaling.

\subsection{EWA as a baseline: Relation to adaptive kernel theories}

A line of recent works have studied feature learning in the mean-field or $\mu$P setting, by deriving theories in which the full $P\times P$ kernels are high-dimensional order parameters adapting to the data in the posterior\cite{seroussi2023natcomm,fischer24critical,lauditi25a} or during gradient descent dynamics \cite{bordelon2022dmft,lauditi25a}, typically together with backward-propagating conjugate or gradient kernels that are also fully adaptive. These are valuable and complex frameworks, but they are still intrinsically high-dimensional: the dominant saddle point must be found in coupled $P\times P$ matrix equations whose complexity, in the proportional regime, is not fundamentally simpler than that of sampling the original $N\times N$ weight posteriors. Ref.~\cite{rubin25a} provides a unifying account at the example of shallow networks, where the dimensionality of the order parameters can furthermore be reduced to $P$-dimensional mean-output discrepancies. In Ref.~\cite{rubin2025mitigating}, steps are being made to simplify the $L$-layer $P\times P$ equations by choosing a lower-dimensional variational Ansatz, which we believe is a very promising future direction. 
Another notable approach is Ref.~\cite{coding-schemes-2025}, which also requires high dimensional optimization but focuses on nonlinearity-dependent inhomogeneities in the activation prior, that can induce discrete or sparse coding schemes in the activation posterior if outputs are further scaled down compared to $\mu$P.
 
Against this background, the EWA provides a strong and missing baseline. It is a direct nonlinear generalization of the type of kernel adaptation that is expected in deep linear networks, and remains easy to compute and interpret for deep architectures. In the present manuscript this lets us study networks with ten hidden layers, going beyond the two- to four hidden layer networks presented in previous adaptive-kernel studies. Moreover, the results show that especially in the regimes at moderate depth and load which are most accessible to sampling experiments, a much simpler theory already explains learning and generalization with high accuracy. For this reason, we believe that the EWA should serve as a baseline when studying the emergence of more directional forms of feature learning in nonlinear networks: comparing only to the infinite-width limit can be misleading, because the large gap between infinite-width NNGP and finite-width phenomenology is mostly captured by the directionally homogenous renormalized-kernel picture.

\subsection{Emergence of deviations from the EWA theory at large depth $L$ and large load $\alpha$}

Aside from the good fit obtained in the majority of the experiments, we observed two types of systematic deviations from the EWA, both appearing when the depth $L$ and the load $\alpha$ become large. In both cases we performed additional sampling experiments, which suggest that the effect is not due to a simple failure or bias of the sampler, even though sampling itself becomes more difficult and has longer convergence times in this regime.

The first type of deviation develops gradually with increasing depth in the Gaussian-data experiments at $\alpha=2.5$, see Figs.~\ref{fig:main-relu-gaussian}(h) and \ref{fig:main-erf-gaussian}(h). There, the EWA approximation appears to deteriorate smoothly as $L$ grows, while notably the agreement remains very good at $L=1$. This could reflect an accumulating error in the predicted width of the rate function of the product $\prod_\ell q_\ell$, or it could be a sign of directional inhomogeneity in the prior that is neglected by the EWA. Furthermore, also finite-size effects are expected to accumulate across layers in deep networks \cite{HaninNica2020,hanin2024random,HaninZlokapa2026}. 

The second type of deviation is qualitatively different. For the CIFAR-10 task with ReLU activations, around $\alpha=2.5$ and $L>5$, we observed an abrupt transition in both training and test error (Fig.~\ref{fig:finite-size-relu-cifar} and Fig.~\ref{Sfig:history_analysis_L7_bistability}). This phenomenon is interesting and distinct from the gradual drift described above, showing all the signatures of a metastability in the posterior where the macro-state described by the EWA loses stability. It is not a finite learning-rate effect of the Langevin sampler: we replicated it with several sampling algorithms that reduce or eliminate finite-step-size bias, and found that increasing $N$ and $P$ sharpens the transition. At present we are not aware of a known mechanism in the literature that could explain this effect. Both types of deviations therefore point to not yet understood phenomena that arise specifically in deep and nonlinear networks, rather than shallow ones, and they provide a natural motivation for further theory development. We have begun follow-up investigations in this direction.

\subsection{Sampling from posteriors with millions of dimensions}

Due to the curse of dimensionality, guarantees on equilibration times for MCMC samplers become uninformative already at $d\gtrsim100$ and particularly in the very high dimensions relevant for deep learning.
Before the seminal work of Li and Sompolinsky on deep linear networks \cite{SompolinskyLinear},
it was therefore often viewed as beyond the capabilities of MCMC methods
to reliably sample from the posterior over the weights of deep neural networks.
This skepticism was also tied to the older picture of deep network loss landscapes as being filled with bad local minima that  would only be avoided by dynamical biases of the optimization algorithms, a view that has since been revised substantially in overparametrized settings: these loss landscapes have relatively flat, non-convex basins connected to the global minima, instead of a proliferation of local minima trapping dynamics \cite{choromanska2015loss-surfaces,kawaguchi2016withoutlocalmin,Garipov2018mode-connectivity,draxler2018nobarriers}.

Here we provided a systematic study showing to which extent MCMC sampling is reliable also for deep and nonlinear networks, across several datasets, activation functions, and prior parametrizations. 
In the experiments presented here, even at system sizes with $L \times N^2 \sim 10^6$, independent Monte-Carlo chains produce consistent posterior statistics over feasible simulation times, including predictor distributions resolved at the level of individual test points (Fig.~\ref{Sfig:history_analysis_L5_wellbehaved}). 
This can partly be explained by the relevant function-space observables being only the $O(P)$ output degrees of freedom. Overall, scaling up the network size and the load parameter $\alpha$ slows down sampling, yet not so severely that at $\alpha \approx 1$ system sizes of $N,P \sim 10^4$ or more would be out of reach.

Nonetheless, exponentially long timescales can be hidden behind apparently flat histories, such that sampling experiments should be interpreted with restraint. Indeed, the metastability phenomenon we report in Figs.~\ref{fig:finite-size-relu-cifar},~\ref{Sfig:history_analysis_L7_bistability} happened for $L=6$ on a timescale of millions of steps for LMC, MALA, and pCN with all predictor statistics seemingly converged. The available evidence thus suggests that large-scale Bayesian sampling in deep networks is feasible enough to be scientifically useful, but also subtle enough that equilibration can not be taken for granted.

\subsection{Beyond the proportional regime}

Our results point to the relative scaling of sample size $P$ and width $N$ as the key control parameter for the complexity of the output prior, and therefore the posterior adaptation structure. In the proportional regime, already the large-deviations structure of the prior becomes important. For MLPs the adaptation captured by the EWA remains low-dimensional, while successfully explaining the empirical learning curves seen in our sampling results.

This suggests that genuinely high-dimensional adaptation in MLPs may require scaling regimes beyond $P \sim N$, where the number of samples grows fast enough to probe richer posterior structure than the one encoded by a renormalized kernel alone. This perspective is consistent with recent work on random teacher-student problems for deep MLPs, where a quadratic number of samples is needed for specialization on the teacher function \cite{cui2023optimal,barbier2025,maillard2024,erba2026}.
The proportional, overparametrized regime is a good proxy for some practical machine learning tasks, while other cases leveraging extreme amounts of data, especially for pre-training of vision- or language models \cite{hoffmann2022computeoptimal} could be better described by quadratic scaling, falling outside the overparametrized regime. Understanding the capabilities and limitations of the most common deep learning architectures as a function of the sample size $P$ relative to the network's width $N$ is therefore an important direction for future work.

\subsection{Outlook}

In this work we have proposed a tractable effective theory for Bayesian deep networks in the proportional regime, showing that a renormalized kernel description can capture finite-width learning curves in nonlinear MLPs far beyond the lazy infinite-width limit and on real-world datasets.

Several directions now appear natural. A first step is to extend the framework to more elaborate convolutional architectures including max-pooling, and to other network families with a richer internal geometry. Secondly, further scaling up sampling experiments could both enable direct comparisons with benchmark performance values and to map systematically if and where the EWA baseline deteriorates as depth and network load increase. Understanding those deviations, and determining whether they reflect finite-size effects or missing ingredients in the effective description, should be a particularly interesting direction for future work towards a theory of learning in deep networks.

\section{Acknowledgements}

R.B., R.P., P.R. and P.B. are supported by $\#$NEXTGENERATIONEU (NGEU). R.B. and P.R. are funded by the Ministry of University and Research (MUR), National Recovery and Resilience Plan (NRRP), project MNESYS (PE0000006) ``A Multiscale integrated approach to the study of the nervous system in health and disease” (DN. 1553 11.10.2022). R.P. and P.B. are funded by MUR project PRIN 2022HSKLK9 and P2022A889F. 
This research benefits from the HPC facility of the University of Parma.

\section*{Data Availability}

The figure files and code to reproduce the results of this paper is available at \cite{zenodo}.

\twocolumngrid
%

\onecolumngrid
\newpage
\appendix


\renewcommand{\thefigure}{S\arabic{figure}}
\renewcommand{\theHfigure}{S\arabic{figure}}
\setcounter{figure}{0}

\section{Non-central equivalent Wishart Ansatz} \label{app:noncentral_EWA}

\subsection{Contractions of the non-central Wishart distribution and definition of $\tilde{Q}_\ell$}\label{app:noncentral_contractions_and_tildeQ}

The non-central Wishart distribution is obtained by allowing a non-zero mean in the Gaussian vectors entering the Wishart construction. Let $G$ be a $P \times N$ matrix, whose columns $G_i$ ($i=1,\dots,N$) are independent and distributed as
\begin{equation}
    G_i \sim \mathcal N_P(\mu_i,V)\,,
\end{equation}
with common covariance matrix $V$ and possibly non-zero means $\mu_i \in \mathbb R^P$. Denoting by $M = (\mu_1,\dots,\mu_N) \in \mathbb R^{P\times N}$ the corresponding mean matrix, the random matrix $G G^\top$ is said to follow a non-central Wishart distribution, which we denote by
\begin{equation}
    R = G G^\top \sim \mathcal W^\mathrm{nc}_P(M M^\top,V,N)\,,
\end{equation}
with non-centrality encoded by $M M^\top$. For $M=0$ one recovers the ordinary central Wishart distribution. Another convention frequently used in the literature is to define the non-centrality matrix as $\Omega = V^{-1} M M^\top$. 

The only property of the non-central Wishart distribution that we will need is the analogue of Eq.~\eqref{eq:propertychi}. Given a fixed vector $s \in \mathbb R^P$, one has
\begin{equation}
    \frac{s^\top R s}{s^\top V s}
    =
    \sum_{i=1}^N (z_i)^2,
    \quad \quad \mathrm{where}\; z_i =\frac{s^\top G_i}{\sqrt{s^\top V s}}\,
    \sim \mathcal N\Big(\frac{s^\top \mu_i}{\sqrt{s^\top V s}}, 1\Big).
\end{equation}
Since the projected variables $z_i$ are independent Gaussian with unit variance, it follows that the normalized contraction is distributed as a non-central chi-squared random variable:
\begin{equation}
    \frac{s^\top R s}{s^\top V s}
    \sim
    \chi_\mathrm{nc}^{2}(N, N\lambda_s)\,,
    \quad \quad \mathrm{where}\;
    N \lambda_s
    =
    \sum_{i=1}^N \frac{(s^\top \mu_i)^2}{s^\top V s}
    =
    \frac{s^\top M M^\top s}{s^\top V s}\,. \label{eq:property_chi_noncentral}
\end{equation}
In the special case relevant to our setting, in which all columns have the same mean $\mu_i=m$, the non-centrality parameter reduces to
\begin{equation}
    N \lambda_s
    =
    N\,\frac{(s^\top m)^2}{s^\top V s}\,.
\end{equation}

\subsubsection*{Definition of $Q_\ell$ variables in the non-central EWA}

Analogously to Eq.~\eqref{eq:iterated_chi2_intoduction} we can introduce the variables $Q_\ell$
\begin{align}
    e^{-\frac{1}{2} \bar{f}^\top K_E^{(L)} \bar{f}} 
    &=  \exp\bigg[  -\frac{1}{2} 
                    \underbrace{
                    \frac{\bar{f}^\top K_E^{(L)} \bar{f}}
                         {\bar{f}^\top \Theta(K_E^{(L-1)}) \bar{f}}\;
                         }_{\coloneq Q_L/N_L}
                    \bar f^\top \Theta(K_E^{(L-1)}) \bar f
            \bigg] \nonumber \\
    =\dots&= \exp\left[  -\frac{1}{2} 
                    \bigg(\prod_{\ell=1}^L \frac{Q_\ell}{N_\ell} \bigg)
                    \bar{f}^\top \Theta^L(C_X) \bar{f}
            \right].
\end{align}
For brevity we now use the short hands $\Sigma_{(\ell)},m_{(\ell)}$ and $\Theta_{(\ell)}$ instead of $\Sigma\big(\Theta^{L-\ell-1}(K^{\ell-1}_E)\big)$, $m\big(\Theta^{L-\ell-1}(K^{\ell-1}_E)\big)$ and $\Theta^{L-\ell}\big(K^{(\ell-1)}_E\big)$. 
Given the non-central EWA
\begin{equation}
    \Theta^{L-\ell}\big(K_E^{(\ell)}\big) | K_E^{(\ell-1)} \sim \mathcal W^\mathrm{nc}_P \left(N_\ell\, m_{(\ell)} , \Sigma_{(\ell)}, N_\ell \right)
\end{equation}
the property Eq.~\eqref{eq:property_chi_noncentral} holds for contractions normalized by the scale matrix, which for non-zero means $m_{(\ell)}$ is the covariance kernel $\Sigma_{(\ell)}$ and not the second moment $\Theta_{(\ell)} = \Sigma_{(\ell)} + m_{(\ell)} m_{(\ell)}^\top$ appearing in the definition of $Q_\ell$.
This is why we introduce the related variables $\tilde Q$ in the non-central case
\begin{align}
\label{eq:tildeQ}
    \tilde Q_\ell  
            &\coloneq \frac{\bar{f}^\top \Theta^{L-\ell}\big(K_E^{(\ell)}\big) \bar{f}}
                  {\frac{1}{N_\ell}\bar{f}^\top \Sigma_{(\ell)} \bar{f}} \nonumber \\
            &= \frac{\bar{f}^\top \big(\Sigma_{(\ell)} + m_{(\ell)} m_{(\ell)}^\top \big)\bar{f}}
                  {\bar{f}^\top \Sigma_{(\ell)} \bar{f}} \;
              \frac{\bar{f}^\top \Theta^{L-\ell}\big(K_E^{(\ell)}\big) \bar{f}}
                  {\frac{1}{N_\ell}\bar{f}^\top \Theta_{(\ell)} \bar{f}} \nonumber \\
            &= \big(1 + \lambda_{\mathrm{nc},\ell} \big) Q_{\ell}
\end{align}
which according to Eq.~\eqref{eq:property_chi_noncentral} follow 
$\tilde Q_{\ell} \sim \chi_\mathrm{nc}^2\big(N_{\ell}, N_{\ell}\,\lambda_{\mathrm{nc},\ell}\big)$ with non-centrality $N_{\ell}\,\lambda_{\mathrm{nc},\ell} = N_{\ell}\,\frac{\bar{f}^\top m_{(\ell)} m_{(\ell)}^\top \bar{f}}{\bar{f}^\top \Sigma_{(\ell)} \bar{f}}$.

With these definitions one can proceed with the arguments given in Section~\ref{sec:noncentral_EWA} to show the asymptotic equivalence of central and non-central EWA. Note however, that if one wanted to write out an explicit effective action in the $L$-layer non-central case, this would require to track dependencies between the $\tilde Q_\ell$ (or equivalently $Q_\ell$) distributions, which through $\lambda_{\mathrm{nc},\ell}$ now explicitly depend on $\bar f$ and $m_{(\ell)}$ and therefore require additional order parameters to decouple, unlike in the central case where the $Q_\ell$ distributions are independent and identical.

\subsection{Non-central EWA for one hidden layer}
\label{sec:EWA_nc_1HL}

For one hidden layer, the action arising from a non-central EWA can be written out without too much complication. This appendix analyses the relation to the central EWA result, showing how task-kernel and task-kernel-mean overlaps control the saddle point solution, and how the distributions of $Q$ and $Q_\mathrm{nc}$ differ while such a difference is absent in the output distribution.

We derive the action by first computing the characteristic function of the output prior Eq.~\eqref{eqprioralternative} for a one hidden layer network under the non-central EWA, with the aim to plug into Eq.~\eqref{eq:Z_from_priorcharfunc} and obtain the posterior partition function by a Gaussian integration.
Recall the definitions of the kernels $C = \frac{1}{N_0 \lambda_0} X X^\top$ and $K^{\mu \nu}_\mathrm{E} = \frac{1}{N_1 \lambda_1}\sum_i^{N_1} \sigma_i^\mu \sigma_i^\nu$ and $\Theta(C) = (\Sigma + m m^\top) / \lambda_1$; with hidden layer activations $\sigma_i^\mu$ i.i.d. across $i$, of mean $m^\mu = \mathbb{E}[\sigma^\mu_i]$ and covariance $\Sigma^{\mu \nu} = \mathrm{Cov}[\sigma^\mu_i \sigma^\nu_i]$.
The characteristic function of the output prior Eq.~\eqref{eqprioralternative} is
\begin{align}
    \varphi (\bar f| X) 
    & = \int_{\mathcal S^+_P} dK_{\mathrm{E}} \rho (K_{\mathrm{E}}|C) e^{-\frac{1}{2}\bar f^\top K_E\bar f}
      = \mathbb{E}\left[ e^{-\frac{1}{2}\bar f^\top K_E\bar f} \right] \\
    & = \mathbb{E}\bigg[ \exp{\bigg(
                            -\frac{1}{2}
                            \underbrace{\frac{\bar f^\top K_\mathrm{E}\bar f}
                                 {\frac{1}{N_1 \lambda_1} \bar f^\top \Sigma \bar f}}
                                 _{\tilde Q \,\sim \,\chi^2_\mathrm{nc}(N_1, N_1 \lambda_\mathrm{nc})}
                            \;\frac{1}{N_1 \lambda_1} \bar f^\top \Sigma \bar f
                            \bigg)}\bigg]
\end{align}
Here $\tilde Q \,\sim \,\chi^2_\mathrm{nc}(N_1, N_1 \lambda_\mathrm{nc})$ with non-centrality parameter $\lambda_\mathrm{nc} = \frac{\bar f^\top m m^\top \bar f}{\bar f^\top \Sigma \bar f}$ under the non-central EWA. This can be seen by decomposing $\tilde Q = \sum_i a_i a_i$ with $a_i = (\bar f^\top \sigma_i) \big/ \sqrt{\bar f^\top \Sigma \bar f}$ and noting that $\mathbb E[a_i] = \bar f^\top m \big/ \sqrt{\bar f^\top \Sigma \bar f}$.

Now expressing the distribution $\rho(\tilde Q) = \int \frac{d\tilde{\bar Q}}{2\pi} \varphi(\tilde{\bar Q}) e^{-i\tilde{\bar Q} \tilde{Q}}$ via its characteristic function
\begin{equation}
\varphi (\tilde{Q}) = \big(1 - 2i \tilde{\bar Q}\big)^{-\frac{N}{2}} \exp \bigg(\frac{iN_1 \lambda_\mathrm{nc} \tilde{\bar Q}}{1 - 2i \tilde{\bar Q}} \bigg), \nonumber
\end{equation}
we find with $\lambda_\mathrm{nc} = \frac{\bar f^\top m m^\top \bar f}{\bar f^\top \Sigma \bar f}$ 
\begin{align}
    \varphi (\bar f| X) 
    & = \int \frac{d\tilde{\bar Q} d\tilde{Q}}{2\pi}  
        \big(1 - 2i \tilde{\bar Q}\big)^{-\frac{N}{2}} 
        \exp \bigg[
                    \frac{i}{1 - 2i \tilde{\bar Q}} \frac{N_1 \tilde{\bar Q}}{\bar f^\top \Sigma \bar f} \bar f^\top m m^\top \bar f 
                    + i \tilde Q \left( -\tilde{\bar Q} + \frac{i}{2 N_1 \lambda_1} \bar f^\top \Sigma \bar f \right)
                    \bigg].
\end{align}
Noting that the second term in the exponent gives precise meaning to $\tilde{\bar Q}$, as 
$\int d\tilde Q e^{i \tilde Q \left( -\tilde{\bar Q} + \frac{i}{2 N_1 \lambda_1} \bar f^\top \Sigma \bar f \right)} = \delta (\tilde{\bar Q} - \frac{i}{2 N_1 \lambda_1} \bar f^\top \Sigma \bar f) $, we can cancel $\frac{N_1 \tilde{\bar Q}}{\bar f^\top \Sigma \bar f} = \frac{i}{2 \lambda_1}$ in the first term of the exponent 
- this leaves the desired quadratic function of $\bar f$, and with the substitutions $- 2i \tilde{\bar Q} = \bar Q$ and $\tilde Q / N = Q$
\begin{align}
    \varphi (\bar f| X) 
    & = \int \frac{d\bar Q d\tilde{Q}}{2\pi}  
        \big(1 - 2i \tilde{\bar Q}\big)^{-\frac{N}{2}} 
        \exp \bigg[
                    - \frac{1}{2 \lambda_1}\frac{1}{1 - 2i \tilde{\bar Q}}  \bar f^\top m m^\top \bar f 
                    - i \tilde Q \tilde{\bar Q} 
                    - \frac{\tilde Q}{2 N_1 \lambda_1} \bar f^\top \Sigma \bar f
                    \bigg] \\
    & \propto \int d\bar Q d\tilde{Q}  
        \big(1 + \bar Q\big)^{-\frac{N}{2}} e^{\frac{N_1}{2} Q \bar Q }\,
        \exp \bigg[
                    - \frac{1}{2} 
                    \bar f^\top \bigg(
                        Q\, \Theta(C) + 
                        \big( \frac{1}{1 + \bar Q} - Q \big)  \frac{m m^\top}{\lambda_1} 
                    \bigg) \bar f
                    \bigg] \\
    & \propto \int d\bar Q d\tilde{Q}  
        \exp \bigg[
                    \frac{N_1}{2} Q \bar Q 
                    - \frac{N_1}{2} \log (1 + \bar Q)
                    - \frac{1}{2} 
                    \bar f^\top \big( \Id \beta^{-1} + K^R_{\mathrm{nc}}(Q,\bar Q)\big) \bar f
                    \bigg]. \label{eq:app_charfunc_prior_1hl_noncentral}
\end{align}
In the last line we defined the renormalized kernel for the non-central EWA, 
\begin{align}
K^R_{\mathrm{nc}} (Q,\bar Q)
&= K^R (Q) + \frac{1}{\lambda_1} \bigg( \frac{1}{1 + \bar Q} - Q \bigg)  m m^\top \\
&=  Q\, \Theta(C) + \frac{1}{\lambda_1} \bigg( \frac{1}{1 + \bar Q} - Q \bigg)  m m^\top. 
  \label{eq:renormalized_noncentral_1hl_kernel}
\end{align}
Finally, plugging into the expression for the partition function of the posterior Eq.~\eqref{eq:Z_from_priorcharfunc} and performing the Gaussian integration, we find 
$Z_\mathrm{nc} \propto \int d\bar Q d\tilde{Q}\, e^{- \frac{N_1}{2} S[Q,\bar Q]}$ with the effective action
\begin{equation}
S_\mathrm{nc} \big[Q,\bar{{Q}}\big] =
-Q\bar{Q}
+\log\left(1+ \bar Q\right)
+\frac{{\alpha}}{P} \log\det\left[ \Id \beta^{-1} + K_{\mathrm{nc}}^{R} (Q,\bar Q )\right]
+\frac{{\alpha}}{P} y^{\top} \left(\Id \beta^{-1} +  K_{\mathrm{nc}}^{R} (Q,\bar Q )\right)^{-1}y. \label{eq:1hl-raw-noncentral-action}
\end{equation}
This recovers the one hidden layer action for non-zero mean activations in Suppl. Sect.~IV of \cite{pacelli2023statistical}, here obtained via the route of a non-central EWA. By analyzing this expression further we here show explicitly how the difference to the central EWA is small, as argued generally in Section~\ref{sec:noncentral_EWA}.  

Using the Sherman--Morrison formula
\begin{equation}
\left(A+uv^{\tr}\right)^{-1}=A^{-1}-\frac{A^{-1}uv^{\tr}A^{-1}}{1+v^{\tr}A^{-1}u}
\end{equation}
and the Matrix determinant lemma
\begin{equation}
\det\left(A+uv^{\tr}\right)=\left(1+v^{\tr}A^{-1}u\right)\det\left(A\right),
\end{equation}
\eqref{eq:1hl-raw-noncentral-action} can be expressed in the more lengthy but better interpretable form written only in terms of the kernel $K^R_\beta(Q)$ appearing in the central EWA:
\begin{align}
S_\mathrm{nc} \left[Q,\bar{Q}\right] 
=& -Q\bar{Q} + \log\left(1+\bar{Q}\right)
   +\frac{{\alpha}}{P} \Tr\log \left[\Id \beta^{-1} + K^R\right]
   +\frac{{\alpha}}{P} y^{\top}\left(\Id \beta^{-1} + K^R\right)^{-1} y \nonumber \\
 & +\frac{{\alpha}}{P} \log \left[  
                1 
                - \frac{{1}}{\lambda_{1}}
                  \left(Q-\frac{{1}}{1+\bar{Q}}\right)
                  m^{\top}\left(\Id \beta^{-1} + K^R\right)^{-1}m
                \right] \label{eq:noncentral_1hl_action_expanded}\\ 
 & +\frac{{\alpha}}{P} 
    y^{\tr}\left[
        \frac{{1}}{\lambda_{1}} \left(Q-\frac{{1}}{1+\bar{Q}}\right)
        \frac{\left(\Id \beta^{-1} + K^R\right)^{-1} mm^{\top} \left(\Id \beta^{-1} + K^R\right)^{-1}}
             {1 - \frac{{1}}{\lambda_{1}} \left(Q-\frac{1}{1 + \bar{Q}}\right)
                                          m^{\top}\left(\Id \beta^{-1} + K^R\right)^{-1} m}
        \right]y \nonumber .
\end{align}
Here the first line is equivalent to the action Eq.~\eqref{eq:action_singleoutput} obtained from the central EWA, as seen by plugging in the explicit saddle-point relation 
$ \frac{1}{Q^{\ast,c}} - 1 = \bar{Q}^{\ast,c}$ which follows directly from $\partial S / \partial \bar{Q} \overset{!}{=} 0$ when setting the second and third line to zero. The second and third line thus represent the difference between central and non-central EWA. Note that at the saddle-point of the central EWA, the quantity $\Delta_Q \coloneq Q-\frac{{1}}{1+\bar{Q}} \overset{\ast,c}{=} 0$ causing the second and third lines to vanish. While the saddle point of the non-central action Eq.~\eqref{eq:noncentral_1hl_action_expanded} can give rise to $\Delta_Q \neq 0$ at finite $N,P$, we find that asymptotically $\Delta_{Q^\ast} \overset{N,P\to\infty}{\to} 0$.

\subsubsection*{Zero-temperature limit}

For ease of exposition, we now focus on the zero-temperature limit where the structure of this action becomes particularly clear.
Setting $\beta \to 0$, we can define the task-dependent (but not $Q, \bar{Q}$ dependent) scalar overlaps
\begin{align}
    M_{yy} &= \frac{1}{P} y^{\tr} \Theta(C)^{-1} y \label{eq:app_Myy}\\
    M_{m y} &= \frac{1}{\sqrt{P}} m^{\tr} \Theta(C)^{-1} y \\
    M_{mm} &= m^{\tr} \Theta(C)^{-1} m \\
    \Gamma_K &= \frac{1}{P} \log [\det \Theta(C)]. \label{eq:app_GammaK}
\end{align}
With the short-hand $\Delta_{Q} = Q-\frac{1}{1 + \bar Q}$, the action becomes
\begin{align}
    S[Q,\bar Q] =  &- Q \bar Q 
                     + \log\left(1+\bar Q\right) 
                     + \alpha \log(Q/\lambda_1)
                     + \alpha \Gamma_K 
                     + \alpha\frac{\lambda_1}{Q} M_{yy}  \nonumber \\
                    &+ \frac{\alpha}{P} \log \left( 1 - \frac{\Delta_Q}{Q} M_{m m}  \right) 
                     + \alpha \lambda_1 \frac{\Delta_Q}
                                             {Q \left(Q - \Delta_Q \;M_{m m} \right)}
                                        M_{m y}^2.  \label{eq:noncental_1hl_action_zerotemp}
\end{align}
Again, the first line corresponds to the central EWA result. Note that the 
saddle-point of the zero temperature, central EWA action only depends on $\alpha$ and the fixed task-to-kernel overlap $M_{yy}$, permitting a simple understanding of the saddle-point solution $Q^\ast$, see App.~\ref{app:zerotemp_saddle_behavior}.

\subsubsection*{Scaling behavior}
The starting observation to analyze the relative scaling and contribution of the $M_{yy},M_{\mu y},M_{\mu \mu}$ overlaps is that the NNGP kernel $\Theta(C)$ corresponds \deleted[id=rev1]{to} to a zero-mean activation kernel with rank-1 spike
$\Theta(C) = \Sigma + m m^\top.$
Since $m^\mu =O(1)$, while the BBP transition happens at the $O\big(1/\sqrt{P}\big)$ scale, the spike always creates a strong outlier Eigenvalue $\lambda_0=O(P)$ as long as $\Sigma$ does not inherit an equally strong low-rank structure from the data distribution.
Diagonalizing the kernel into the Eigenvalue-Eigenvector pairs $\{\lambda_i, v_i\}_{i=1..P}$ 
we can define the projections of $m$ and $y$ on the Eigenspaces, $\hat{m}_i = m^\top v_i$ and $\hat{y}_i = y^\top v_i$. When inverting the kernel, the Eigenvectors stay the same and the Eigenvalues are simply inverted, giving the pairs 
$\{\lambda_i^{-1}, v_i\}_{i=1..P}$.

Therefore, we can understand \deleted[id=rev1]{the} the scaling of $M_{yy},M_{\mu y},M_{\mu \mu}$ through the behavior of the kernel spectrum and the alignment of the Eigenspaces with $m,y$. In particular, $v_0$ is typically highly localized to $m$ and the outlier eigenvalue $\lambda_0^{-1}$ becomes the smallest eigenvalue of the inverse kernel, which instead is dominated by the tail of the spectrum. The exact scaling indeed still depends non-trivially on the data-dependent decay of the spectrum and the relative overlaps of $m,y$ with the corresponding Eigenspaces. To avoid a lengthy discussion we rely on the general argument of asymptotic equivalence to the central EWA presented in Section~\ref{sec:noncentral_EWA}, and here restrict ourselves to present numerical results on the behavior of $M_{yy},M_{\mu y},M_{\mu \mu}$ and $\Delta_{Q^\ast}$ in Appendix~\ref{appsec:numerical_evidence_noncentral} and a few observations:

\begin{itemize}
    \item For real data distributions such as the images of CIFAR-10, the dimensionality of the ground-truth data manifold is fixed as the number of samples $P$ taken from the distribution increases. A well known consequence is that the Eigenvalues of the kernel at fixed spectral index $i$ grow as $\lambda_i \sim P$, while in relative terms the new eigenvalues added in the tail of the spectrum as $P$ increases, assuming the modes at $i=O(P)$ still approximate the population spectrum,  typically follow a power-law decay $\lambda_{O(P)} /P \sim P^{-\gamma}$ with kernel- and data-dependent exponent $\gamma > 1$. Such a power law decay holds for the population spectrum of finite-smoothness kernels such as the ReLU NNGP \cite{Bietti2021,Birman1977}, while for analytic kernels the decay may be faster. Therefore, the tail comes to dominate the inverse spectrum more and more, e.g. here with the mean $\frac{1}{P}\sum_i\lambda_i^{-1} \sim P^{\gamma -1}$.
    \item Correspondingly, if the label vector $y$ projects uniformly on the Eigenspaces of the kernel, also $M_{yy}$ grows as $\sim P^{\gamma - 1}$. For easier tasks where $y$ is largely localized to the upper bulk of the spectrum, $M_{yy}=O(1)$.
    \item The logarithmic term $\frac{\alpha}{P} \log \left( 1 - \frac{\Delta_Q}{Q} M_{m m}  \right)$ trivially vanishes for $P\to \infty$ as long as $\frac{\Delta_Q}{Q} M_{m m} \ll 1$ to avoid the divergence. This behavior is encouraged both since $M_{m m}\overset{P\to\infty}{\to}0$ if $v_0$ is mostly localized to $m$, and since the saddle-point can be thought of as a perturbed version of the central EWA saddle located at $\Delta_{Q^{\ast,c}}=0$.
    \item $M^2_{my}$ which controls the last term in Eq.~\eqref{eq:noncental_1hl_action_zerotemp} is determined by the overlap of the projections of $m$ and $y$ on the Eigenspaces. The contributions of each $\hat m_i \lambda_i^{-1} \hat y_i$ can be positive or negative and therefore partly cancel. If $m||y$, then  $M_{my} \propto \frac{1}{\sqrt{P}} M_{mm}$ vanishes at a faster rate than $M_{mm}$. 
    A fast decay exponent $\gamma$ of the tail spectrum can lead to slowly decaying or even constant $M_{my}$, however the same mechanism increases also $M_{yy}$, such that the hierarchy between the two terms remains.
\end{itemize}

\begin{figure}
    \centering
    \includegraphics[width=\linewidth]{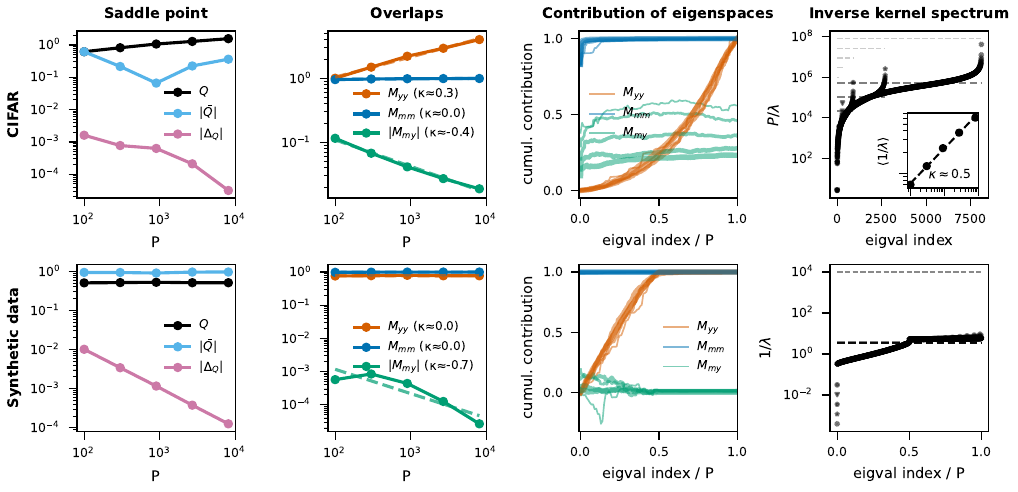}
    \caption{
    Analysis of the non-central EWA action Eq.~\eqref{eq:noncental_1hl_action_zerotemp} for one hidden layer ReLU networks, varying the dataset size $P,N$ at fixed $\alpha$. (First row) Results for regression on CIFAR-10 classes \{0,1\} with fixed input dimension $N_0=784$ while the hidden layer scales as $N=P/\alpha$ with $\alpha=2$. (Second row) Results for Gaussian data linear labels $y=w^\ast X$, here both $\alpha_0=\alpha =2$. (First column) Behavior of the saddle-point solution $Q^\ast,\bar Q^\ast$ and resulting $\Delta_Q$ which indicates the deviation from the central EWA solution. (Second column) The data-dependent overlaps $M_{yy},M_{my},M_{mm}$ determining the action. Exponents $\kappa$ of power-law fits to each line are shown in the legend and indicated by dashed lines plotted below the empirical results. (Third \replaced[id=rev1]{column}{row}) Shows by descending Eigenvalue order the contribution of $\Theta(C)$ eigen-spaces to the three overlaps, e.g. $M_{my}=\frac{1}{\sqrt{P}}\sum_i \lambda_i^{-1} m^\top v_i\; y^\top v_i$. (Fourth \replaced[id=rev1]{column}{row}) $\lambda_i^{-1}$ constituting the spectrum of the inverse kernel $(\Theta(C) + \epsilon\Id)^{-1}$. Different markers indicate the spectrum for $P=[100, 300, 900, 2700, 8100]$. For CIFAR with fixed $N_0$ the spectra are rescaled by $P$ to show collapse of the shared part of the spectra; this is not needed for the synthetic data with $\alpha_0$. Black dashed lines indicate average values, gray dashed lines the ceiling due to regularization by $\epsilon=10^{-4}$. The inset shows the means $(\sum_i \lambda_i^{-1}) /P $ for each $P$ and corresponding power-law fit with exponent $\kappa$ (dashed line).
    }
    \label{Sfig:noncentral_Qs_and_Ms_analysis}
\end{figure}

\subsection{Numerical evidence of the equivalence of central and non-central approaches} \label{appsec:numerical_evidence_noncentral}

The behaviors discussed above are shown numerically in Fig.~\ref{Sfig:noncentral_Qs_and_Ms_analysis} for the CIFAR-10 $\{0,1\}$ and Gaussian data tasks used in the main text. For MNIST, and synthetic datasets with class imbalance or with labels $y$ correlated to the mean vector $m$, we have found qualitatively similar results (not shown).
In all cases, and as following from the general arguments presented in the main text Section~\ref{sec:noncentral_EWA}, the quantity $\Delta_Q$ controlling any differences between central and non-central theories vanishes with increasing system size $P,N$. For the synthetic Gaussian task this happens as $\Delta_Q\sim P^{-1}$. The second column of Fig.~\ref{Sfig:noncentral_Qs_and_Ms_analysis} shows this is due to similarly vanishing $|M_{my}|$.

As seen in the third column panels, the majority contribution to $M_{mm}$ is due to the outlier eigenvalue. This shows that also in the fixed data dimension case of CIFAR, where not only the mean-associated outlier but all eigenvalues grow proportionally with $P$, the outlier eigenvector $v_0$ is largely localized and $m$ has non-vanishing overlaps only with a few top eigenvectors. $M_{my}$ also shows a significant contribution from the outlier eigenvalue, and while the contributions of remaining directions are of significant size they have (for these choices of task and kernel) almost random signs, causing these contributions to cancel. To visualize the size of this cancellation, the green lines in the third column were normalized by the total size of unsigned contributions. Finally, the contributions to $M_{yy}$ are more evenly spread across the full spectrum. This is task dependent but expected of a regime where adding data samples increases the accuracy of the kernel predictor.

$M_{yy}$ can be growing with $P$, as in the case of CIFAR where numerically $M_{yy}\sim P^{0.3}$. This growth is due to the increasing weight of the tail eigenvalues; the last 30\% of eigenvalues for each $P$ contribute about half of the size of $M_{yy}$ for CIFAR (orange lines in third column), while the mean value of the inverse spectrum increases with $P$ (inset of fourth column).

Lastly, as expected the inverse NNGP spectrum of the Gaussian data (bottom right panel) is composed of three easily interpretable parts: The outlier eigenvalue due to the non-zero mean scaling with $P$, and two bulks of eigenvalues. These arise from the fact that $C=XX^\top$ is Wishart and not of full rank since $P=2N$, creating two bulks of $N$ nonzero and $N$ zero eigenvalues. Trivially the linear label vector $y=w^\ast X$ can only have overlap with the nonzero part of the spectrum, which explains the behavior of the $M_{yy}$ and $M_{my}$ contributions in the third column. The ReLU NNGP kernel function $\Theta$, having an infinite-dimensional associated reproducing kernel Hilbert space (RKHS), ensures that $\Theta(C)$ is full-rank such that also the second bulk of eigenvalues is non-zero, independently of small regularization (compare separation of gray dashed line and inverse spectrum).

\subsection{\added[id=rev1]{Derivation of Eq.~\eqref{eq:small_noncentrality}: self-averaging and suppression of the non-centrality parameter}} \label{app:selfaveraging-noncentrality}

\added[id=rev1]{In this appendix we show the self-averaging and smallness of $\lambda_{\mathrm{nc}}(\bar f)=\frac{(\bar{f}^\top m)^2}{\bar{f}^\top \Sigma \bar{f}}$ under the $\bar f$ measure, used in Sec.~\ref{sec:noncentral_EWA} to explain why the non-central EWA can asymptotically be replaced by the central EWA for generic depth.}
Given that the numerator $(\bar f^\top m)^2$ only depends on a single projection $m$ and this direction contributes negligibly to $\bar f^\top \Sigma \bar f$, self-averaging of the numerator and denominator can be assessed separately. 
In the numerator, $(\bar f^\top m)^2 =O(1)$ for typical $\bar f \sim \mathcal N(\mu, K_\beta)$ due to the small inverse outlier eigenvalue $\sim 1/\| m\|^{2}$ asymptotically localized to the $m$ direction. 
A convenient scalar controlling the typical size of the denominator is the effective dimension
\begin{equation}
    t \coloneq \Tr\left[\Sigma \, K_\beta \right]
       = \Tr\left[ \Sigma \left(\beta^{-1}\Id + Q(\Sigma + m m^\top)   \right)^{-1}\right].
    \label{eq:teff_def}
\end{equation}
\added[id=rev1]{To see this, consider again that }in the noncentral EWA one encounters $\bar f $ integrals of the form
\begin{equation}
\int d\bar f\; \exp\!\Big(-\tfrac12 \bar f^\top K_\beta^{-1}\,\bar f + i y^\top \bar f\Big)\,(\cdots),
\qquad K_\beta=(\beta^{-1}I+Q\,\Theta)^{-1},
\label{eq:appteff_shifted_gaussian}
\end{equation}
with $\Theta=\Sigma+mm^\top$ and $Q>0$. Completing the square gives
\begin{equation}
-\tfrac12 \bar f^\top K_\beta^{-1} \bar f+i y^\top \bar f
=
-\tfrac12(\bar f-\mu)^\top K_\beta^{-1} (\bar f-\mu)+\tfrac12\,y^\top K_\beta \,y,
\qquad \mu\coloneq i K_\beta \, y.
\end{equation}
Thus the $\bar f$-dependence in Eq.~\eqref{eq:appteff_shifted_gaussian} is that of a complex-mean Gaussian with covariance $K_\beta$. In particular, fluctuations of quadratic forms such as $\bar f^\top\Sigma\bar f$ are then governed by $K_\beta$, while the linear term only adds the deterministic shift $\mu$.
Write $\bar f=\mu+g$ with $g\sim\mathcal N(0,K_\beta)$. Then
\begin{equation}
D=\bar f^\top\Sigma\bar f
= g^\top\Sigma g + 2\mu^\top\Sigma g + \mu^\top\Sigma\mu
\label{eq:appteff_D_decomp}
\end{equation}
and mean and variance of the pure fluctuation term are 
\begin{align}
\mathbb E[g^\top\Sigma g] &= \Tr(\Sigma K_\beta) \equiv t, \\
\mathrm{Var}[g^\top\Sigma g] &= 2\,\Tr\!\big((\Sigma K_\beta)^2\big).
\end{align}
Furthermore, $\Sigma$ and $K_\beta^{-1}=\beta^{-1}I+Q\,(\Sigma + m m^\top)$ are positive semi-definite and the eigenvalues of $\Sigma K_\beta$ are $\in(0,Q^{-1}]$, so that
\begin{equation}
\Tr\!\big((\Sigma K_\beta)^2\big)\le Q^{2}\Tr(\Sigma K_\beta) = Q^2 t
\qquad\Rightarrow\qquad
\frac{\sqrt{\mathrm{Var}[g^\top\Sigma g]}}{\mathbb E[g^\top\Sigma g]}
\le \sqrt{\frac{2}{t}}Q.
\end{equation}
Hence whenever $t=\Tr(\Sigma K_\beta)$ diverges, $g^\top\Sigma g$ is self-averaging.
\replaced[id=rev1]{We will argue below}{In the main text, it is argued} that $t=O(P)$ and therefore relative fluctuations are $O(P^{-1/2})$.

The mean-shift related terms in Eq.~\eqref{eq:appteff_D_decomp} do not change this behavior: $2\mu^\top\Sigma g$ is linear in $g$ and has variance
$4\,\mu^\top\Sigma K_\beta\Sigma\mu$, while $\mu^\top\Sigma\mu$ is deterministic. For the  $\|y\|^2=O(P)$ scaling relevant here, these contributions are at most $O(P)$ even if $y$ correlates significantly with the mean direction $m$, and therefore do not affect the fact that $D$ is extensive and concentrates around its mean on relative scale $P^{-1/2}$.

\replaced[id=rev1]{Therefore,}{A short calculation in Appendix~\ref{app:noncentral_teff} shows that} $\bar f^\top \Sigma \bar f = O(t)$ with relative fluctuations $\sim t^{\frac12}$, and is \deleted[id=rev1]{therefore} self-averaging if $t = O(P)$.
Indeed, in Eq.~\eqref{eq:teff_def} at low temperature $\beta^{-1}\ll1$ and aside from the negligible $m$ direction, $\Sigma$ and $K_\beta$ admit almost the same diagonalization. Along an eigenmode with $\Sigma$-eigenvalue $\lambda_i$ and corresponding $K_\beta$-eigenvalue $[\beta^{-1}+Q\lambda_i]^{-1}$, the contribution to $t$ is then
$\frac{\lambda_i}{\beta^{-1}+Q\lambda_i} z_i^2$ with $z_i$ standard normal. Hence for the extensive number of $\lambda_i \gtrsim \beta^{-1}$ modes one has $\lambda_i(\beta^{-1}+Q\lambda_i)^{-1}=O(1)$ and their contributions add up to $t \sim P,$ with relative fluctuations of order $P^{-1/2}$. In particular, the denominator $\bar f^\top \Sigma \bar f$ then concentrates away from zero, so $1/(\bar{f}^\top \Sigma \bar f)$ is also self-averaging and $\langle1/(\bar{f}^\top \Sigma \bar f)\rangle_{\bar f} \sim 1/t=O(1/P)$.
\added[id=rev1]{We note that the relevant spectrum for this argument is that of the nonlinear activation covariance, whose rank is not limited by the input dimensionality, even when that dimensionality is small and fixed. Nonetheless, particular combinations of low-rank data distributions and activation functions can violate the $t = O(P)$ condition due to rapidly decaying spectral tails.}
Combining numerator and denominator we have for the dominant $\bar f$ configurations $\lambda_\mathrm{nc} = O(1/P)$ and correspondingly \added[id=rev1]{Eq.~\eqref{eq:small_noncentrality} follows from}
\begin{equation}
    \rho_{\chi^2_{\mathrm{nc}}}\!\Big(\tilde Q_{\ell};N_\ell, N_\ell\lambda_{\mathrm{nc}}\Big)
    =
    \rho_{\chi^2}\!\big(Q_{\ell};N_\ell\big)\Big(1+O(1/P)\Big).
\end{equation}

\begin{figure*}
    \includegraphics[width=\textwidth]{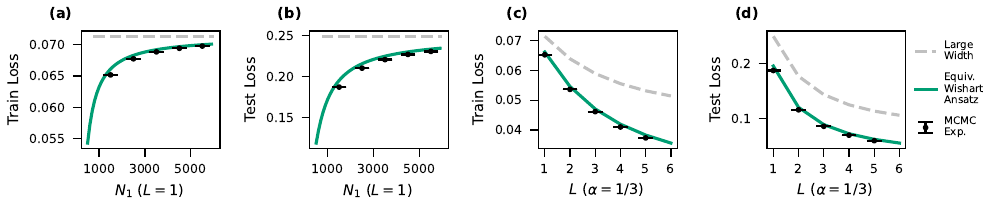} 
    \caption{\newtext{Comparison between the learning curves obtained via the Equivalent Wishart Ansatz and numerical experiments on synthetic data with an equicorrelated kernel and labels aligned with the spike direction. Numerical samples from the Bayesian posterior (black dots) are compared against the large-width limit predictions (gray dashed lines) and the results of the EWA theory (green solid lines). Panels (a) and (b) display the training loss and the generalization error of a one-hidden-layer network ($L=1$) as a function of the width $N_1$; panels (c) and (d) display the same quantities as a function of the number of hidden layers $L$, at fixed width $N_\ell=1500$ $\forall \ell\ge 1$, i.e., $\alpha=P/N_\ell=1/3$. We keep the input dimensionality and the number of patterns fixed at $N_0=1000$ and $P=500$, and the equicorrelation parameter at $\rho=0.5$. These simulations refer to the Erf activation function, with Gaussian priors $\lambda=1$ and temperature $T=0.1$. For all panels, we sample from the posterior using Langevin Monte Carlo with a learning rate $\eta=0.001$ and use $P_t=500$ test samples.}}
    \label{fig:C-with-spike}
\end{figure*}

\newtext{
\subsection{The equicorrelated dataset: an instructive example}
\label{app:equicorrelated}

The contraction property, Eq.~\eqref{eq:propertychi}, is exact for Wishart matrices and is therefore built into the EWA. For the true kernel ensemble it is an approximation whose accuracy depends on the data and also on the direction $\bar f$. An instructive example is provided by data with an equicorrelated Gram matrix,
\begin{equation}
C=(1-\rho)\,\Id+\rho\,\mathbf{1}_P\mathbf{1}_P^\top,\qquad 0<\rho<1,
\end{equation}
for which the pre-activations of each hidden neuron can be written as
\begin{equation}
    h^\mu=\sqrt{\rho}\,g+\sqrt{1-\rho}\,z^\mu, \quad\quad g,z^1,\ldots,z^P\sim\mathcal{N}(0,1)\, .
\end{equation}
Consider a one-hidden-layer network with Erf activation and unit prior variances, and the direction $v_0=\mathbf{1}_P/\sqrt{P}$, which is the eigenvector of $C$ associated with its eigenvalue of order $P$. In the direction $\bar{f}=v_0$, the contraction is
\begin{equation}
    v_0^\top K_\text{E} v_0=\frac{1}{NP}\sum_i\sum_{\mu\nu}\sigma(h^\mu_i)\sigma(h^\nu_i) = \frac{P}{N}\sum_i \left( \frac{1}{P}\sum_\mu \sigma(h^\mu_i) \right)^2\, .
\end{equation}
In the previous equation, the average of the post-activations over the examples converges by the law of large numbers to
\begin{equation}
    m(g)=\mathbb{E}_z[\sigma(\sqrt{\rho}\,g+\sqrt{1-\rho}\,z)]\, , 
\end{equation} 
which is a nonlinear function of a single Gaussian variable. Since $|m(g)|<1$, the normalized squared projection $r_i=m(g_i)^2/b$, with $b=\mathbb{E}[m(g)^2]$ is bounded. As a consequence, the rate function of $q=\sum_{i=1}^{N}r_i / N$ is infinite outside a bounded interval, while the Wishart rate function of Eq.~\eqref{eq:single-q-rate-func} is finite for every $q>0$, meaning that along $v_0$ the two differ at leading order. 

We now show that this direction nevertheless does not contribute to the posterior, considering the case of labels aligned with it, $y=\mathbf{1}_P$. Starting from Eq.~\eqref{eq:Z_from_priorcharfunc}, we have
\begin{align}
Z &\propto \mathbb{E}_{K_\text{E}}\!\left[\int d\bar f\;e^{-\frac12\bar f^\top\left(\beta^{-1}\Id+K_\text{E}\right)\bar f+i\,\bar f\cdot y}\right] \\
&\propto \mathbb{E}_{K_\text{E}}\!\left[\det\big(\beta^{-1}\Id+K_\text{E}\big)^{-1/2}\,e^{-\frac12\,y^\top\left(\beta^{-1}\Id+K_\text{E}\right)^{-1}y}\right].
\end{align}
Writing $K_\text{E}$ in blocks, with $v_0$ as the first direction,
\begin{equation}
K_\text{E}=\begin{pmatrix}k_{00}&k_{0\perp}^\top\\ k_{0\perp}&K_\perp\end{pmatrix},\qquad
k_{00}=v_0^\top K_E\,v_0\simeq P\,b\,q,\qquad
s_0=\beta^{-1}+k_{00}-k_{0\perp}^\top\big(\beta^{-1}\Id+K_\perp\big)^{-1}k_{0\perp},
\end{equation}
where $K_\perp$ is the $(P-1)\times(P-1)$ block orthogonal to $v_0$ and $s_0$ is its Schur complement, one has exactly $\det(\beta^{-1}\Id+K_\text{E})=s_0\,\det(\beta^{-1}\Id+K_\perp)$ and $y^\top(\beta^{-1}\Id+K_\text{E})^{-1}y=P/s_0$, and therefore
\begin{equation}
Z=\mathbb{E}_{K_\text{E}}\!\Big[\det\big(\beta^{-1}\Id+K_\perp\big)^{-1/2}\;s_0^{-1/2}\,e^{-P/(2s_0)}\Big].
\label{eq:Z-split}
\end{equation}
At finite temperature $s_0$ is of order $P$ for every typical realization of the kernel so that the factor $s_0^{-1/2}e^{-P/(2s_0)}$, which contains the whole dependence on the direction $v_0$ and on the labels, is thus of order $P^{-1/2}$ and contributes to $\ln Z$ a term of order $\ln P$ (negligible with respect to the contribution of the determinant, which is of order $P$). This means that in the proportional regime the main contribution comes from the other $P-1$ orthogonal directions to the spike. The mechanism is analogous as the one that suppresses the difference between central and non-central Wishart distributions for non-zero-mean activations, discussed in App.~\ref{app:selfaveraging-noncentrality} above, with the spike generated here by the data instead of the mean of the activations.

We also tested the EWA in this setting with numerical simulations. To realize the equicorrelated example exactly, we built synthetic inputs whose Gram matrix coincides with $C$, with no sampling noise: denoting by $P_{\rm tot}=P+P_{t}$ the total number of training and test examples, we draw a matrix $O\in\mathbb{R}^{N_0\times P_{\rm tot}}$ with orthonormal columns, which requires $N_0\ge P_{\rm tot}$, and define the input matrix $X=\sqrt{N_0}\,C^{1/2}O^\top$, so that $XX^\top/N_0=C$ holds exactly and the first-layer pre-activations are distributed as $h\sim\mathcal N(0,C)$. All labels are set to $y^\mu=1$, i.e., they are aligned with the spike direction. The results are shown in Fig.~\ref{fig:C-with-spike}. In agreement with the argument above, the EWA remains accurate in this setting: we find good agreement for the learning curves both as a function of the width and as a function of the depth.}

\section{Additional details on the Large Deviation Analysis}
\label{appsec:rate-function-prior}

\subsection{Layer by layer and non central EWA rate function}
We here discuss more in depth some aspects on the large deviation analysis performed in Sec.~\ref{subsec:numerical-validation-ewa}.
Recall that the variables under investigation are the $q_\ell$ variables, defined as
\begin{equation}
    \label{eq:q_ell}
    q_\ell = \frac{\bar f^\top \Theta^{L-\ell}(K_\mathrm{E}^{(\ell)}) \bar f}{\bar f^\top \Theta^{L-\ell+1}(K_\mathrm{E}^{(\ell-1)}) \bar f}
\end{equation}
and viewed as sequences over the corresponding $N_\ell$. We keep a generic $N_\ell$, but all the plots will show for simplicity results obtained by sampling networks with the same number of neurons across layers: $N_\ell=N \, \forall \ell$.
Here we distinguish explicitly between central and non-central EWA, and we will provide additional numerical evidence on the claims made in the main text.
\newline
We first show that the normalized quantity in Eq.~\eqref{eq:q_ell}, introduced in the main text in the context of the central EWA, is actually left unchanged even for the non-central case. Indeed, as shown in Sec.~\ref{app:noncentral_contractions_and_tildeQ}, the non-central EWA implies that $\tilde Q_\ell \sim \chi_{\text{nc}}^2(N_\ell, N_\ell \lambda_{\text{nc},\ell})$. As a consequence, the normalized variable:
\begin{equation}
    \begin{aligned}
        \tilde q_\ell &\coloneq \frac{\tilde Q_\ell}{\mathbb{E}[\tilde Q_\ell]} = \frac{\tilde Q_\ell}{N_\ell(1+\lambda_{\text{nc},\ell})} = q_\ell.
    \end{aligned}
\end{equation}
Where the last equality follows from Eq.~\eqref{eq:tildeQ}. In particular, we can use the same sampling algorithm for both the central and non central case, the only difference being the specific function implemented by the NNGP kernel function $\Theta$. Naturally, the central and non-central EWA will generally lead to a different shape of the theoretical rate function, as will be now described, but the way in which $q_\ell$ and $\tilde q_\ell$ are calculated is the same, even if their distributions are different. In order to derive the layer-wise rate function for the non-central EWA, let us consider again the fact that $\tilde Q_\ell^{(N_\ell)} \sim \chi^2_{\text{nc}}(N_\ell, N_\ell \lambda_\ell)$. The moment generating function of the non-central chi-squared distribution is:
\begin{equation}
    M_{\tilde Q_\ell^{(N_\ell)}}(t) = (1-2t)^{-N_\ell/2} \exp \left\{ \frac{N_\ell \lambda_{\text{nc},\ell} t}{1-2t} \right\}, \qquad t<1/2.
\end{equation}
It follows that the moment generating function of the normalized $\tilde q_\ell$ are:
\begin{equation}
    M_{\tilde q_\ell^{(N_\ell)}}(t) = \left(1-\frac{2t}{N_\ell(1+\lambda_{\text{nc},\ell})}\right)^{-N_\ell/2} \exp \left\{ \frac{N_\ell \lambda_{\text{nc},\ell} t}{N_\ell(1+\lambda_{\text{nc},\ell})-2t} \right\}, \qquad t<N_\ell(1+\lambda_{\text{nc},\ell})/2.
\end{equation}
Therefore, the scaled cumulant generating functions are:
\begin{align}
    \Lambda_\ell(t) &= \lim_{N_\ell\to\infty} \frac{1}{a_{N_\ell}} \ln M_{\tilde q_\ell^{(N_\ell)}}(a_{N_\ell}t) \\
    &= \frac{\lambda_{\text{nc},\ell} t}{1+\lambda_{\text{nc},\ell}-2t} - \frac{1}{2} \ln \left( 1-\frac{2t}{1+\lambda_{\text{nc},\ell}} \right), \qquad t<(1+\lambda_{\text{nc},\ell})/2.
\end{align}
where the scales for these LDP turned out to be $a_{N_\ell}=N_\ell$. Finally, the layer by layer rate functions for the non-central EWA are obtained via Legendre–Fenchel tranform:
\begin{equation}
    \mathcal{I}_\ell (x) = \sup_{t<\frac{1+\lambda_{\text{nc},\ell}}{2}} \left\{ tx - \frac{\lambda_{\text{nc},\ell} t}{1+\lambda_{\text{nc},\ell}-2t} + \frac{1}{2} \ln \left( 1-\frac{2t}{1+\lambda_{\text{nc},\ell}} \right) \right\}.
\end{equation}
And this expression can be easily computed numerically. Notice that this time, differently than in the central case, the rate function is in principle different for different layers, but only through the non centrality parameter $\lambda_{\text{nc},\ell}$, recovering the central result for $\lambda_{\text{nc},\ell}=0$. This is due to the fact that, differently than in the central case, the $\tilde q_\ell$ variable are not decoupled across layers, since they conditionally depend on the previous layer kernel through the non centrality parameter, which we recall it is given by:
\begin{align}
    \lambda_{\mathrm{nc},\ell} = \frac{\bar{f}^\top m_{(\ell)} m_{(\ell)}^\top \bar{f}}{\bar{f}^\top \Sigma_{(\ell)} \bar{f}},
\end{align}
where $m_{(\ell)}$ and $\Sigma_{\ell}$ are calculated using $K_\mathrm{E}^{(\ell-1)}$, see Sec.~\ref{app:noncentral_contractions_and_tildeQ}. In our numerical experiments, we choose to control not the $\lambda_{\textrm{nc}}$ parameter but the overlap between $\bar f^\top$ and the last layer $m_{(L)}$, since this is the quantity that quantifies the contributions of non-central effects at the posterior level, so in the final integration over $d\bar f$. \replaced[id=rev1]{
The sampling scheme, which we call Conditional Sampling Algorithm, allows to get proper samples from each of the conditional distributions $\tilde q_\ell \mid K_{\mathrm{E}}^{(l-1)}$, and its scheme is summarized in Alg.~\ref{algorithm:conditional_sampling}.
}{For this reason, we need to modify the sampling algorithm used in the main text for the central case in order to get proper samples from each of the conditional distributions $\tilde q_\ell \mid K_{\mathrm{E}}^{(l-1)}$. On a practical level, the difference between this modified sampling algorithm, that we call Conditional Sampling Algorithm, and the simpler one used to sample from the central joint distribution is that in this case the denominator of Eq.~\eqref{eq:q_ell} is kept fixed across samples. It follows in particular that there is no difference between the two sampling procedures for $1$-hidden layer}

\begin{figure}[h!http]
    \includegraphics[width=\textwidth]{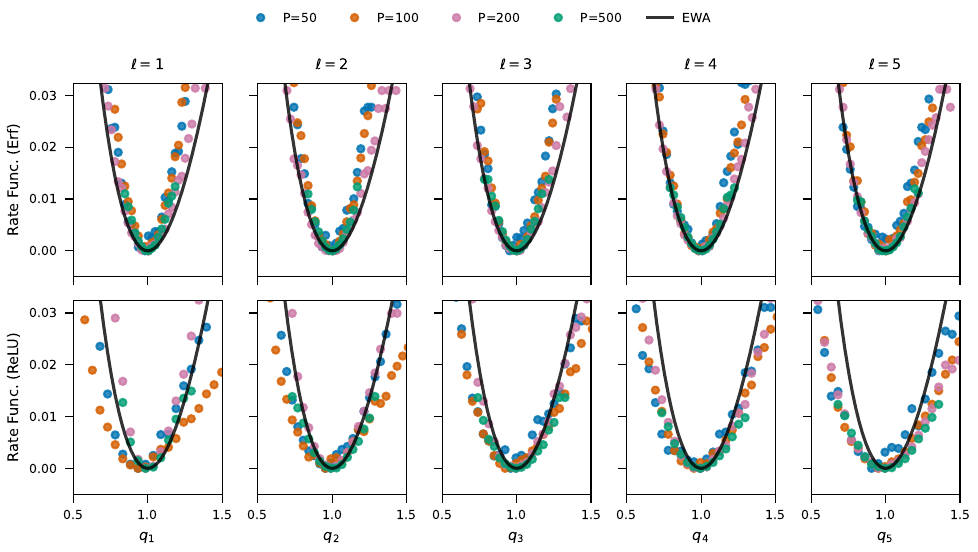} 
    \caption{\textbf{Layer by layer rate function for Erf and ReLU networks}. Numerical samples of the empirical rate function for the CIFAR-10 dataset (coloured dots) are compared to the expected theoretical rate functions for each individual layer (black dashed lines) for a deep network with $L=5$ hidden layers for both Erf (first row) and ReLU (second row) activation function. For both networks, the load is $\alpha=1.0$ and for the ReLU network the overlap parameter is zero. The empirical rate function is obtained using $5000$ samples from the Conditional Sampling Algorithm for each size of the dataset ranging in $P\in\{50,100,200,500\}$, as usually done to assess the asymptotic convergence of the empirical rate function to the theoretical one.}
    \label{fig:layers_rate_func}
\end{figure}

\begin{figure}[h!http]
    \includegraphics[width=\textwidth]{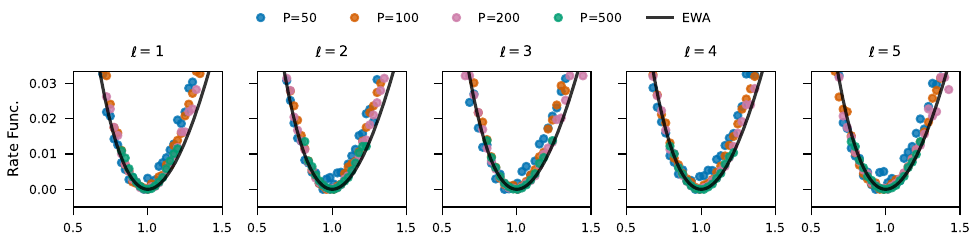}
    \caption{\textbf{Independent resampling of the layer by layer rate function.} The plot shows the same quantities \replaced[id=CK]{as in}{than} the first row of Fig.~\ref{fig:layers_rate_func}, but obtained from a different realization of the $\bar f$ vector. As expected, the conclusions from the resampling are left unchanged.}
    \label{fig:layer_rate_func_double}
\end{figure}

\begin{figure}[h!http]
    \includegraphics[width=\textwidth]{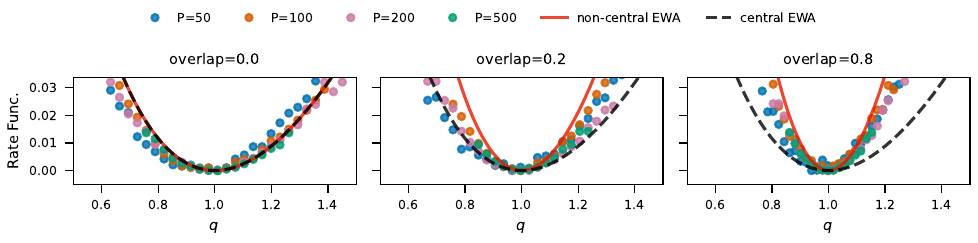}
    \caption{\textbf{Central vs non-central EWA rate function for a ReLU 1-hidden layer network.} Numerical samples of the empirical rate function for the CIFAR-10 dataset (coloured dots) are compared to the expected theoretical rate function under the central EWA (black dashed lines) and non-central EWA, (red lines) for a shallow network with ReLU activation function and $\alpha=1.0$. The different plots correspond to a different value of the overlap parameter: zero overlap (first column), small overlap (second column) and large overlap (third column). The empirical rate function is obtained using $5000$ samples for each size of the dataset ranging in $P\in\{50,100,200,500\}$, as usually done to assess the asymptotic convergence of the empirical rate function to the theoretical one.}
    \label{fig:rate_func_relu1HL}
\end{figure}

\begin{center}
\begin{minipage}{0.8\textwidth}
\begin{algorithm}[H]
\caption{Conditional Sampling Algorithm}
\label{algorithm:conditional_sampling}
\begin{algorithmic}[1]

\Require $K_0, \text{overlap},P,\alpha,L$

\State Sample $K_{\text{list}}^{*}:=(K_0,K_1^*,\dots,K_L^*)$
\State Compute $m_{L}$ using $K_{L-1}^{*}$
\State Sample $\bar f$ with the given overlap with $m_{L}$

\For{$\ell=1$ to $L$}
    \State Compute $\Theta^{L-\ell}(K_{\ell-1}^{*})$~\label{step:calculate_kernel_iteration}
    \State Compute $m_{\ell}$ with ~\eqref{step:calculate_kernel_iteration}~\label{step:calculate_ml}
    \State Compute $\lambda_{\mathrm{nc},\ell}$ using $K_{\ell-1}^*$ and ~\eqref{step:calculate_ml}
    \State Compute $\Theta^{L-\ell+1}(K_{\ell-1}^{*})$ iterating once more ~\eqref{step:calculate_kernel_iteration}
    \State Compute denominator contraction

    \For{$M=1$ to $N_{\text{samples}}$}
        \State Sample a new $K_{\ell}$ using $K_{\ell-1}^*$
        \State Compute $\Theta^{L-\ell}(K_{\ell})$
        \State Compute numerator contraction and $q_{\ell}^{(M)}$
    \EndFor

\EndFor
\end{algorithmic}
\end{algorithm}
\end{minipage}
\end{center}
\deleted[id=rev1]{As described in the main text, in the case of the central EWA independent samples from the joint distribution of $(q_1, \dots, q_L)$ are needed, in order to get proper samples of random variable $\mathcal{Q}=\prod_\ell q_\ell$.} Even though the predictive power of the EWA only require that the variable $\mathcal{Q}$ satisfies the LDP, we provide numerical results \added[id=rev1]{with direct prior sampling} concerning the LDP at the level of each individual layer also in the central case. \deleted[id=rev1]{Additionally, in} In order to numerically check that the $q_\ell$ variables are indeed decoupled across layers, as predicted by the central EWA, we use the conditional algorithm to estimate the layer by layer rate function in the central case: if the corresponding empirical rate functions are in agreement with the theoretical ones predicted by the EWA it means that the $q_\ell$ are indeed decoupled. 
Fig.~\ref{fig:layers_rate_func} shows numerical evidence on how the EWA is able to catch the statistical properties of the $q_\ell$ variable even at the level of each individual layer, especially in the Erf case (first row). We systematically observe some asymptotic deviation for ReLU networks (second row), \deleted[id=rev1]{suggesting that the deviation is then absorbed at the product level, see Fig.~\ref{fig:product-rate-function}} \added[id=rev1]{consistent with the observations made in the main text, see Sec.~\ref{subsec:numerical-validation-ewa}} and the following Sec.~\ref{app:toymodel-product-rate-func-deviations} for an explanation using a toy model with independent variables. 
It is worth mentioning how the result of the sampling procedure is stable across different realization of the $\bar f$ vectors: From Eq.~\eqref{eq:q_ell} it follows that the normalization of the $\bar f$ is not relevant, and different realization of the $\bar f$ vectors sampled uniformly from the unit $P$-dimensional sphere leads to the same conclusions, as expected. See Fig.~\ref{fig:layer_rate_func_double} for an example.
Fig.~\ref{fig:rate_func_relu1HL} shows instead how the non-central EWA is conceptually a more principled ansatz at the prior level, in the case of a shallow architecture for which the effective action was derived in Sec.~\ref{sec:EWA_nc_1HL}. The plot shows how the theoretical rate function of the non-central EWA is able to describe the empirical sampled points, especially for large values of the overlap parameters when the difference between central and non-central EWA becomes more evident. The reason why these differences become irrelevant at the level of the prior and posterior output distribution in the proportional regime is explained in Sec.~\ref{sec:noncentral_EWA}.

\subsection{Toy model for deviations from the theoretical product rate function}\label{app:toymodel-product-rate-func-deviations}

To understand how deviations in the width of layer-wise rate functions influence the product rate function, consider a simple toy model. Assume that the variables $x_\ell >0$ are independent and admit rate functions of the same speed of the form
\begin{equation}
I_\ell(x) = a_\ell I_x(x),
\end{equation}
where $I_x$ is a common base rate function and the prefactors $a_\ell$ represent quenched disorder around the homogeneous value $a_\ell=1$.
For the product
$y=\prod_{\ell=1}^L x_\ell$
the contraction principle gives
\begin{equation}
I_y(y;\{a_\ell\})=\inf_{\prod_{\ell=1}^L x_\ell = y}\sum_{\ell=1}^L a_\ell I_x(x_\ell).
\end{equation}
Expanding around the typical point $x_\ast=\mathrm{argmin\,I_x}$ such that
$I_x(x)\simeq \frac{\kappa}{2}(x-x_\ast)^2,$
and writing $x_\ell=x_\ast+\delta_\ell$, the product constraint expands to leading order as $x_\ast^{L-1}\sum_{\ell=1}^L \delta_\ell \simeq y - y_\ast,$ with $y_\ast=x_\ast^L$.
The rate function then reduces to the quadratic minimization problem
\begin{equation}
I_y(y;\{a_\ell\}) \simeq \inf_{\sum_\ell \delta_\ell = \frac{y-x_\ast^L}{x_\ast^{L-1}}}\frac{\kappa}{2}\sum_{\ell=1}^L a_\ell \delta_\ell^2,
\end{equation}
whose solution is
\begin{equation}
I_y(y;\{a_\ell\}) 
\simeq \frac{\kappa}{2}\,\frac{(y-x_\ast^L)^2}{x_\ast^{2L-2}\sum_{\ell=1}^L a_\ell^{-1}} 
= \bigg(\frac{1}{L} \sum_{\ell=1}^L a_\ell^{-1}\bigg)^{-1} I_y(y;\{a_\ell =1\}) .
\end{equation}
Thus the disorder enters only through the harmonic mean $\frac{1}{L}\sum_\ell a_\ell^{-1}$.

This model with independent $x_\ell$ immediately explains two qualitative features. First, the effect of the width-disorder is self-averaging and not cumulative. The homogeneous approximation therefore does not deteriorate in the product but can often be better than at individual layers. Second, the harmonic mean makes the rate function less sensitive to large values $a_\ell>1$ than to small values $a_\ell<1$, since increasingly large values of any given $a_\ell$ decrease $a_\ell^{-1}$ only weakly, whereas decreasing $a_\ell$ below unity enhances $a_\ell^{-1}$ more than proportionally. This corresponds to the dominance of the softest modes in the tails of the distribution of the product, while stiffer modes contribute less.

\begin{figure}
    \centering
    \includegraphics[width=\linewidth]{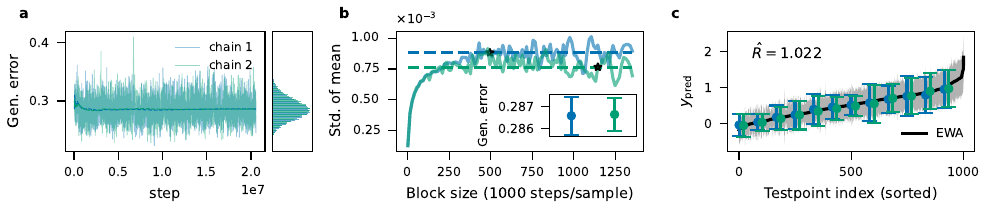}
    \caption{
    Analysis of a well-behaved Langevin Monte-Carlo history, at the example of two independent LMC chains for the $L=5$ point of Fig.~\ref{fig:finite-size-relu-cifar}. 
    (a) Full sampling history of the generalization error for two MCMC chains with independent weight initialization and noise realizations. Thin dashed lines show the running mean for each chain, and both sample histograms are shown overlaid on the right.
    (b) Estimation of the standard deviation of the mean via the blocking method. Both chains show a plateau of the error estimate for blocks greater than 500 samples (here corresponding to 500K LMC steps with learning rate $\eta=10^{-3}$), indicating the collection of independent samples from the posterior at such scales. The inset shows the corresponding empirical mean with its estimated error for both chains.
    (c) Finer grained per-testpoint analysis of chain-to-chain and chain-to-theory agreement. The black line and gray area show the output mean $\langle y_\mathrm{pred}\rangle$ and $\mathrm{Std}[y_\mathrm{pred}]$ for each of the $1000$ points of the test-set, as predicted by the EWA theory and sorted by their mean. Overlaid are the corresponding empirical results from both chains, displayed for a subset of test-points (green moved slightly to right to improve legibility). The $\hat R$ value close to $1$ as well as uniform agreement in per-point statistics across chains indicate good thermalization. 
    Parameters: $5$hL-ReLU network on CIFAR-10 classes $\{0,1\}$, $N=200$, $P=500$, $\gamma=1$ ($=$SP). Sampler LMC with $\eta=10^{-3}$ at $T=10^{-2}$.
    }
    \label{Sfig:history_analysis_L5_wellbehaved}
\end{figure}

\begin{figure}
    \centering
    \includegraphics[width=\linewidth]{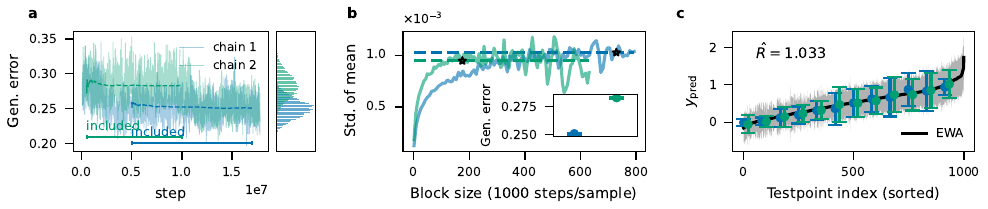}
    \caption{
    Analysis of a metastable MCMC history, showing two independent LMC chains for the $L=7$ point of Fig.~\ref{fig:finite-size-relu-cifar}. 
    For a description of the panel characteristics see Fig:~\ref{Sfig:history_analysis_L5_wellbehaved}.
    Here the analysis in panels (b, c) as well as running means and histograms are restricted to the two windows shown in the full history traces of panel (a). Note that judging from the plateaus of the blocking method in (b), LMC appears as if thermalized both in the metastable and the tentative true posterior states. The per-point comparison in panel (c) confirms that the metastable distribution is close to the distribution predicted by the EWA, and after the transition the predictions remain similar. The value of $\hat R $ was tracked during runtime and refers to the final state including all samples. 
    Parameters and sampler as in Fig.~\ref{Sfig:history_analysis_L5_wellbehaved}, but with $7$ hidden layers.
    }
    \label{Sfig:history_analysis_L7_bistability}
\end{figure}

\begin{figure}
    \centering
    \includegraphics[width=\linewidth]{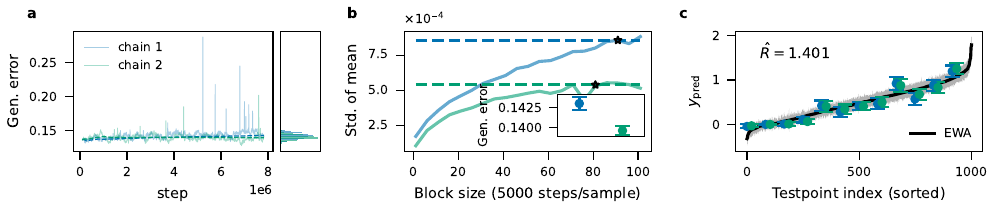}
    \caption{
    Analysis of a poorly converged MCMC history in a difficult sampling regime, showing two independent LMC chains for the $N=2500$ point in $\mu$-parametrization of Fig.~\ref{fig:muPvsSP} requiring a temperature of $T=4\times10^{-5}$. 
    For a description of the panel characteristics see Fig:~\ref{Sfig:history_analysis_L5_wellbehaved}.
    Due to the larger system size this sampling run includes only 8M steps as wall-time was restricted to $20$ hours. The history shows both instability events and relatively long autocorrelation time. Both the missing plateau in the blocking plot (b) indicates that the error bars in the inset are still underestimated, in agreement with the still significant difference of empirical means of the two chains. Noteworthy is that the per-point $y_\mathrm{pred}$ distributions in panel (c) do not all behave uniformly: Mostly the sampling results agree approximately in value and ordering with the theory prediction, while a smaller subset of points shows more pronounced differences. This small but discernible pattern was also observed on other $\mu$P sampling runs closer to convergence.
    Parameters: $4$hL-ReLU network on CIFAR-10 classes $\{0,1\}$, $N=2500$, $P=1000$, $\gamma=50$ ($=\mu$P). Sampler LMC with $\eta=0.25$ at $T=4\times 10^{-5}$.
    }
    \label{Sfig:history_analysis_muP_difficult_sampling}
\end{figure}

\subsection{Rate function beyond the Gaussian regime: tilted prior and thermodynamic integration}
\label{app:tilting_and_thermointegration}

\newtext{
The empirical rate functions of Figs. \ref{fig:layers_rate_func} - \ref{fig:rate_func_relu1HL} are estimated by a direct sampling from the prior. Since a value of $\mathcal Q$ with rate $\mathcal I$ is observed with probability that scale as $\sim e^{-N\mathcal I}$, direct sampling cannot resolve effectively the tail of the distribution, meaning that an exponentially large number of sample would be needed. The saddle point of the effective action \eqref{eq:action_singleoutput} sits instead at $\mathcal{Q}^*=\prod_\ell q^*_\ell\neq 1$, potentially in the tail of the distribution. To test the EWA also at more extreme values of $\mathcal{Q}$, we sample the \emph{true} MLP prior under a tilted distribution that makes such values of $\mathcal{Q}$ typical, and we reconstruct the rate function by thermodynamic integration \cite{neal1993probabilistic}. The EWA enters only in the choice of the tilt range, not in the estimator.
\newline ~ \newline
\textbf{Observable.} \\
Throughout this section $N_\ell=N\ \forall \ell$ and $\bar f\in\mathbb{R}^P$ is fixed. Multiplying the $q_\ell$ to form the $\mathcal Q$ observable, the intermediate contractions cancel and
\begin{equation}
\mathcal{Q}(\theta)=\prod_{\ell=1}^{L}q_\ell=\frac{\bar f^\top K_{\rm E}^{(L)}\bar f}{\bar f^\top \Theta^{L}(C)\,\bar f}\,,
\label{eq:Qtelescope}
\end{equation}
where the denominator is a constant and the numerator depends on the hidden-layers weights, which we keep denoting by $\theta$, only through the last-layer empirical kernel $K_{\rm E}^{(L)}$. Evaluating $\mathcal{Q}$ therefore requires a single forward pass on the training inputs, and none of the intermediate maps $\Theta^{L-\ell}(K_{\rm E}^{(\ell)})$.
\newline ~ \newline
\textbf{Tilted prior.} \\
For $t\in\mathbb{R}$ we define the (polynomially) tilted prior measure
\begin{equation}
dp_t(\theta):=\frac{\mathcal{Q}(\theta)^{Nt}}{M_{\ln\mathcal{Q}}(Nt)}\,dp(\theta),
\qquad
M_{\ln\mathcal{Q}}(Nt)=\mathbb{E}\big[e^{Nt\ln\mathcal{Q}}\big]=\mathbb{E}\big[\mathcal{Q}^{Nt}\big],
\label{eq:tilted_prior}
\end{equation}
and we denote by $\mathbb{E}_t$ and $\mathrm{Var}_t$ the expectation and variance under $p_t$, with $t=0$ being the prior. Notice that we implement polynomial tilt and not exponential tilt or, equivalently, an exponential tilt of $\ln \mathcal Q$ instead of $\mathcal Q$. Indeed, the exponential tilt is well defined only for $L=1$, while for $L\ge 2$ it is not defined under the EWA because the law of $\mathcal Q$ is described by a heavy-tail distribution, hence its normalization diverges under the EWA for every $t>0$. On the contrary, the polynomial tilt is normalizable under the EWA for all $t>-1/2$ (corresponding to the entire domain of $\mathcal{Q}$; the positive real axis), and its cumulant generating function is known in closed form at finite $N$. The contraction principle can then be used to map the result back to $\mathcal Q$.
\newline
$M_{\ln\mathcal{Q}}$ is the moment generating function of $\ln\mathcal{Q}$, so that the scaled cumulant generating function and its derivatives are given by
\begin{equation}
\Lambda^{(N)}_{\ln\mathcal{Q}}(t):=\frac1N\ln M_{\ln\mathcal{Q}}(Nt),
\qquad
\Lambda'^{(N)}_{\ln\mathcal{Q}}(t)=\mathbb{E}_t[\ln\mathcal{Q}] \eqcolon u(t),
\qquad
\Lambda''^{(N)}_{\ln\mathcal{Q}}=N\,\mathrm{Var}_t[\ln\mathcal{Q}] \eqcolon v(t) \,.
\label{eq:scgf_N}
\end{equation}
$\Lambda^{(N)}_{\ln\mathcal{Q}}$ is convex and vanishes at $t=0$. In calculating its Legendre transform along $u(t)$, the $\sup$ is attained at $s=t$:
\begin{equation}
\mathcal{I}^{(N)}_{\ln\mathcal{Q}}\big(u(t)\big)\coloneq\sup_{s}\big\{s\,u(t)-\Lambda^{(N)}_{\ln\mathcal{Q}}(s)\big\}
= t\,u(t)-\Lambda^{(N)}_{\ln\mathcal{Q}}(t).
\label{eq:legendre_N}
\end{equation}
By the G\"artner--Ellis theorem, if $\Lambda^{(N)}_{\ln\mathcal{Q}}$ converges to a differentiable limit as $N\to\infty$, then $\ln\mathcal{Q}$ satisfies an LDP with speed $N$ and rate function $\mathcal{I}_{\ln\mathcal{Q}}=\lim_N\mathcal{I}^{(N)}_{\ln\mathcal{Q}}$. Since $y=e^{u}$ is a continuous bijection, the contraction principle gives $\mathcal{I}_{\mathcal{Q}}(y)=\mathcal{I}_{\ln\mathcal{Q}}(\ln y)$. Hence
\begin{equation}
y(t)\coloneq e^{u(t)},\qquad
\mathcal{I}^{(N)}_{\mathcal{Q}}\big(y(t)\big)=t\,\ln y(t)-\Lambda^{(N)}_{\ln\mathcal{Q}}(t)
\label{eq:rate_param}
\end{equation}
is a finite-$N$ estimate of the rate function of $\mathcal{Q}$ at the point $y(t)$. It involves only expectations under $p_t$ and makes no reference to the EWA. The point $y(t)$ is, up to $O(1/N)$, the mean of $\mathcal{Q}$ under the tilted prior. Indeed
\begin{equation}
\begin{aligned}
    \mathbb{E}_t[\mathcal{Q}] &= M_{\ln\mathcal{Q}}(Nt+1)/M_{\ln\mathcal{Q}}(Nt) = \exp\{ N[\Lambda^{(N)}_{\ln\mathcal{Q}}(t+1/N) - \Lambda^{(N)}_{\ln\mathcal{Q}}(t)] \} \\
    &\Longrightarrow \quad \ln\mathbb{E}_t[\mathcal{Q}]=u(t)+O(1/N).
\end{aligned}
\end{equation}
\newline ~ \newline
\textbf{Theoretical tilt under the EWA.} \\
Under the EWA the $\mathcal Q$ variable is distributed as the product of $L$ independent (and also iid in the case $N_\ell=N$) Gamma random variables
\begin{equation}
    \mathcal Q \overset{\text{EWA}}{\sim} \prod_{\ell=1}^L \Gamma\left(\frac{N}{2}, \frac{2}{N} \right).
\end{equation}
With $\mathbb{E}[q_\ell^{\,n}]=(2/N)^{n}\,\Gamma(N/2+n)/\Gamma(N/2)$ for $n>-N/2$, hence for $t>-1/2$
\begin{align}
\Lambda^{(N)}_{\ln\mathcal{Q}}(t)&\overset{\rm EWA}{=}L\Big[t\ln\frac{2}{N}+\frac1N\ln\Gamma\Big(\frac N2(1+2t)\Big)-\frac1N\ln\Gamma\Big(\frac N2\Big)\Big],
\label{eq:scgf_EWA}\\
u(t)&\overset{\rm EWA}{=}L\Big[\ln\frac{2}{N}+\psi\Big(\frac N2(1+2t)\Big)\Big]=L\ln(1+2t)+O(L/N),
\label{eq:u_EWA}\\
N\,\mathrm{Var}_t[\ln\mathcal{Q}]&\overset{\rm EWA}{=}LN\,\psi^{(1)}\Big(\frac N2(1+2t)\Big)=\frac{2L}{1+2t}+O(L/N),
\label{eq:var_EWA}
\end{align}
where $\psi$ and $\psi^{(1)}$ are the digamma and trigamma functions respectively. In terms of the tilted mean
\begin{equation}
y_{\rm EWA}(t)=e^{u(t)}=(1+2t)^{L}
\quad\Longleftrightarrow\quad
t_{\rm EWA}(y)=\tfrac12\big(y^{1/L}-1\big),
\label{eq:y_EWA}
\end{equation}
and the latter is used in the numerical procedure only to estimate the range of the tilting parameter, without ever entering in the estimator.
For linear networks the EWA is exact and Eqs.~\eqref{eq:scgf_EWA}-\eqref{eq:var_EWA} hold at any $N$, which provides an exact benchmark of the numerical procedure, including also an exact analysis of the finite size effects.
\newline ~ \newline
\textbf{Sampling.} \\
The tilted distribution \eqref{eq:tilted_prior} is defined via a change of measure with respect to a Gaussian reference distribution (the prior), making the preconditioned Crank–Nicolson (pCN) \cite{cotter2013mcmc} method well suited for sampling from it. The Metropolis ratio of the pCN step contains only the tilt, and a proposal $\theta'$ is accepted with probability
\begin{equation}
\alpha(\theta, \theta') = \min\left\{1,\ \exp\left(Nt\,[\ln\mathcal{Q}(\theta')-\ln\mathcal{Q}(\theta)]\right)\right\}.
\label{eq:acc_tilt}
\end{equation}
The move is gradient-free and acts on the hidden-layer weights. At $t=0$ every proposal is accepted and the method yields independent samples of the prior. For more details about the pCN method see App. \ref{appsec:additional-mcmc-samplers}.
\newline ~ \newline
\textbf{Thermodynamic integration.} \\
Since $\Lambda^{(N)}_{\ln\mathcal{Q}}(0)=0$, Eq.~\eqref{eq:scgf_N} gives
\begin{equation}
\Lambda^{(N)}_{\ln\mathcal{Q}}(t)=\int_{0}^{t}u(s)\,ds .
\label{eq:TI}
\end{equation}
We estimate $u$ through its Monte Carlo estimator on a ladder of tilts $0=t_0,t_1,t_2,\dots$, built separately for the branch $t>0$, up to $t_{\rm high}$, and for the branch $t<0$, down to $t_{\rm low}>-1/2$ (the two branches are integrated independently). The spacing is chosen such that consecutive tilted priors overlap. Between neighboring tilts the mean of $\ln\mathcal{Q}$ shifts by 
\begin{equation}
\Delta u = \Lambda'^{(N)}_{\ln\mathcal{Q}}(t+\Delta t) - \Lambda'^{(N)}_{\ln\mathcal{Q}}(t) \simeq \Delta t \Lambda''^{(N)}_{\ln\mathcal{Q}}(t) = \Delta t N\,\mathrm{Var}_t[\ln\mathcal{Q}]\,.
\end{equation}
Requiring this shift to equal one standard deviation, $\Delta u\simeq(\mathrm{Var}_t[\ln\mathcal{Q}])^{1/2}$, and using Eq.~\eqref{eq:var_EWA}, we obtain
\begin{equation}
\Delta t\simeq c \sqrt{\frac{1+2t}{2NL}} \qquad c \sim O(1)\,,
\label{eq:dt}
\end{equation}
and in the simulations we use $c=0.5$.
Each chain is started from the last configuration of the previous tilt, so that it is already in the bulk of the new tilted prior and the required burn-in period is expected to be small. The cost of the procedure grows as $\sqrt{N}$, against the $e^{-N\mathcal{I}_{\mathcal{Q}}(y)}$ independent samples needed by direct sampling. At each tilt $t_k$ we measure the sample mean $\hat u_k$ and the sample variance $\hat v_k$ of $\ln\mathcal{Q}$, and we integrate Eq.~\eqref{eq:TI} with the cubic Hermite rule, which uses both the integrand and its derivative $N\,\mathrm{Var}_t[\ln\mathcal{Q}]$ at the nodes:
\begin{equation}
\hat\Lambda_{k+1}=\hat\Lambda_{k}+\frac{h_k}{2}\big(\hat u_k+\hat u_{k+1}\big)+\frac{N h_k^{2}}{12}\big(\hat v_k-\hat v_{k+1}\big),
\qquad h_k=t_{k+1}-t_k,\quad \hat\Lambda_0=0,
\label{eq:hermite}
\end{equation}
valid for both signs of $h_k$. From Eq.~\eqref{eq:rate_param}, the estimate of the rate function is
\begin{equation}
\hat{\mathcal{I}}_{\mathcal{Q}}(y_k)=t_k\,\hat u_k-\hat\Lambda_k,\qquad y_k=e^{\hat u_k}.
\label{eq:I_hat}
\end{equation}
The procedure is summarised in Alg.~\ref{alg:TI}. 
Equilibration along the ladder was checked by running each branch outward and back and verifying that the two estimates agree within statistical errors.
} 
\begin{algorithm}[H]
\caption{Tilted sampling and thermodynamic integration (one branch, one sweep)}
\label{alg:TI}
\begin{algorithmic}[1]
\Require data $X$, test vector $\bar f$, $N$, $L$, $c$, end point $t_{\rm end}\in\{t_{\rm low},t_{\rm high}\}$
\State $t\gets 0$,\ $\hat\Lambda\gets 0$; sample $\theta$ from the prior and compute $\hat u,\hat v$ from $\ln\mathcal{Q}(\theta)$, Eq.~\eqref{eq:Qtelescope}
\While{$|t|<|t_{\rm end}|$}
  \State $h\gets \mathrm{sign}(t_{\rm end}) c \sqrt{(1+2t)/(2NL)}$;\quad $t'\gets t+h$ \Comment{Eq.~\eqref{eq:dt}}
  \State run pCN at tilt $t'$ from the current $\theta$, accepting with Eq.~\eqref{eq:acc_tilt}
  \State (after burn in) compute sample mean $\hat u'$ and variance $\hat v'$ of $\ln\mathcal{Q}$
  \State $\hat\Lambda\gets\hat\Lambda+\tfrac{h}{2}(\hat u+\hat u')+\tfrac{N h^{2}}{12}(\hat v-\hat v')$ \Comment{Eq.~\eqref{eq:hermite}}
  \State store $y=e^{\hat u'}$, $\hat\Lambda$ and $\hat{\mathcal{I}}_{\mathcal{Q}}(y)=t'\hat u'-\hat\Lambda$ \Comment{Eq.~\eqref{eq:I_hat}}
  \State $t\gets t'$,\ $\hat u\gets\hat u'$,\ $\hat v\gets\hat v'$
\EndWhile
\end{algorithmic}
\end{algorithm}

\begin{figure*}[htbp]
    \centering
    \includegraphics[width=0.48\textwidth]{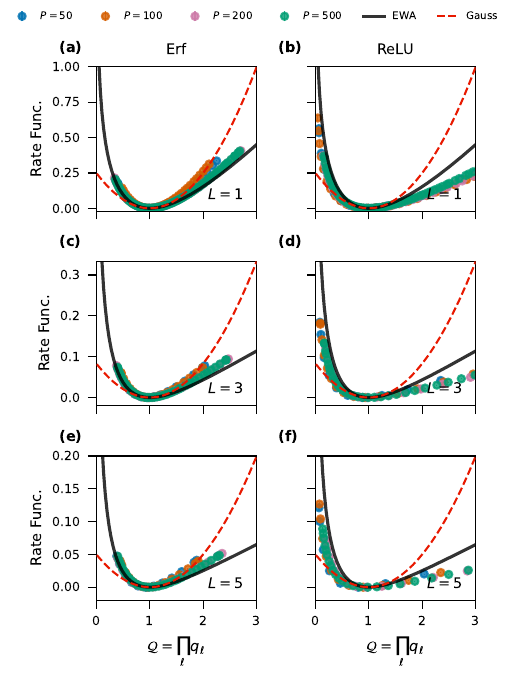}
    \hfill
    \includegraphics[width=0.48\textwidth]{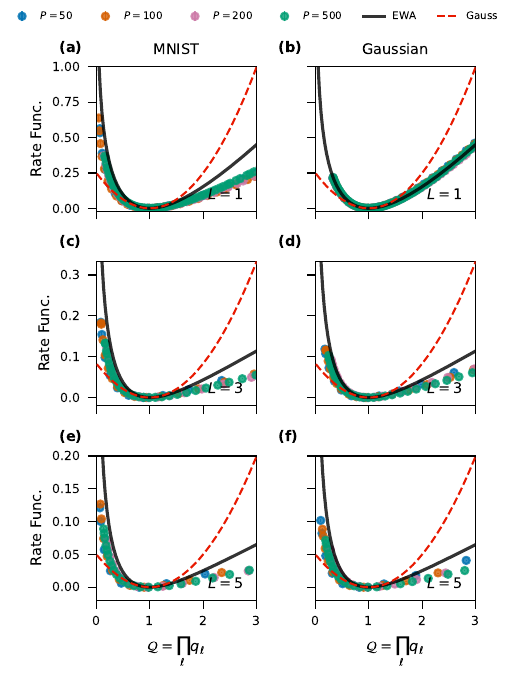}
    \caption{\newtext{\textbf{Supplementary results obtained using the tilting method.} 
    \textbf{Left panel:} Results for the MNIST dataset across both activation functions (analogous to the corresponding CIFAR-10 results in the main text, Fig.~\ref{fig:product-rate-function}). 
    \textbf{Right panel:} Results for MNIST and Gaussian data using the ReLU activation function (analogous to the Erf activation results in the main text, Fig.~\ref{fig:rate_function_TI_dataset}).}}
    \label{fig:TI_panel_appendix}
\end{figure*}

\added[id=rev1]{
The results obtained with the tilting method are reported in the main text. For completeness, we show in Fig.~\ref{fig:TI_panel_appendix} additional plots regarding results for the MNIST dataset and both activations (left panel) and for MNIST and Gaussian data for ReLU activation (right panel), while we refer the reader to Sec.~\ref{subsec:numerical-validation-ewa} for the discussion. 
}

\section{Ease and difficulty of sampling the overparametrized DNN posterior}
\label{appsec:ease-and-difficulty-of-sampling-the-overparametrized-MLP-posterior}

Generally, sampling from the posterior of a DNN is a formidable task due to its very high dimensionality and multi-modality. From the outset, it can only become feasible since we are interested solely in low-dimensional summary statistics such as the generalization error, or at most the marginals of the output posterior ($P$ scalar distributions), instead of the full joint distribution in $N\times N$ parameter space. Difficulties can especially be expected at low temperature and high network load $\alpha = P/N$. In such cases, due to the non-convex loss landscape in the parameter space, Monte Carlo sampling could realistically exhibit several systematic biases in the exploration of the configuration space, breaking convergence to the true posterior on any feasible time scale and trustworthiness of numerical outcomes. There are multiple reasons to expect a breakdown of Monte Carlo sampling: for example, networks can get stuck in individual branches of the non-convex loss landscape and require exponential time to escape \cite{izmailov2021bayesian}, or the simulation can slow down due to poor gradient propagation across layers such that some parameter groups are sampled on much slower time scales than others \cite{glorot2010understanding}. Nevertheless Monte Carlo practitioners have different strategies to numerically probe the stability of their simulations. Below, we present the algorithms we employed for sampling, together with the statistical tools we implemented to assess the quality of our simulations. Discussing examples representative of our results, we show that at our working scale Bayesian experiments involving non-linear DNNs are feasible.

\subsection{Additional MCMC samplers}
\label{appsec:additional-mcmc-samplers}

A possible approach to probe the convergence of Monte Carlo simulations is to compare outcomes from different algorithms. For this reason, in hard regimes where one simultaneously needs large depth and training set size we employed several different algorithms. We chose algorithms that do not share the same paradigm for exploring the network configuration space. In particular, we implemented a pure gradient-based Monte Carlo sampler (Langevin Monte Carlo -- LMC), a pure energy-based Monte Carlo sampler (the preconditioned Crank--Nicolson MCMC algorithm -- pCN), a mixed gradient-based and energy-based Monte Carlo algorithm (the Metropolis Adjusted Langevin Algorithm -- MALA), and a momentum-driven Monte Carlo algorithm (the Hamiltonian Monte Carlo algorithm with No-U-Turn Sampler -- NUTS HMC). In the manuscript, Sec.~\ref{subsec:algorithms-used}, we already discussed the LMC implementation, since we mainly used this algorithm for testing the EWA in different learning scenarios (see the section below for details on how we assessed the sampling robustness of this algorithm). Here we provide a brief description of the routines we used to implement the other algorithms:
\begin{itemize}
    \item \underline{Metropolis Adjusted Langevin Algorithm -- MALA} \\
    The MALA algorithm is an exact Bayesian sampler built on the discretized Langevin equation. While more expensive, its main advantage compared to LMC is that it samples from the true posterior, avoiding additional systematics arising from the finite learning rate. Since the standard discretized Langevin transition
    \begin{equation}
        \label{eq:transitions}
        \theta \rightarrow \theta' = \theta - \epsilon \partial_\theta\left[ \mathcal{L}_{\mathrm{reg.}}\right]_{\theta} + \sqrt{2\epsilon T}\eta
    \end{equation}
    is not symmetric, i.e. $P_\beta(\theta\to\theta')\ne P_\beta(\theta'\to\theta)$, it is not enough to correct the transition $\theta\to\theta'$ with a Metropolis step to ensure that detailed balance is satisfied. On the contrary, one has to take into account the full transition probability between two states $\theta$ and $\theta'$, which basically amounts to accepting the new state $\theta'$ as a proposal from state $\theta$ with probability
    \begin{align}
        \label{eq:acc_prob}
        p_\beta(\theta, \theta') = \min \left( 1, \frac{P_\beta(\theta')P_\beta(\theta'\to\theta)}{P_\beta(\theta)P_\beta(\theta\to\theta')} \right)\, ,
    \end{align}
    where the transition amplitudes can be easily obtained from Eq.~\eqref{eq:transitions}, namely:
    \begin{equation}
        \label{eq:MALA_transitions}
        P_\beta(\theta\to\theta') = P_\beta(\eta = [\theta' - \theta + \epsilon \partial_\theta \mathcal{L}_{\mathrm{reg.}}]/\sqrt{2\epsilon T} ) \propto \exp{\left\{-\frac{1}{4\epsilon T}\left( \theta' - \theta + \epsilon \partial_\theta \mathcal{L}_{\mathrm{reg.}} \right)^2\right\}}\, .
    \end{equation}
    So, MALA simply amounts to iterating the gradient-based proposal in Eq.~\eqref{eq:transitions} and the energy-based acceptance step in Eq.~\eqref{eq:acc_prob}. The price one pays to avoid finite learning rate effects is not only the computation of the loss at each step, but also the computation of the transition amplitudes. We note also from Eqs.~\ref{eq:acc_prob} and~\ref{eq:MALA_transitions} that the acceptance rate depends on the choice of learning rate, requiring in principle extra preliminary simulations to fix the value of $\epsilon$ for each set of parameters in order to obtain a fixed target acceptance probability. In our simulations we fixed $\epsilon=10^{-3}$, which allowed us to keep the average acceptance probability in the range $p_\beta\approx 0.8$ for $L=5$ and $p_\beta\approx 0.4$ for $L=10$.
    \item \underline{Preconditioned Crank--Nicolson MCMC algorithm -- pCN} \\
    The preconditioned Crank--Nicolson is a Monte Carlo sampling algorithm with the special feature of preserving a good acceptance rate even in the high-dimensional limit (which is not the case for MALA, for example, where $p_\beta\to0$ as $N_\ell\to\infty$). The pCN proposal is gradient-free, namely:
    \begin{equation}
        \theta\to\theta'=\sqrt{1-\phi^2}\theta + \frac{\phi}{\sqrt{\lambda}}\xi\, , \qquad \xi\sim\mathcal{N}(0,\mathbb{1})\, ,
    \end{equation}
    where $\phi\in [0,1]$ is a free parameter to be optimized to ensure a reliable acceptance rate. The new proposal $\theta'$ is accepted with probability
    \begin{equation}
        \label{eq:acc_step_pcn}
        p_\beta(\theta,\theta')=\min[1, \exp(-\beta(\mathcal{L}(\theta')-\mathcal{L}(\theta)))] \, .
    \end{equation}
    Note that, unlike MALA, the acceptance step of pCN is computed only on the 
    likelihood and not on the full regularized loss. It is also worth mentioning that the pCN proposals always remain on the shell of the prior norm, rather than drifting away from it, diffusing through the acceptance step in Eq.~\eqref{eq:acc_step_pcn}. In this work, we fixed $\phi=0.002$, which yields an average acceptance probability in the range from $p_\beta \approx 0.48$ (for $L=5$) to $p_\beta\approx 0.27$ (for $L=10$).
    \item \underline{No-U-Turn Hamiltonian Monte Carlo -- NUTS HMC}\\
    The NUTS HMC algorithm is a state-of-the-art sampler which is particularly suitable for Bayesian models. While it is well known that the vanilla HMC algorithm is a sampler with fast mixing capabilities due to momentum-driven phase-space exploration (resulting in short decorrelation timescales), it is also well known that HMC requires preliminary runs to set the hyperparameters of the simulation, to which the performance of the algorithm is very sensitive. The most crucial hyperparameters of the simulation are the integration step size and the trajectory length. In particular, a wrong choice of the latter is chancy for the simulation, as it can yield trajectories which fold back onto the same track. NUTS HMC builds on top of the HMC algorithm and solves the problem of fixing the trajectory length. Roughly speaking, the algorithm tries different trajectories forward and backward in time, increasing the trajectory length until the system is far from loops back to a previous position in the trajectory (in jargon, a U-turn), and samples one trajectory among those generated this way to ensure detailed balance. A full description of the sampling strategy, which is beyond the scope of this manuscript, is presented in Ref.~\cite{hoffman2014no}. In this work we used the NUTS HMC sampler provided by the \texttt{NumPyro} Python package~\cite{numpyro}, which also provides optimal fine-tuning of the learning rate. It can be shown that this algorithm performs at least as well as optimally fine-tuned HMC~\cite{hoffman2014no}. Before starting the sampling, we fixed a pre-calibration phase of $500$ steps: during this warm-up phase, the system was first brought to thermalization in $O(10)$ steps, and then the remaining steps were used to fix the optimal trajectory length and integration step size to achieve an optimal acceptance rate. Typical values of the step size $\epsilon$ are in the range $10^{-4}<\epsilon<10^{-3}$, while for the trajectory length $\tau$ we obtained an optimal $\tau \approx O(10^3)$, which yielded typical average acceptance probabilities $0.7<p_\beta<0.95$.
\end{itemize}

\subsection{Methods used to assess sampling convergence}
\label{appsubsec:additional-details-on-monte-carlo-sampling}

The estimation of the autocorrelation time $\tau_{\text{int}}$~\cite{Wolff2004} is a well-known tool for the computation of statistical errors associated with Monte Carlo measurements. In this work we used the blocking method, which gives an indirect measure of the autocorrelation time by means of the statistical estimation of the standard deviation of the mean. It operates by partitioning the data into blocks and averaging the values within each block, thereby creating new samples of reduced length and decreased autocorrelations. As the block size grows, the error estimate obtained from the naive standard error of the mean (i.e. neglecting autocorrelation effects) on the blocked samples becomes increasingly accurate, eventually converging to a plateau once the blocks are effectively uncorrelated. The plateau value $\delta$ corresponds to the actual statistical error associated with the mean, and the integrated autocorrelation time can be computed using the relation $\delta^2 = \sigma_0^2 2\tau_{\text{int}}/N_{\text{samples}}$, where $\sigma_0$ is the standard deviation of the original samples. Note that from the previous formula one can also obtain the number of independent samples $N_{\text{ind. samples}}=N_{\text{samples}}/(2\tau_{\text{int}})$. In Fig.~\ref{Sfig:history_analysis_L5_wellbehaved} (b) we show a representative example of the estimation of the statistical error using the blocking method in the case of a MLP with $5$ hidden layers using the LMC algorithm (note also that the autocorrelation times depend on the sampling algorithm at hand). It is worth mentioning that the blocking method can also be used to preliminarily probe the convergence properties of the simulation towards the equilibrium distribution. Typically, if the algorithm is unable to explore the configuration space consistently, for example being trapped in a sub-branch of the loss landscape, the blocking method fails, meaning that the plateau cannot be reached even for very large block sizes due to slow propagation of modes in the sampling time. We report in Fig.~\ref{Sfig:history_analysis_muP_difficult_sampling} an example of a poorly converged LMC simulation: in panel (b) it is shown that a plateau cannot be reached, indicating sampling issues. 

We also assessed the convergence of Bayesian sampling using multiple parallel chains, i.e. simulating two independent Monte Carlo histories obtained with the same algorithm, but with different initial configurations and different seeds for the random number generator. In particular, we computed the Gelman--Rubin statistic $\hat{R}$~\cite{rhat}, which allows to quantify the convergence of Monte Carlo simulations. Here we computed $\hat{R}$ for the predictor of each test example and reported the averaged value. A value of the Gelman--Rubin coefficient $\hat{R}\approx1$ signals robust convergence properties of the simulations, as shown in the representative example in Fig.~\ref{Sfig:history_analysis_L5_wellbehaved}, panel (c). On the contrary, in Fig.~\ref{Sfig:history_analysis_muP_difficult_sampling} (c), the value $\hat{R}\approx 1.4$ signals poorly converged simulations, in agreement with what we discussed above using the blocking method. Also the blocking method benefits from multiple chain simulations: since the autocorrelation time depends only on the stochastic evolution implemented and not on the initial configuration or the random number generator, different chains should exhibit the same plateau. This is again the case in Fig.~\ref{Sfig:history_analysis_L5_wellbehaved} (b), where the two plateau are indistinguishable up to statistical noise, while in Fig.~\ref{Sfig:history_analysis_muP_difficult_sampling} (b) the two plateau 
are inconsistent across the two chains. An example of the metastability effects discussed in Sec.~\ref{subsec:emergent-performance-transitions} is reported in Fig.~\ref{Sfig:history_analysis_L7_bistability}. In this case, in panel (b), it can be seen that when restricting to subsets of samples belonging to the two different states, the blocking method shows the same plateau and does not indicate that the simulations are far from equilibrium. In other words, the two slices of sampling histories appear as locally stable macrostates.

\section[Spatial kernel renormalization in deep CNNs]{\added[id=rev1]{Spatial kernel renormalization in deep CNNs}}
\label{app:cnn-spatial}

\underline{\added[id=rev1]{Spatial paths and kernel blocks.}} \added[id=rev1]{We give the construction of $\mathcal H, \mathcal Q$ and $K_{\mathcal Q}^{(R)}$ in Eq.~\eqref{eq:cnn-path-kernel-renorm} for $L$ layers, starting for readability with zero-mean activations and a flattened scalar readout.  As in the main section, $\mathcal M_\ell$ is the filter mask, $M_\ell=|\mathcal M_\ell|$ its number of offsets (e.g., four for a two dimensional $2\times2$ filter), $p_\ell$ the number of spatial positions and $A_\ell$ the number of channels.
To compactly treat various convolutional geometries, we denote $i\oplus_\ell m$ for the position in layer $\ell-1$ contributing to position $i$ in layer $\ell$ by filter offset $m$. For example, in a one-dimensional convolution with no padding, one would have $i\oplus_\ell m=S_\ell i+m$ with $S_\ell$ the stride. This notation can also cover padding at the borders, and when applying the corresponding boundary prescription our derivations below are compatible with  zero-, periodic- and replication padding. We place the weight-prior precisions $\lambda$ in $\omega_\ell$ and suppress their possible index $\ell$ for brevity, $\Theta$ is then the nonlinear kernel map independent of $\lambda$. Denoting the deterministic pre-activation kernels in the large-width limit as $Z_\ell$, their recursion reads}
\begin{equation}
\begin{gathered}
\added[id=rev1]{\relax Z_1=\omega_1(C),\qquad
 Z_\ell=\omega_\ell\!\left(\Theta(Z_{\ell-1})\right),}\\
\added[id=rev1]{\relax [\omega_\ell(K)]^{\mu\nu}_{ij}
 =\frac{1}{M_\ell\lambda}\sum_{m\in\mathcal M_\ell}
 [K]^{\mu\nu}_{i\oplus_\ell m,j\oplus_\ell m}\,.}
\end{gathered}
\label{eq:cnn-iw-spatial}
\end{equation}
\added[id=rev1]{Here $C$ is the input covariance averaged over input channels, and the recursion for $Z_\ell$ applies for $\ell\geq2$. Each example-pair block $Z_\ell^{\mu\nu}$ or $V_\ell^{\mu\nu}$ has two spatial (-patch) indices. These reference kernels are computed once for the given inputs and architecture.}
\added[id=rev1]{\relax Each example-pair block $Z_\ell^{\mu\nu}$ or $\Theta(Z_\ell)^{\mu\nu}$ has two spatial patch indices $ij$, each of which can be a multi-index for multi-dimensional convolutional filters. The large-width reference kernels are computed once for the given inputs and architecture.}

\added[id=rev1]{The spatial contributions to the output can be indexed by paths through the convolutional layers. For example, consider a second-layer $2\times2$ filter with stride one that maps a $3\times3$ first-layer spatial grid to a $2\times2$ grid, without need for padding. Each of the four final positions has four filter offsets, giving sixteen paths to nine first-layer positions. Several paths therefore reach the same position, due to overlapping filter masks. 
At general depth, the path index at layer $\ell$ records the remaining offsets and the final position, $\pi_\ell=(m_{\ell+1},\ldots,m_L,i_L)$, with $r_\ell=p_L\prod_{k=\ell+1}^L M_k$ possible values. Starting at $i_L$ and following these offsets backwards specifies the visited positions. Along the full path $\pi=\pi_1$, the visited positions are then written as $i_L(\pi)=i_L$ and $i_{\ell-1}(\pi)=i_\ell(\pi)\oplus_\ell m_\ell$ for $\ell=L,\ldots,2$. The first offset in a path $\pi$ is $m_2$, since the first convolution and activation are already included in $\Theta(Z_1)$. This again comes from the difference between one-hidden layer and deep CNNs; for $L=1$ no paths need to be counted as filter positions can only overlap on the deterministic input images.}

\added[id=rev1]{\relax An entry of $\mathcal H^{\mu\nu}$ pairs two such paths: it starts with their first-layer kernel, multiplies the gains at the pairs of positions visited in later layers, and includes the convolutional weight normalizations. Explicitly,}
\begin{equation}
\added[id=rev1]{\relax \mathcal H^{\mu\nu}_{\pi\pi'}=
 [\Theta(Z_1)]^{\mu\nu}_{i_1(\pi),i_1(\pi')}
 \prod_{\ell=2}^L
 \frac{[b_\ell]^{\mu\nu}_{i_\ell(\pi),i_\ell(\pi')}}{M_\ell\lambda},
 \qquad [b_\ell]^{\mu\nu}_{ij}=
 \frac{[\Theta(Z_\ell)]^{\mu\nu}_{ij}}{[Z_\ell]^{\mu\nu}_{ij}}\,.}
\label{eq:cnn-path-feature}
\end{equation}

\added[id=rev1]{The readout will then contribute a separate factor $1/(p_L\lambda)$ to the final output covariance in Eq.~\eqref{eq:cnn-forward-wishart}. The gains are evaluated at the large-width kernels and remain fixed as the $\mathcal Q_\ell$ vary. Thus the nonlinear kernel map is represented by a fixed entrywise multiplication, with a gain defined by the covariance ratio above. Two paths ending at the same first-layer position can visit different positions above and acquire different gains. This is why they are retained as separate coordinates. For a linear activation all gains are $b_\ell=1$, and their contributions can instead be collected on the spatial positions, as in the exact linear construction of~\cite{Bassetti:JMLR:2024}.}

\added[id=rev1]{The matrix $\boldsymbol{\mathcal H}$ formed by stacking these blocks on the joint example and path indices is positive semidefinite. To see this for a zero-mean activation such as Erf and positive preactivation variances, expand the activation in Hermite polynomials. Its covariance is a sum of entrywise powers of the preactivation correlation matrix, with outer-product coefficients. There is no constant term, so division by $Z_\ell$ reduces each power by one and leaves a positive semidefinite matrix. By the Schur product theorem, entry-wise products of positive-semidefinite matrices retain that property. Therefore, evaluating the gain matrices on the visited path positions and taking their entrywise products, as well as with the corresponding entries of $\Theta(Z_1)$ preserves positive semidefiniteness. Thus $\boldsymbol{\mathcal H}$ is a valid covariance for the forward Wishart representation below.}

\underline{\added[id=rev1]{\relax Forward and backward Wishart representations.}} \added[id=rev1]{\relax Once $\boldsymbol{\mathcal H}$ is fixed, the EWA can be formulated as a sequence of auxiliary random kernels at each layer, which have two indices running over paths truncated at that layer. We denote these $Pr_\ell\times Pr_\ell$ path-indexed matrices by $\widehat H_\ell$. The $\widehat H_\ell$ already incorporate the subsequent gains along the indexed paths, making them analogous to the $\Theta^{L-\ell}(K_{\mathrm E}^{(\ell)})$ of the MLP derivation. Now, moving forward by one layer removes one offset from the indexed paths. Writing $\pi_\ell=(m_{\ell+1},\pi_{\ell+1})$, this operation acts as the partial trace $[\Tr_{m_{\ell+1}}(\widehat H_\ell)]^{\mu\nu}_{\pi_{\ell+1}\pi'_{\ell+1}}=\sum_{m_{\ell+1}}[\widehat H_\ell]^{\mu\nu}_{(m_{\ell+1},\pi_{\ell+1}),(m_{\ell+1},\pi'_{\ell+1})}$ shortening the paths by one layer. Only equal offsets are summed, reflecting that distinct filter weights are independent. At the last layer, correspondingly define $[\Tr_{i_L}(\widehat H_L)]^{\mu\nu}=\sum_{i_L}[\widehat H_L]^{\mu\nu}_{i_Li_L}$ for the flattened readout. The normalization factors of $\omega_{\ell+1}$ are already included in $\mathcal H$ which will form the first-layer anchor of the chain, and the product of gains remains intact in the same way as all scale matrices in Eq.~\eqref{eq:EWA-scalematix} refer to the final activation layer.  With $A_\ell$ channels in layer $\ell$, the forward representation is therefore}
\begin{align}
\added[id=rev1]{\relax\widehat H_1}
 &\added[id=rev1]{\relax \sim\mathcal W_{Pr_1}
       (\boldsymbol{\mathcal H}/A_1,A_1),} \nonumber\\
\added[id=rev1]{\relax\widehat H_{\ell+1}
       \mid\widehat H_\ell}
 &\added[id=rev1]{\relax \sim\mathcal W_{Pr_{\ell+1}}
       \big(\Tr_{m_{\ell+1}}(\widehat H_\ell)/A_{\ell+1},A_{\ell+1}\big),}
 \label{eq:cnn-forward-wishart}\\
\added[id=rev1]{\relax f\mid\widehat H_L}
 &\added[id=rev1]{\relax \sim\mathcal N\!\left(0,\frac{1}{p_L\lambda} \Tr_{i_L}(\widehat H_L)\right).} \nonumber
\end{align}
\added[id=rev1]{\relax The first Wishart has mean $\boldsymbol{\mathcal H}$; each subsequent conditional mean is the offset trace of the preceding kernel. For a nonlinear network these random kernels describe the propagated contributions to the output, with all gains fixed by Eq.~\eqref{eq:cnn-path-feature}.}

\added[id=rev1]{\relax We now integrate backwards, as in the main-text calculation. The identity $\Tr[\Tr_{m_{\ell+1}}(K)\,\mathcal Q_{\ell+1}]=\Tr[K(\Id_{M_{\ell+1}}\otimes\mathcal Q_{\ell+1})]$ on each example-pair block transfers the offset trace to the order parameter matrix: each filter offset gets a copy of the same $\mathcal Q_{\ell+1}$. Applying the Wishart transform successively builds the cumulative product matrices $\mathcal Q_\ell$ }
\begin{equation}
\begin{gathered}
\added[id=rev1]{\relax \mathcal Q_L\sim\mathcal W_{r_L}(\Id_{r_L}/A_L,A_L),}\\
\added[id=rev1]{\relax \mathcal Q_\ell\mid\mathcal Q_{\ell+1}
 \sim\mathcal W_{r_\ell}\!\left(
 \Id_{M_{\ell+1}}\otimes\mathcal Q_{\ell+1} / A_\ell,A_\ell\right),}\\
\added[id=rev1]{\relax \mathcal Q=\mathcal Q_1.}
\end{gathered}
\label{eq:cnn-backward-Qs}
\end{equation}
\added[id=rev1]{\relax The conditional mean is $\Id_{M_{\ell+1}}\otimes\mathcal Q_{\ell+1}$, and the unconditional means are identities. Equivalently, the correlations across layers, which make it natural to work with the cumulative products $\mathcal Q_\ell$, can be decoupled by introducing the independent and normalized layer matrices}
\begin{equation}
\begin{gathered}
\added[id=rev1]{\relax q_\ell\sim\mathcal W_{r_\ell}(\Id_{r_\ell}/A_\ell,A_\ell)
 \quad \mathrm{independently } \;\; \forall \ell,}\\
\added[id=rev1]{\relax \mathcal Q_\ell
 =(\Id_{M_{\ell+1}}\otimes\mathcal Q_{\ell+1}^{1/2})\,q_\ell\,
  (\Id_{M_{\ell+1}}\otimes\mathcal Q_{\ell+1}^{1/2}),
 \quad \ell<L.}
\end{gathered}
\label{eq:cnn-independent-innovations}
\end{equation}
\added[id=rev1]{\relax Thus $\mathcal Q_\ell$ accumulates the fluctuations from layers $\ell,\ldots,L$, while $q_\ell$ represents the independent fluctuation introduced at layer $\ell$. For the implementation of the theory, we minimize the effective action over the $q_\ell$ since their rate function has a simple form, and construct the corresponding $\mathcal Q$ via Eq.~\eqref{eq:cnn-independent-innovations}. Note that the independent variables $q_\ell$ are matrices of shape $(r_\ell\times r_\ell)$ with $r_\ell$ the number of paths passing from layer $\ell$ to the output, and in this sense are not a purely local object of layer $\ell$. Zooming out, the forward and backward representations are equivalent at finite width within the EWA. For linear activations both are exact integral representations of the finite-width prior.}

\underline{\added[id=rev1]{Renormalized kernel and numerical evaluation.}} \added[id=rev1]{Conditional on any fixed value of the order parameters $\mathcal Q$ in the backward representation, the output prior becomes Gaussian. Its covariance is obtained by contracting the path indices of the fixed feature blocks with $\mathcal Q$, leading to the renormalization scheme of Eq.~\ref{eq:cnn-path-kernel-renorm}}
\begin{equation}
\added[id=rev1]{\relax [K^{(\mathrm R)}_{\mathcal Q}]^{\mu\nu}
 =\frac{1}{p_L\lambda}\sum_{\pi,\pi'}\mathcal H^{\mu\nu}_{\pi\pi'}\mathcal Q_{\pi\pi'}\,.}
\label{eq:app_cnn-path-kernel-renorm}
\end{equation}
\added[id=rev1]{For $\mathcal Q =\Id$, only equal paths contribute. Summing their filter offsets successively uses $Z_\ell\odot b_\ell=\Theta(Z_\ell)$ and recovers the large-width output kernel. A general $\mathcal Q$ instead changes the weights of these contributions and also couples distinct paths. Equation~\eqref{eq:app_cnn-path-kernel-renorm} therefore gives a spatially dependent change of the kernel.}

\added[id=rev1]{In path coordinates, the backward representation replaces the $Pr_\ell$-dimensional forward random kernels by $r_\ell$-dimensional order parameters. The effective action is given by combining the joint $\{q_1,...,q_L\}$ rate with the Gaussian evidence for square loss or its conditional Laplace approximation for cross-entropy, as in the rest of the manuscript.  During optimization, for each evaluation point in the independent basis of the $q_l$, we construct the chain of $\mathcal Q_L, \mathcal Q_{L-1},..., \mathcal Q$ and evaluate Eq.~\eqref{eq:app_cnn-path-kernel-renorm} from the fixed $\mathcal H$ blocks in order to evaluate the evidence term for that $K^{(\mathrm R)}_{\mathcal Q}$. The EWA theory curves use the saddle-point $Q_\ast$ of the effective action. App.~\ref{appsec:numerical-computation-of-the-saddle-point} describes the parametrization and minimization. The same kernel construction with training and test examples after fixing the saddle-point $Q_\ast$ supplies the covariance blocks for prediction.}

\underline{\added[id=rev1]{Pooling and multiple outputs.}} \added[id=rev1]{Global average pooling (GAP) averages the final activations before applying the readout weights. Therefore, it still depends on covariances between distinct final positions.  Let $\tau=(m_2,\ldots,m_L)$ denote the offsets without the final position, and write $\mathcal H^{\rm flat}$ for Eq.~\eqref{eq:cnn-path-feature}. The pooled block is}
\begin{equation}
\added[id=rev1]{\relax \mathcal H^{\mathrm{GAP},\mu\nu}_{\tau\tau'}
 =\frac{1}{p_L}\sum_{i,j=1}^{p_L}
 \mathcal H^{\mathrm{flat},\mu\nu}_{(\tau,i),(\tau',j)}\,.}
\label{eq:cnn-gap-feature}
\end{equation}\added[id=rev1]{\relax Together with the readout factor $1/(p_L\lambda_{L+1})$ in Eq.~\eqref{eq:app_cnn-path-kernel-renorm}, this gives the $1/(p_L^2\lambda_{L+1})$ normalization of the two spatial averages. The final spatial index is now absent, so $r_L=1$ and $r_\ell=\prod_{k=\ell+1}^L M_k$. The same conditional Wishart law and kernel contraction apply on these reduced path coordinates, as the GAP operation is linear. In particular, GAP leaves nontrivial spatial $\mathcal Q_\ell$ matrices in earlier layers even though the last-layer $\mathcal q_L$ is reduced to a single free parameter.}

\added[id=rev1]{\relax For $D$ outputs, the forward representation has $D$ conditionally independent Gaussian readouts of the same hidden features. The backward transform acts on combined path and output coordinates, with dimension $d_\ell=D r_\ell$. The laws in Eqs.~\eqref{eq:cnn-backward-Qs}--\eqref{eq:cnn-independent-innovations} hold with $r_\ell$ replaced by $d_\ell$. The scalar contraction in Eq.~\eqref{eq:app_cnn-path-kernel-renorm} becomes}
\begin{equation}
\added[id=rev1]{\relax [K^{(\mathrm R)}_{\mathcal Q}]_{\mu a,\nu b}
 =\frac{1}{p_L\lambda}\sum_{\pi,\pi'}\mathcal H^{\mu\nu}_{\pi\pi'}
       [\mathcal Q]_{\pi a,\pi'b}\,.}
\label{eq:cnn-multiple-output-kernel}
\end{equation}
\added[id=rev1]{\relax Setting $q_\ell=\Id_{d_\ell}$ at every layer gives independent outputs and the large-width kernel. At the posterior saddle the order parameters can also couple output coordinates, and allow a spatial renormalization that differs between outputs.}

\underline{\added[id=rev1]{Nonzero activation mean.}} \added[id=rev1]{An off-diagonal entry of the nonlinear kernel map $[\Theta(K)]^{\mu\nu}_{ij} = \Theta \big( [K]^{\mu\mu}_{ii}, [K]^{\nu\nu}_{jj}, [K]^{\mu\nu}_{ij})$ depends on the cross covariance $[K]^{\mu\nu}_{ij}$ and on the two variances $[K]^{\mu\mu}_{ii}$ and $[K]^{\nu\nu}_{jj}$. For ReLU, these variances determine the activation means, $m_{\ell,i}^\mu=\sqrt{[Z_\ell]^{\mu\mu}_{ii}/(2\pi)}$. In the following, we use a bold $\boldsymbol{m}_\ell$ to distinguish the $P$-dimensional mean activation vector from the filter mask offsets $m_\ell$ of the previous paragraphs. Now, the second moment $\Theta(Z_\ell)$  contains the mean outer product $\boldsymbol{m}_\ell\boldsymbol{m}_\ell^\top$ even at zero cross covariance. Dividing this contribution by $Z_\ell$ is singular in the gain construction Eq.~\eqref{eq:cnn-path-feature}, and we therefore proceed by separating the mean contribution from the centered gain}
\begin{equation}
\added[id=rev1]{[b_\ell^{\rm c}]^{\mu\nu}_{ij}
 =\frac{[\Theta(Z_\ell)]^{\mu\nu}_{ij}-m_{\ell,i}^\mu m_{\ell,j}^\nu}
 {[Z_\ell]^{\mu\nu}_{ij}}\,.}
\label{eq:cnn-centered-gain}
\end{equation}
\added[id=rev1]{At fixed positive variances, the centered covariance vanishes at zero cross covariance. The gain there is defined by continuity and equals the local slope in the cross-covariance argument, e.g. $1/4$ for ReLU. At nonzero reference covariance, of course this ratio gives corresponding local gain values, which in fact do not generally equal the derivative of the map. We show further below that this construction correctly recovers the MLP case for a single spatial position and thus a single path. The Hermite argument above, now applied to the centered activation, shows that $b_\ell^{\rm c}$ is still positive semidefinite on the joint example and position indices.}

\added[id=rev1]{We now keep both $\boldsymbol{m}_\ell$ and $b_\ell^{\rm c}$ frozen at the deterministic large-width values; no independent fluctuations of the mean features are introduced, which would require additional order parameters. This is consistent with the central-EWA reduction of the noncentral EWA in Sec.~\ref{sec:noncentral_EWA}, where the nonzero activation mean enters the reference second moment, and corrections specific to noncentral fluctuation statistics are suppressed in the proportional limit under the assumptions discussed there. Here the separation makes the multiplicative construction of $\mathcal H$ regular while retaining the mean contribution in the fixed kernel blocks, without the need to introduce additional order parameters: This is the CNN analogue of using the central EWA also for activation functions with nonzero mean.}

\added[id=rev1]{\relax To construct $\mathcal H$, the mean contribution must also be assigned to the path coordinates. For a fixed remaining path $\pi_\ell$, there are $\prod_{k=2}^{\ell}M_k$ choices of the earlier offsets. We here choose to distribute the average mean feature equally across all these origin paths, which could be called a coherent gauge. Tracing over earlier filter offsets reproduces one full copy of $\boldsymbol{m}_\ell\boldsymbol{m}_\ell^\top$. Write $\mathsf P_k K$ for the matrix obtained by evaluating $K$ on the positions visited at layer $k$, i.e., $[\mathsf P_k K]^{\mu\nu}_{\pi\pi'}=[K]^{\mu\nu}_{i_k(\pi),i_k(\pi')}$. To construct final $\mathcal H$, we iteratively construct the deterministic intermediate blocks $\mathcal H^{(\ell)}$. With example indices implicit, their recursion is}
\begin{equation}
\begin{aligned}
\added[id=rev1]{\relax \mathcal H^{(1)}} &\added[id=rev1]{\relax =\mathsf P_1 \Theta(Z_1),}\\
\added[id=rev1]{\relax \mathcal H^{(\ell)}} &\added[id=rev1]{\relax =
 \frac{\mathcal H^{(\ell-1)}\odot\mathsf P_\ell b_\ell^{\rm c}}{M_\ell\lambda_\ell}
 +\frac{\mathsf P_\ell(\boldsymbol{m}_\ell\boldsymbol{m}_\ell^\top)}{\prod_{k=2}^{\ell}M_k},
 \quad \ell=2,\ldots,L.}
\end{aligned}
\label{eq:cnn-mean-branches}
\end{equation}
\added[id=rev1]{The first term propagates the preceding block through the centered gain and the second inserts the fixed mean contribution at this layer. Both terms preserve positivity on the joint example and path indices. 
Averaging the final positions of $\mathcal H=\mathcal H^{(L)}$ as in Eq.~\eqref{eq:cnn-gap-feature} gives the GAP counterpart, and again the random chain $\widehat H_\ell$ in Eq.~\eqref{eq:cnn-forward-wishart} starts from the completed $\mathcal H$ matrix. Under successive offset traces, the two terms reconstruct the centered covariance and the mean outer product at each layer, so $\mathcal Q=\Id$ again recovers the large-width kernel. Both contributions enter the same contraction with $\mathcal Q$ in Eq.~\eqref{eq:app_cnn-path-kernel-renorm}, with the central Wishart distributions unchanged.}

\underline{\added[id=rev1]{\relax Reduction to the central EWA for MLPs.}} \added[id=rev1]{\relax Consider one spatial position per hidden layer and pointwise hidden-to-hidden convolutions, so $p_\ell=1$ and $M_\ell=1$. There is a single path, and the operator $\mathsf P_\ell$ and  partial trace $\Tr_{m_{\ell}}$ are identities. Starting from $\mathcal H^{(1)}=\Theta(Z_1)$, Eq.~\eqref{eq:cnn-mean-branches} gives by induction}
\begin{equation}
\added[id=rev1]{\relax \mathcal H^{(\ell)}
 =\frac{\Theta(Z_{\ell-1})}{\lambda_\ell}\odot b_\ell^{\rm c}
   +\boldsymbol{m}_\ell\boldsymbol{m}_\ell^\top
 =Z_\ell\odot b_\ell^{\rm c}+\boldsymbol{m}_\ell\boldsymbol{m}_\ell^\top
 =\Theta(Z_\ell)\,.}
\label{eq:cnn-single-position-limit}
\end{equation}
\added[id=rev1]{\relax Thus $\mathcal H=\Theta(Z_L)$, including for zero-mean activations by setting $\boldsymbol{m}_\ell=0$. For a scalar output, Eq.~\eqref{eq:cnn-independent-innovations} becomes $\mathcal Q_L=q_L$ and $\mathcal Q_\ell=q_\ell\mathcal Q_{\ell+1}$, with independent $q_\ell\sim\chi^2_{A_\ell}/A_\ell$. Consequently $K^{(\mathrm R)}=(\prod_{\ell=1}^L q_\ell)\Theta^L(C)$, exactly the MLP result of Eq.~\eqref{eq:renormalized_kernel}, with $A_\ell=N_\ell$ and the weight precisions part of $\Theta$. For $D$ outputs, the remaining $D\times D$ matrices obey the corresponding Wishart product construction of Sec.~\ref{subsec:multiple-outputs} and recover Eq.~\eqref{eq:multiple-output-kernelrenorm}. The two constructions therefore coincide at the level of the central-EWA prior, including the separated mean contributions, before any saddle-point approximation.}

\begin{figure}
    \centering
    \includegraphics[width=\linewidth]{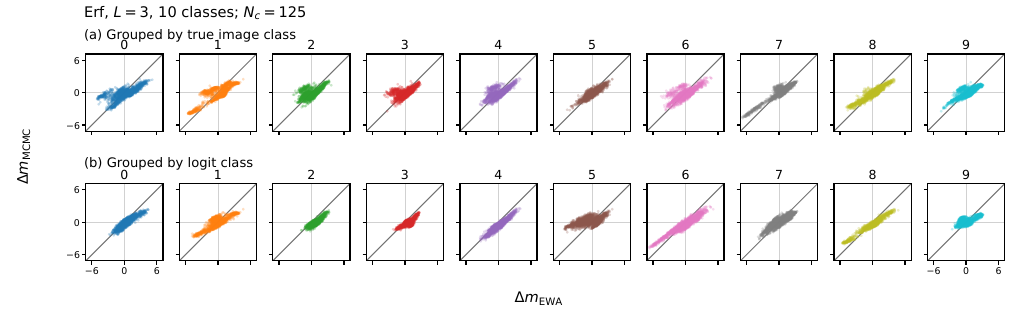}
    \caption[Class-resolved shifts of the mean logits.]{\textbf{\added[id=rev1]{Class-resolved shifts of the mean logits.}} \added[id=rev1]{For the three-layer Erf classifier in Fig.~\ref{fig:cnn}(c) at $N_c=125$, each point compares the EWA and sampled shifts of one posterior-mean centered logit relative to the large-width posterior. Both rows contain the same ten logits for each of the 2000 test images. The upper row groups points by the true image class, the lower row by the class of the logit coordinate. The diagonal indicates agreement.}}
    \label{Sfig:cnn-class-shifts}
\end{figure}

\begin{figure}
    \centering
    \includegraphics[width=\linewidth]{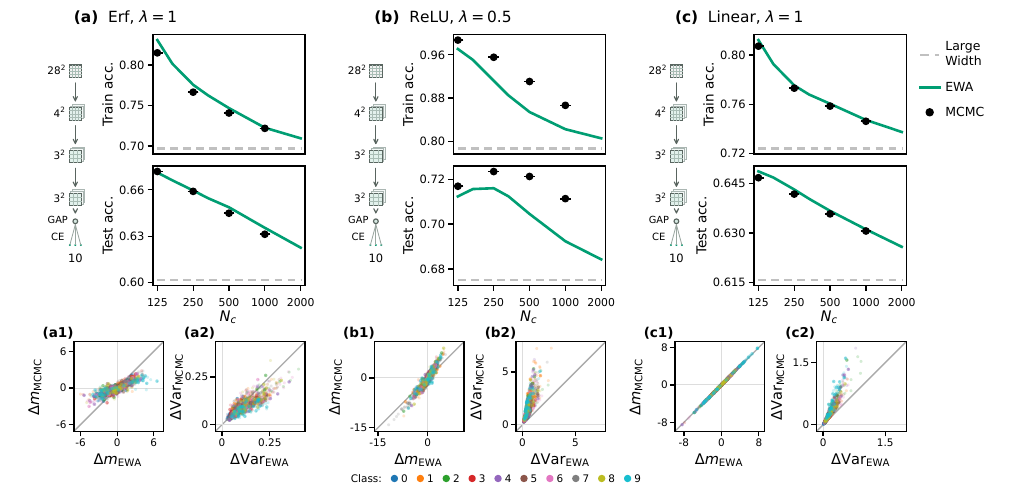}
    \caption[Ten-class classification at fixed GAP architecture.]{\textbf{\added[id=rev1]{Ten-class classification at fixed GAP architecture.}} \added[id=rev1]{All three CNNs shown have filters of size $7\times7,2\times2,1\times1$, strides $7,1,1$, and global average pooling in the last layer. The dataset and temperature are as in Fig.~\ref{fig:cnn}(c). From left to right the activations are Erf, ReLU and linear. The precision value $\lambda=0.5$ for ReLU prevents representation shrinkage across layers, to yield a fair comparison to Erf and Linear. The upper panels show posterior-averaged accuracies versus $N_c$. The lower panels compare mean and marginal variance shifts of the centered test logits at $N_c=125$ with the same axis conventions as Fig.~\ref{fig:cnn}. Colors denote the true class label. }}
    \label{Sfig:cnn-gap-activations}
\end{figure}

\underline{\added[id=rev1]{Further comparisons with posterior sampling.}} \added[id=rev1]{Figure~\ref{Sfig:cnn-class-shifts} shows the mean-logit shifts of Fig.~\ref{fig:cnn}(c1) resolved by class. Both splitting by true class (a) and by logit identity (b), a clear association between predicted and sampled changes is retained, with different spreads and slopes across classes. The aggregate test accuracy prediction close to the sampled value therefore coexists with appreciable spread in the prediction of individual means. It is still notable that the EWA largely correctly predicts the direction of mean-change on the level of individual test points in this finite-width CNN model, at a load $\alpha=4$ that is far from small.}

\added[id=rev1]{In Figure~\ref{Sfig:cnn-gap-activations} we use the three-layer CNN architecture with global average pooling of Fig.~\ref{fig:cnn}(a), but with 10 outputs and cross-entropy loss to test additional models in this challenging setting. With this overall CNN architecture fixed, Fig.~\ref{Sfig:cnn-gap-activations} compares the effect of three different activation functions: Erf (a), ReLU (b) and linear activations (c). For the Erf and linear models the accuracy curves track the samples. For ReLU the theory underestimates the training and test accuracies, to a large part due to an overestimation of the variance shifts (c2). For the linear CNN classifier (c) the Wishart distributions entering the EWA are exact, the only approximations introduced by the EWA theory line are the saddle-point approximation in the $Q$ and the Laplace approximation of the non-Gaussian logit posterior. The correspondence for mean shifts is close for this linear control, although its sampled variance shifts are larger than predicted. This appears to be a finite-size error introduced by the saddle-point approximation in the for this model $756$-dimensional space determining $\mathcal Q$. This substantial number arises largely from the multiplication by $9$ output logit contrasts with the number of $(4,1,1)$ GAP paths at the level of the respective hidden layers, giving matrix dimensions $36\times36$, $9\times9$ and $9\times9$ for $q_1, q_2, q_3$, respectively. The Laplace approximation does not seem to contribute strongly to the error, we verified on a subset of points that sampling the actual non-Gaussian posterior did not correct the variance prediction (not shown).}

\section{Details on datasets}
\label{appsec:details-on-datasets}

\underline{MNIST and CIFAR10.}
The datasets are filtered for $P$ training and $P_t=1000$ test samples of the selected categorical labels, \{“0”,“1”\} for MNIST and \{“cars”,“planes”\} for CIFAR10, and the labels mapped to the numerical values $y^\mu \in\{0,1\}$. In the case of CIFAR10, images where downsampled from $32\times 32$ to $28\times 28$ using bilinear interpolation with \texttt{torchvision.transforms.Resize()} and grayscaled, increasing slightly the difficulty of the task and resulting in the same $N_0=784$ input dimension as for MNIST.
The samples are then flattened and standardized by mean and standard deviation of the training set. To be precise, let $\mathcal{D}_{P}=\{ x^\mu, y^\mu\}_{i=1}^{P}$ denote the training dataset, where each $x^\mu \in \mathbb{R}^{N_0}$ is a flattened and grayscale input sample, $y^\mu\in\mathbb R$, and let $\mathcal{D}^{\text{test}}_{P_t}=\{ \tilde{x}^\mu, \tilde{y}^\mu\}_{i=1}^{P_t}$ denote the corresponding test dataset. The empirical mean and standard deviation are computed exclusively on the training set:
\begin{equation}
m_{\mathcal D} = \frac{1}{P N_0} \sum_{\mu=1}^{P} \sum_{i_0=1}^{N_0} x_{i_0}^\mu, 
\qquad
\sigma_{\mathcal D} = \sqrt{\frac{1}{P N_0} \sum_{\mu=1}^{P} \sum_{i_0=1}^{N_0} (x_{i_0}^\mu - m_{\mathcal D})^2}.
\end{equation}
The standardization is then applied to both training and test datasets using the same statistics:
\begin{equation}
    x^\mu \to \frac{x^\mu - m_{\mathcal{D}}}{\sigma_{\mathcal{D}}} \quad \forall\ x^\mu\in\mathcal{D}_P, \qquad
    \tilde{x}^\mu \to \frac{\tilde{x}^\mu - m_{\mathcal{D}}}{\sigma_{\mathcal{D}}} \quad \forall\ \tilde{x}^\mu\in\mathcal{D}_{P_t}^{\text{test}}. 
\end{equation}
This procedure ensures that the training data has global zero mean and unit variance, while the test data is transformed consistently using the training statistics, thereby avoiding any information leakage. 

\underline{\added[id=rev1]{FashionMNIST.}} \added[id=rev1]{Compared to MNIST, which is strongly clustered by class in the pixel space, and CIFAR10, which instead is a highly difficult task especially with reduced training set sizes, we use FashionMNIST as a suitable task of intermediate difficulty. For the experiments with all 10 classes in Figs.~\ref{fig:multioutput-mlp-depth},~\ref{fig:cnn} and ~\ref{Sfig:cnn-gap-activations}, we subsample class-balanced sets of training and test images from the respective official splits. For the experiments with scalar outputs, we used classes 2 and 4 ("Pullover" vs. "Coat") and mapped the labels to numerical values $y^\mu \in \{0,1\}$, as in the MNIST and CIFAR10 pipeline.
We standardize the pixels using the mean and standard deviation of the full training split. The CNN experiments use 2D convolutions and therefore get directly get the 28x28 pixel images as inputs, while for inputs to MLPs the images are flattened to $N_0=784$ dimensional vectors. }

\underline{Gaussian dataset.}
In addition to real datasets, we consider a synthetic linear regression task constructed as in classic teacher--student frameworks. Let again $N_0$ denote the input dimensionality, which in the experiments shown in the main text is set to $N_0 = 300$. The training inputs are independently drawn from an isotropic Gaussian distribution:
\begin{equation}
    x^\mu \sim \mathcal{N}(0, \Id_{N_0} ), \qquad \mu = 1, \dots, P,
\end{equation}
and analogously for the test inputs $\{\tilde{x}^\mu\}_{\mu=1}^{P_t}$. 
A fixed teacher vector $w^\ast \in \mathbb{S}^{N_0}$ is drawn from the unit sphere by sampling and normalizing a Gaussian vector $w^* \sim \mathcal{N}(0, \Id_{N_0}), \;\; w^* \rightarrow w^*/\|w^*\|$.
The corresponding labels are generated through the noiseless linear rule defined by the teacher vector:
\begin{equation}
    y^\mu = (x^\mu)^\top w^*, \qquad \tilde{y}^\mu = (\tilde{x}^\mu)^\top w^* \, .
\end{equation}
By construction, all $x^\mu_i$ and $y^\mu$ are standardized, and the marginal distributions of the labels inherit the Gaussian statistics of the inputs; since $\|w^*\| = 1$, it follows that $y^\mu \sim \mathcal{N}(0,1)$ and similarly for the test labels. The important point to note here is that there is no match between the teacher, a linear model, and the student architecture, a nonlinear MLP of depth $L$. Therefore, even though the task is a simple linear regression, from the perspective of teacher-student models it is a non-trivial task for the student MLP to solve, which suffers from prior mismatch.

\section{Numerical computation of the saddle-points}
\label{appsec:numerical-computation-of-the-saddle-point}

The code for theory and sampling experiments is available at Ref.~\cite{zenodo} and built around the GitHub package~\cite{deepbays}.
Finding saddle-points of the low-dimensional effective action is possible without significant difficulties, and requires negligible compute compared to the sampling experiments. 

\underline{Single-output theory.}
At non-zero temperature, evaluating the effective action Eq.~\eqref{eq:action_singleoutput} requires to compute the inverse $y^\top \big[ \beta^{-1} \Id + K^{(R)}_{\mathcal Q}\big]^{-1} y$ each time that $\mathcal Q =\prod_\ell q_\ell$ is changed, which determines the computational complexity of finding the saddle-points $q_\ell^\ast$. Note that at zero temperature the order parameters can be pulled out as $\mathcal Q^{-1} y^\top \,\big[\Theta^L(C)\big]^{-1} y$ and the contraction of the kernel inverse computed once can be reused. 

We implemented the kernel functions and effective action in \texttt{PyTorch}, where the most efficient routine to obtain a projection of an inverse matrix $K^{-1}y$ is \texttt{torch.linalg.solve()}, and also gradients of the action can be computed automatically. However, optimizing the runtime of the action evaluation was overall not necessary here. First we note that due to the absence of conjugate fields the stationary points of the action Eq.~\eqref{eq:action_singleoutput} are indeed minima and never saddle-points, which simplifies the optimization. We used \texttt{scipy.optimize.fsolve()} to perform the $L$-dimensional minimization over ${q_1 \dots q_L}$ in the general case, with initialization $q_\ell =1\; \forall \ell$ at the infinite-width value. Since in the settings shown here the widths $N_\ell$ are the same in all layers, also the action becomes symmetric and it is possible to optimize over $q_\ell = q\; \forall \ell$, reducing to a one-dimensional minimization. It can be shown that a unique minimum exists (at finite temperature where the kernel is always invertible). The minimum can be found using a fixed \texttt{scipy.optimize.brentq} bracket in log space $[\log q_\mathrm{min}, \log q_\mathrm{max}]$ with $q_\mathrm{min}=10^{-2}$, $q_\mathrm{max}=10^{3}$, and on the strictly monotonous transform $\mathrm{arcsinh}\big(S(e^{\log q})\big)$, which proved to converge very fast and reliably also for the $\mu$P settings where $q^\ast \gg1$. The hyperbolic arcsine and log are used here to improve the conditioning of $S(q)$, as the minimum can be sharp. Due to $q>0$ the $\log q$ always exists, while $S(q)$ can be negative and therefore $\mathrm{arcsinh}(S)$ is used.

\underline{Non-central theory for one hidden layer (App.~\ref{app:noncentral_EWA}).}
The non-central theory implementing Eq.~\eqref{eq:noncentral_1hl_action_expanded} was only needed for Fig.~\ref{Sfig:noncentral_Qs_and_Ms_analysis}. The action Eq.~\eqref{eq:noncentral_1hl_action_expanded} contains a conjugate variable $\bar Q$, and optimization is two dimensional with the stationary points being saddles and no longer minima. We used \texttt{scipy.optimize.fsolve()} with initialization at the infinite-width values $Q_0=1,\,\bar Q_0 =0$. Since the temperature is zero, the quantities defined in Eqs.~\eqref{eq:app_Myy}-\eqref{eq:app_GammaK} need to be computed only once.

\underline{\added[id=rev1]{Multiple outputs.}} \added[id=rev1]{For equal hidden-layer widths $N_\ell=N$, we determine $\mathcal Q^\ast$ by minimizing $N\mathcal I_L(\mathcal Q)-\log Z(Y\mid\mathcal Q)$, where $Z(Y\mid\mathcal Q)$ is the evidence given $\mathcal Q$. For an $r\times r$ matrix the Wishart product rate is $2\mathcal I_L(\mathcal Q)=L\Tr(\mathcal Q^{1/L})-\log\det\mathcal Q-Lr$. For square loss $r=D$, and the product structure of Eq.~\eqref{eq:multiple-output-kernelrenorm} allows us to evaluate the Gaussian evidence in the eigenbases of $\Theta^L(C)$ and $\mathcal Q$, thereby avoiding to construct the full dense $PD\times PD$ matrix. For both losses we use L-BFGS-B with analytic gradients in the independent entries of a real symmetric matrix $H$ in log space. Writing $\mathcal Q=\exp(H)$ ensures that $\mathcal Q$ is positive definite at every iteration. For each theory point, two initializations, one at $H=0$, are checked to give the same minimizer to numerical precision.}

\underline{\added[id=rev1]{Cross-entropy likelihood.}} \added[id=rev1]{A shift common to all $D$ logits of an example leaves its softmax probabilities unchanged. We therefore use the $r=D-1$ contrasts $g^\mu=B^\top f^\mu$, where the columns of $B$ form an orthonormal basis perpendicular to the vector of ones. The centered logits $Bg^\mu$ retain all pairwise logit differences, and the common shift integrates out of the evidence. For each trial $\mathcal Q$, let $K$ be the joint prior covariance of the training contrasts. We find their posterior mode $\hat g$ by damped Newton minimization of $\Phi(g)=\ell(g)+g^\top K^{-1}g/2$, with $\ell(g)=-\beta\sum_\mu\log\pi_{y^\mu}(Bg^\mu)$. Writing $W=\nabla^2\ell(\hat g)$, the Gaussian approximation has mean $\hat g$ and covariance $(K^{-1}+W)^{-1}$. The corresponding Laplace approximation to the conditional evidence gives}
\begin{equation*}
    \added[id=rev1]{-\log Z(Y\mid\mathcal Q)\simeq
    \Phi(\hat g)+\tfrac12\log\det\!\left(\Id+W^{1/2}KW^{1/2}\right)\,.}
\end{equation*}
\added[id=rev1]{The outer minimization over $\mathcal Q$ then uses the L-BFGS-B solver as described above. At $\mathcal Q^\ast$ we obtain the test distribution by Gaussian conditioning. For each example we use $8192$ scrambled Sobol points to integrate the indicator of correct classification over its marginal Gaussian distribution. Averaging these probabilities over the training or test examples gives the respective accuracies.}

\underline{CNN theory.}
\added[id=rev1]{We first precompute the large-width spatial covariances defining $\mathcal H$ (App.~\ref{app:cnn-spatial}). 
For implementation of the effective action, the independent $q_\ell$ in Eq.~\eqref{eq:cnn-independent-innovations} are used and we construct the cumulative matrices $\mathcal Q_\ell$ via that relation at each evaluation. For common width $N_c$, the leading negative log prior is $(N_c/2)\sum_\ell[\Tr q_\ell-\log\det q_\ell-d_\ell]$, where $d_\ell$ is the dimension of $q_\ell$. With unequal channel counts, the $\ell$-th term has prefactor $A_\ell/2$. We minimize the sum of this cost and the negative conditional log evidence jointly over the symmetric logarithms of the $q_\ell$, using L-BFGS-B with analytic gradients. The $q_\ell$ given by the matrix exponential of the optimization variables are therefore again guaranteed to be positive definite at each step. The conditional evidence is Gaussian for square loss and is evaluated by the logit Laplace approximation above for cross-entropy.  To evaluate the large-width (NNGP) limit, all innovations $q_\ell$ are set to the identity.}
\deleted[id=rev1]{For networks with convolutional layers, we optimized the effective action by performing gradient-based optimization using the Adam optimizer \texttt{torch.optim.Adam}. To enforce a symmetric matrix $\mathcal Q$ in the optimization, it is reshaped into a vectorized form containing only the true free degrees of freedom, and the derivatives are computed with respect to this vector. Starting from an initial vectorized guess, the optimization loop continues until either the maximum number of epochs is reached or the variation in the loss function remains below a prescribed tolerance threshold for several consecutive iterations. In our simulations, we used a learning rate of $5 \times 10^{-4}$, a tolerance of $10^{-6}$, and a maximum number of epochs equal to $1000$, which provided converged results.}

\begin{figure}
    \centering
    \includegraphics[width=\linewidth]{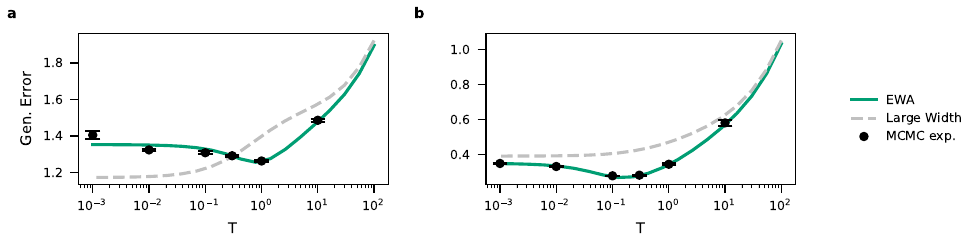}
    \caption{
    Experiments varying the temperature on the representative $L=5$, $N=200$, $P=200$ point with ReLU activation functions. The difference between the two panels is the dataset: (a) Random Gaussian data with linear teacher, at $T=0.1$ corresponds to $P=200$, $L=5$ from Fig.~\ref{fig:main-relu-gaussian}. (b) CIFAR-10, at $T=0.1$ corresponds to $P=200$, $L=5$ from Fig.~\ref{fig:main-relu-cifar}.
    Prior precision $\lambda=1/2$, and LMC sampler with learning rate $\eta=0.001$ as in the two related figures.
    }
    \label{Sfig:vary_temperature_Gaussian_and_Cifar}
\end{figure}

\begin{figure}
    \centering
    \includegraphics[width=\linewidth]{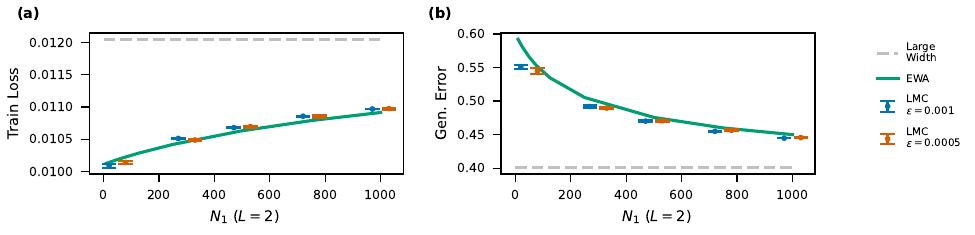}
    \caption{
    Analysis of the systematic effects arising from the finite learning rate on the representative points $L=2$, $P=250$, $P_t=1000$ with Erf activation function, varying the number of neurons $N_1=N_2=N$. In panel (a) we report the train loss, while in panel (b) the generalization error. Circles with error bars refer to numerical experiments using the LMC algorithm at different learning rates, $\eta=10^{-3}$ (red) and $\eta=5\cdot 10^{-4}$ (blue), against the EWA predictions (green solid lines) and the large-width limit (gray dashed lines). In both panels we consider a CIFAR-10 learning task at temperature $T=0.01$, prior precision $\lambda_0=1/5$ and $\lambda_1=\lambda_2=1.0$. Points indicating numerical experiments are computed at the same number of neurons and are shifted only for ease of viewing.}
    \label{Sfig:diff-lr}
\end{figure}

\begin{figure*}
    \includegraphics[width=\textwidth]{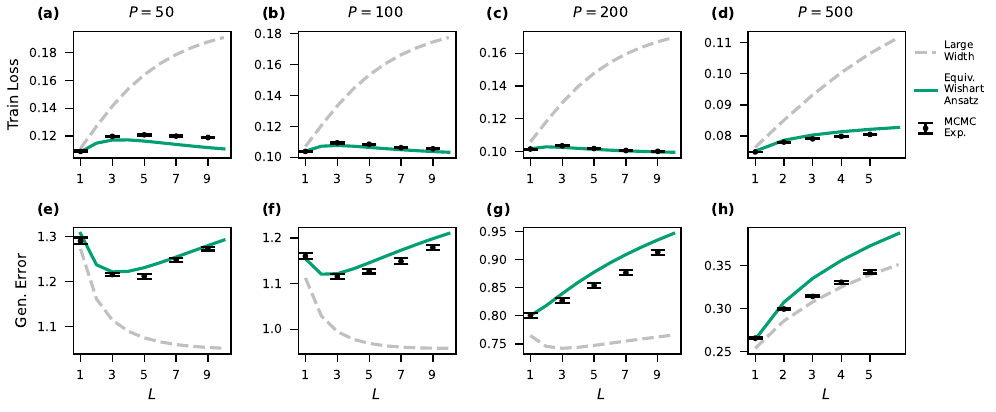} 
    \caption{Comparison between the learning curves obtained via the Equivalent Wishart Ansatz and numerical experiments for zero-mean activation functions on the Gaussian dataset. Numerical samples from the Bayesian posterior (black dots) are compared against the large-width limit predictions (gray dashed lines) and the results of the EWA theory (green solid lines). Both the training loss (first row) and the generalization error (second row) are displayed as a function of the number of hidden layers $L$. We keep the number of neurons and input dimensionality fixed at $N_\ell=200\, \forall \ell$ and $N_0 = 300$, while varying the number of patterns $P$ across different columns ($P$ is constant within each column). These simulations refer to Erf activation function, with Gaussian priors $\lambda = 1$ and temperature $T=0.1$. For all panels, we sample from the posterior using Langevin Monte Carlo with a learning rate $\eta=0.001$ and use $P_t = 1000$ test samples.}
    \label{fig:main-erf-gaussian}
\end{figure*}

\begin{figure*}
    \includegraphics[width=\textwidth]{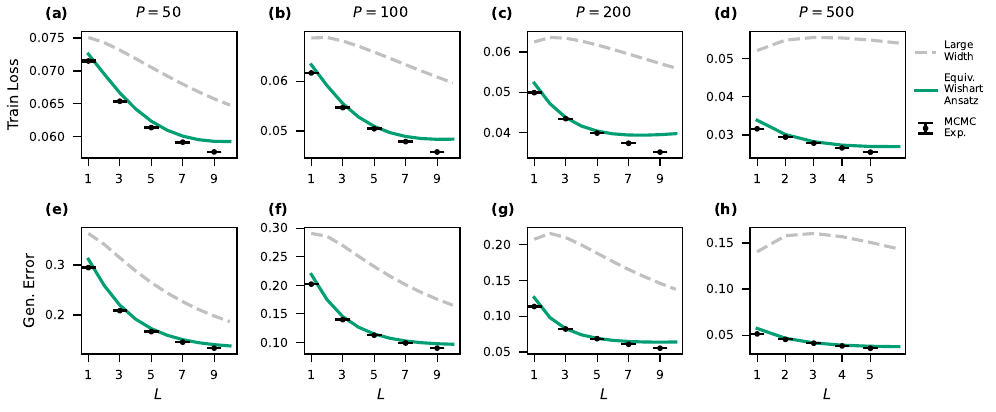} 
    \caption{Comparison between the learning curves obtained via the Equivalent Wishart Ansatz and numerical experiments for non-zero mean activation functions on the MNIST dataset. Numerical samples from the Bayesian posterior (black dots) are compared against the large-width limit predictions (gray dashed lines) and the results of the EWA theory (green solid lines). Both the training loss (first row) and the generalization error (second row) are displayed as a function of the number of hidden layers $L$. We keep the number of neurons and test examples fixed at $N_\ell=200\, \forall \ell$ and $P_t=1000$, while varying the number of patterns $P$ across different columns ($P$ is constant within each column). These simulations refer to ReLU activation function, with critical Gaussian priors $\lambda = 1/2$ and temperature $T=0.1$. For all panels, we sample from the posterior using Langevin Monte Carlo with a learning rate $\eta=0.001$.}
    \label{fig:main-relu-mnist}
\end{figure*}

\begin{figure*}
    \includegraphics[width=\textwidth]{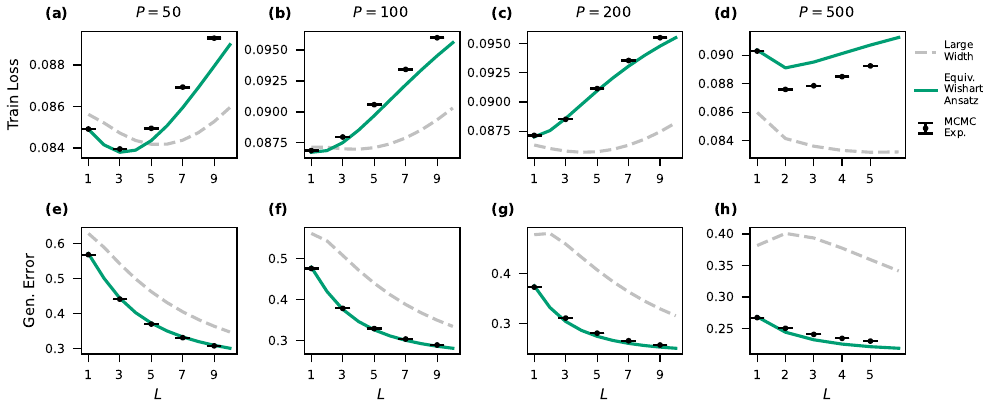} 
    \caption{Comparison between the learning curves obtained via the Equivalent Wishart Ansatz and numerical experiments for non-zero mean activation functions on the CIFAR-10 dataset. Numerical samples from the Bayesian posterior (black dots) are compared against the large-width limit predictions (gray dashed lines) and the results of the EWA theory (green solid lines). Both the training loss (first row) and the generalization error (second row) are displayed as a function of the number of hidden layers $L$. We keep the number of neurons and test examples fixed at $N_\ell=200\, \forall \ell$ and $P_t=1000$, while varying the number of patterns $P$ across different columns ($P$ is constant within each column). These simulations refer to ReLU activation function, with critical Gaussian priors $\lambda = 1/2$ and temperature $T=0.1$. For all panels, we sample from the posterior using Langevin Monte Carlo with a learning rate $\eta=0.001$.}
    \label{fig:main-relu-cifar}
\end{figure*}

\section{Interpretation of the EWA saddle-point at zero temperature} \label{app:zerotemp_saddle_behavior}

Here we describe how the saddle-point $q_1^\ast \dots q_L^\ast$ for single output MLPs behaves depending the relation between task and kernel structure. The interpretation is especially clear at zero temperature, due to a simplification of the minimization of the action. We start with the one-hidden layer case, where an explicit solution is available. For a lucid discussion in the deep linear network case, see also \cite{SompolinskyLinear} Suppl.Sects. I+II.

\subsection{Zero-temperature solution of the saddle-point for $L=1$}

At zero temperature and $L=1$, the effective action Eq.~\eqref{eq:action_singleoutput} is with $q:=q_1$
\begin{equation}
    S(q) = q - \log q 
    + \frac{\alpha}{P} \log \det \left[\beta K^{(\text{R})}_{q}\right] 
    + \frac{\alpha}{P} y^\top \left[K^{(\text{R})}_{q}\right]^{-1}y.
\end{equation}
With the definition of the scalar overlap $M_{yy}:= \frac{1}{P}\, y^\top \Theta(C)^{-1} y$ (compare the discussion of the non-central case around Eq.~\eqref{eq:noncental_1hl_action_zerotemp}), this becomes
\begin{equation}
    S(q) = q + (\alpha - 1) \log(q) + \alpha\frac{1}{q} M_{yy} +\mathrm{const.} \label{eq:app_1hlzerotemp_action}
\end{equation}
First, observe that in the EWA action the only term through which the data influence the saddle point is the scalar $M_{yy}$. This is because, while $\log \det \Theta(C)$ is data-dependent, it is a constant offset that does not depend on $q$. There is indeed a single minimum for all $\alpha, M_{yy} > 0$, both positive scalars by construction, given explicitly by the solution of a quadratic equation:
\begin{equation}
    q^\ast = -\frac{\alpha - 1}{2} + \sqrt{\left(\frac{\alpha - 1}{2}\right)^2 + \alpha M_{yy}}.  \label{eq:app_1hlzerotemp_saddlepoint}
\end{equation}
Only the positive branch of solutions is physical due to $q>0$, a physical constraint which is explicit in the log, but arises from the introduction of $Q=Nq$ as a $\mathcal{\chi}_N^2$-distributed variable. As expected from Eq.~\eqref{eq:app_1hlzerotemp_action}, we see that the saddle point is always growing with an increase in the task-kernel overlap $M_{yy}$. 

Note that the vicinity of $\alpha = 1$ seems at first glance prone to produce non-monotonic behavior, also because in Eq.~\eqref{eq:app_1hlzerotemp_action} the log term changes sign and thus pulls the solution either towards bigger or smaller $q$, respectively. However it turns out this is not the case, instead the solution is strictly monotonically increasing with $\alpha$ for $M_{yy}>1$, or monotonically decreasing for $M_{yy}<1$, as can be shown easily by calculating the derivative. The value of the solution $q^\ast$ indeed varies monotonically with the load $\alpha$ between the infinite-width value $q^\ast=1$ for $\alpha=0$ and the limiting value $q^\ast \to M_{yy}$ for $\alpha \to \infty$. The behavior of Eq.~\eqref{eq:app_1hlzerotemp_saddlepoint} for $M_{yy}\in\{1/5,\,1,\,5\}$ is shown in Fig.~\ref{Sfig:L-hl_zerotemp_saddlepoint}(a).

\underline{Interpretation in terms of generalization properties.}
Since in the zero-temperature case considered here, $q^\ast$ cancels out in the expression for the mean predictor Eq.~\eqref{eq:bias-definition}, while the predictor variance Eq.~\eqref{eq:sigma2-definition} is proportional to $q^\ast$, we obtain the following picture:
Compared to the infinite-width limit at zero temperature, in the proportional regime the EWA for shallow networks of $L=1$ predicts a change in the generalization error by modifying the predictor variance $\mathrm{Var}(\hat{y}_{0}) \propto  q^\ast$. 
If $M_{yy} > 1$, then $q^\ast > 1$ and the generalization error is increased compared to the infinite-width limit. If $M_{yy} < 1$, then $q^\ast < 1$ and the generalization error is decreased instead.
This behavior becomes more pronounced the larger $\alpha$ is, with a maximum factor $M_{yy}$ multiplying the infinite-width predictor variance at $\alpha \gg 1$ in both cases.

\subsection{Behavior of the zero-temperature saddle-point for general depth $L$}

\begin{figure}
    \centering
    \includegraphics[width=\linewidth]{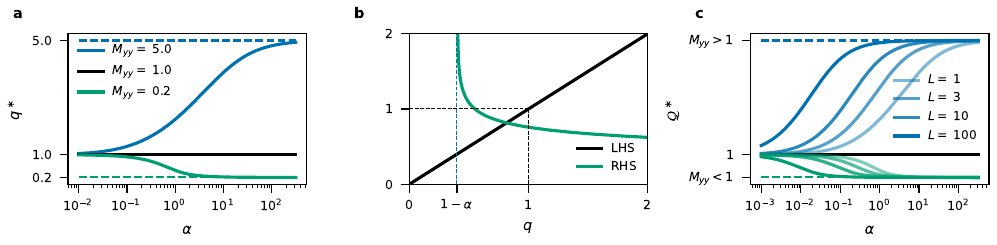}
    \caption{
    Behavior of the saddle point $\mathcal{Q}^\ast$ in the zero-temperature limit as a function of $\alpha$ and the task-kernel overlap $M_{yy}=\frac{1}{P}y^\top \big(\Theta^L(C)\big)^{-1}\, y$.
    (a) Monotonic behavior of $q^\ast$ in zero-temperature $L=1$ theory given by explicit solution Eq.~\eqref{eq:app_1hlzerotemp_saddlepoint}, for $ M_{yy}=5$ (blue), $M_{yy}=1$ (black) and $M_{yy}=1/5$ (green).
    (b) Visualization of the intersection between right- and left-hand sides of the implicit solution Eq.~\eqref{eq:app_Lhl_zerotemp_saddlepoint_implicit}. Increasing $\alpha$ shifts the pole of the RHS towards the left, $M_{yy}$ stretches the RHS in the y-direction. Here $\alpha=0.6$, $M_{yy}=1/4$, $L=5$.
    (c) Behavior of $\mathcal{Q}^\ast$ in the $L$-layer case given by implicit solution Eq.~\eqref{eq:app_Lhl_zerotemp_saddlepoint_implicit}, for $M_{yy}>1$ (blue), $M_{yy}=1$ (black) and $M_{yy}<1$ (green). The intensity of the lines encodes the depth, here with $L=[1, 3, 10, 100]$. Note that the characteristic range where $\mathcal{Q}^\ast$ transitions is at $\alpha_c \approx L^{-1}$.
    }
    \label{Sfig:L-hl_zerotemp_saddlepoint}
\end{figure}

The phenomenology of the one hidden layer action extends with very similar conclusions to the deep case, even though the resulting polynomial equation for $\mathcal Q^\ast$ does not  have a simple explicit solution.

For an $L$-layer MLP with single output dimension and rectangular aspect $N_\ell = N \quad \forall \ell$, the minimum of the action Eq.~\eqref{eq:action_singleoutput} at zero temperature becomes equivalent to the minimum of
\begin{equation}
    S_V[q] = L\, q + L(\alpha - 1) \log(q) + \alpha\frac{1}{q^L} M_{yy} +\mathrm{const.} \label{eq:app_Lhl_zerotemp_action}
\end{equation}
Here we used that $q^\ast_\ell = q^\ast \; \forall \ell$ due to the contraction principle, see Eq.~\eqref{eq:product-rate-function}, and the task-kernel overlap is now $M_{yy}=\frac{1}{P} y^\top \big(\Theta^L(C)\big)^{-1}\,y$. Compared to Eq.~\eqref{eq:app_1hlzerotemp_action}, the roots of the derivative are no longer determined by a quadratic function - instead we have the equation of state
\begin{align}
    \left. \frac{\partial S}{\partial q} \right|_{q^\ast} \overset{!}{=} 0 
    \qquad \Rightarrow \qquad 
    \begin{cases}
        q^\ast = 1,  &\quad \mathrm{if}\; \alpha M_{yy} = 0 \\
        (q^\ast)^L\, (q^\ast + \alpha - 1)  =  \alpha M_{yy}    
        & \quad \mathrm{if}\; \alpha M_{yy} \neq 0
      \end{cases} \label{eq:app_Lhl_zerotemp_eq-of-state}
\end{align}
Due to the insolvability of the quintics established by the Abel-Ruffini Theorem, for $L\geq 5$ there is in general no explicit solution to this equation. Nonetheless the implicit solution 
\begin{equation}
    q^\ast = \left( \frac{\alpha M_{yy}}{q^\ast + \alpha - 1} \right)^{\frac{1}{L}} \label{eq:app_Lhl_zerotemp_saddlepoint_implicit}
\end{equation}
allows us to show that the qualitative phenomenology of the $L=1$ case is preserved.
First, this equation still always has a unique solution: The positive half-space of the RHS (where $q^\ast > 1 - \alpha$) is a function decreasing monotonically from $+\infty$ to $0$, such that for any $\alpha$ there always is exactly one intersection with the LHS, and the negative half-space of the RHS can never intersect with the LHS for physical values $q^\ast > 0$.

Second, we find that 
$\bar{Q}_\ast = 1$ for $\alpha = 0$,
$\bar{Q}_\ast = \sqrt[L+1]{M_{yy}}$ for $\alpha = 1$, and
$\bar{Q}_\ast = \sqrt[L]{M_{yy}}$ for $\alpha \to \infty$; and monotonicity can again be confirmed by taking a derivative of the RHS.
Taking into account that for the deep case the factor multiplying the kernel, and correspondingly also the predictor variance, is $\mathcal{Q}^\ast = (q^\ast)^L$ we find qualitatively similar behavior as for $L=1$ also in the deep case, shown in Fig.~\ref{Sfig:L-hl_zerotemp_saddlepoint}(c).

Third, we ask how depth changes the size of $\mathcal Q^\ast$ at intermediate values of $\alpha$. Note that Eq.~\eqref{eq:app_Lhl_zerotemp_saddlepoint_implicit} can be solved for $\alpha$ as 
\begin{equation}
    \alpha(q^\ast,M_{yy},L) = \frac{q^\ast - 1}{(q^\ast)^{-L} M_{yy} \; - 1};
\end{equation}
plugging in the value $(q^\ast_h)^L = (M_{yy}+1)/2$ halfway between the two limiting values $\{1, M_{yy}\}$, this gives to first order in $1/L$
\begin{equation}
    \alpha(q^\ast_h,M_{yy},L) \propto \exp\left[\frac{1}{L} \log\bigg(\frac{M_{yy}+1}{2}\bigg)\right] -1 = \frac{1}{L} \log\bigg(\frac{M_{yy}+1}{2}\bigg) + O(L^{-2}).
\end{equation}
\underline{Interpretation in the $L$-layer case.} As for the shallow network, also in the deep case for $M_{yy}>1$ the generalization error is increased, and for $M_{yy}<1$ decreased with respect to the infinite-width limit. This is because the predictor variance $\mathrm{Var}(\hat{y}_{0}) \propto  \mathcal{Q}^\ast$, with a maximum factor $\mathcal{Q}^\ast\to M_{yy}$ for $\alpha\gg1$. 
The difference to the shallow case is that depth accelerates the transition to larger factors $\mathcal Q^\ast$ and therefore the effect of finite-width: The value $\alpha_h$ where $\mathcal Q^\ast$ is halfway between the limiting values $\{1, M_{yy}\}$ scales with depth as $\alpha_h \propto 1/L$. This again implies that the product $L\alpha$ is the effective load parameter controlling the size of finite-width effects in deep networks.

\begin{figure*}
    \includegraphics[width=\textwidth]{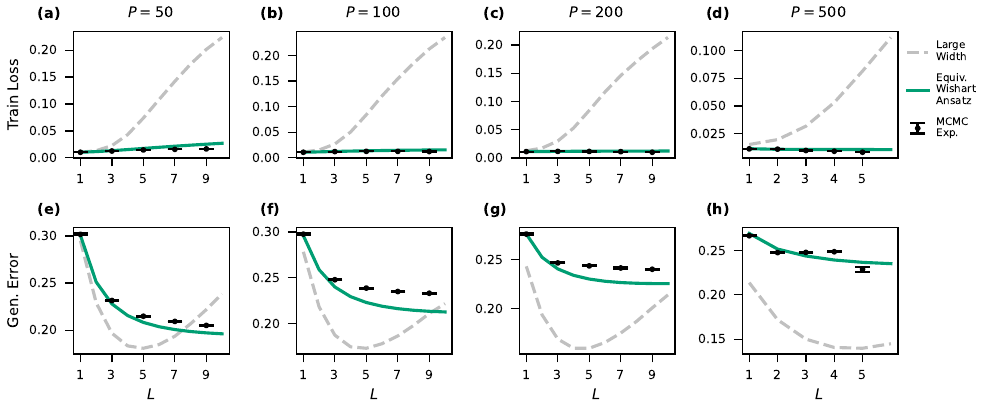} 
    \caption{\newtext{Comparison between the learning curves obtained via the Equivalent Wishart Ansatz and numerical experiments for ReLU networks away from criticality on the CIFAR-10 dataset. Numerical samples from the Bayesian posterior (black dots) are compared against the large-width limit predictions (gray dashed lines) and the results of the EWA theory (green solid lines). Both the training loss (first row) and the generalization error (second row) are displayed as a function of the number of hidden layers $L$. We keep the number of neurons and test examples fixed at $N_\ell = 200 \ \forall\ell$ and $P_t = 1000$, while varying the number of patterns $P$ across different columns ($P$ is constant within each column). These simulations refer to ReLU activation function, with non-critical Gaussian priors $\lambda=1$ and temperature $T = 0.01$. For all panels, we sample from the posterior using Langevin Monte Carlo with a learning rate $\eta=0.001$.}}
    \label{fig:off-crit}
\end{figure*}

\end{document}